\documentclass{article}

\usepackage[preprint]{neurips_2024}

\usepackage{enumitem}
\usepackage[utf8]{inputenc} 
\usepackage[T1]{fontenc}    
\usepackage{xcolor}         
\definecolor{darkblue}{rgb}{0, 0.12, 0.55}
\definecolor{darkgreen}{rgb}{0, 0.55, 0.12}
\definecolor{darkred}{rgb}{0.6,0,0}
\definecolor{darkgreen}{rgb}{0,0.6,0}
\definecolor{Gray}{gray}{0.9}
\usepackage[breaklinks=true,
            colorlinks,
            linkcolor = darkred,
            urlcolor  = darkblue, 
            citecolor = teal,
            bookmarks = false]{hyperref}
\definecolor{myblue}{HTML}{52aeb5}
\definecolor{myred}{HTML}{cf9d97}
\usepackage{url}            
\usepackage{booktabs}       
\usepackage{amsfonts}       
\usepackage{nicefrac}       
\usepackage{microtype}      
\usepackage{graphicx}
\usepackage{cleveref}
\usepackage{amsthm}
\usepackage{multirow}
\usepackage{booktabs}
\usepackage{wrapfig}
\usepackage{listings}
\usepackage{subcaption}
\usepackage{tikz}
\usepackage{tcolorbox}
\usepackage[absolute,overlay]{textpos}
\usepackage{transparent}
\usepackage{tcolorbox}
\usepackage{colortbl} 
\usepackage{natbib}
\usepackage{tipa}
\usepackage{tcolorbox}
\usepackage{colortbl} 
\usepackage{overpic}
\usepackage{bbding}
\usepackage{pifont}

\usepackage{etoc}
\etocdepthtag.toc{mtchapter}
\etocsettagdepth{mtchapter}{subsection}
\etocsettagdepth{mtappendix}{none}

\newcommand{\myPara}[1]{\vspace{.05in}\noindent\textbf{#1}}

\newtheorem{remark}{Remark}

\lstdefinestyle{pythonstyle}{
    language=Python,
    basicstyle=\ttfamily\small,
    keywordstyle=\bfseries\color{blue},
    stringstyle=\color{red},
    commentstyle=\color{gray},
    morecomment=[l][\color{magenta}]{\#},
    numbers=left,
    numberstyle=\tiny\color{gray},
    stepnumber=1,
    numbersep=10pt,
    backgroundcolor=\color{white},
    showspaces=false,
    showstringspaces=false,
    showtabs=false,
    frame=single,
    tabsize=4,
    captionpos=b,
    breaklines=true,
    breakatwhitespace=false,
    morekeywords={as,assert,async,await,break,continue,def,del,elif,except,finally,from,global,import,lambda,nonlocal,pass,raise,try,with,yield},
    escapeinside={(*@}{@*)},
}

\crefname{theorem}{Theorem}{Theorem}
\crefname{lemma}{Lemma}{Lemma}
\crefname{remark}{Remark}{Remark}
\crefname{figure}{Fig.}{Fig.}
\crefname{section}{Sec.}{Sec.}
\crefname{equation}{Eq.}{Eq.}
\crefname{table}{Tab.}{Tab.}
\crefname{algorithm}{Alg.}{Alg.}
\crefname{appendix}{Appendix}{Appendix} 

\title{
Pay Attention and Move Better: Harnessing Attention 
for Interactive Motion Generation and Training-free Editing
}

\author{%
Ling-Hao Chen$^{1, 2}$\thanks{This work was done while Ling-Hao Chen was an intern at IDEA Research.}, Shunlin Lu$^3$ , Wenxun Dai$^1$, Zhiyang Dou$^{4}$, Xuan Ju$^5$, Jingbo Wang$^6$\\
\textbf{Taku Komura}$^{4}$, \textbf{Lei Zhang}$^{2}$\thanks{Corresponding author.} \\
$^1$Tsinghua University, $^2$International Digital Economy Acadamy (IDEA Research)\\
$^3$The Chinese University of Hong Kong, Shenzhen, $^4$The University of Hong Kong\\
$^5$The Chinese University of Hong Kong, $^6$Shanghai AI Laboratory\\
Project page: \url{https://lhchen.top/MotionCLR}
}

\begin{document}

\maketitle

\begin{figure*}[!h]
    \centering
    \vspace{-3em}
    \includegraphics[width=\linewidth]{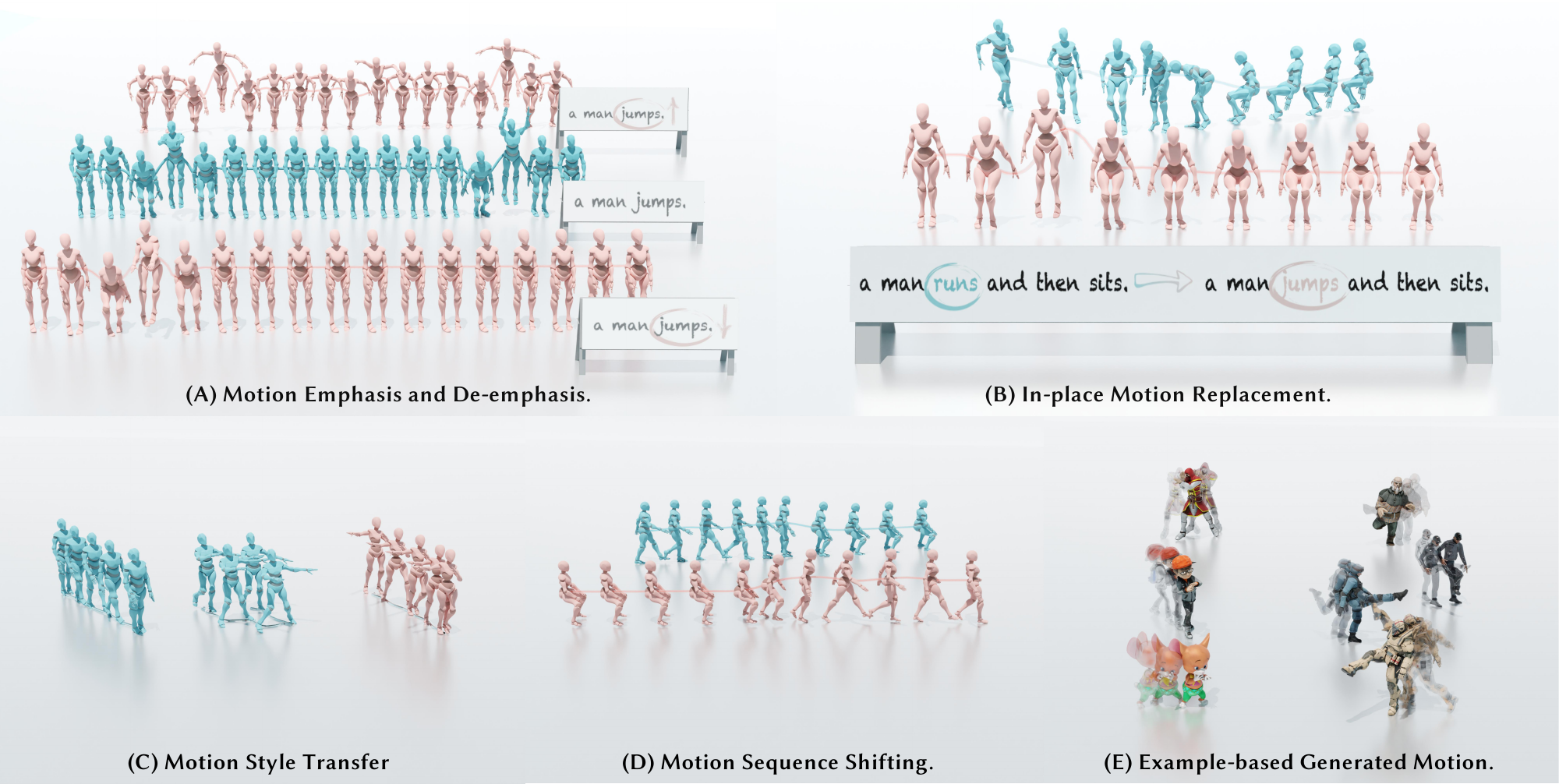}
    \vspace{-1.4em}
    \caption{{\bf We propose MotionCLR, supporting \textit{interactive} motion generation and versatile editing.} The \textbf{\textcolor{myblue}{blue}} and \textbf{\textcolor{myred}{red}} characters represent original and edited motions. 
    (A) Motion deemphasis and emphasis via adjusting the weight of ``\texttt{jump}''. 
    (B) In-place replacing the action of ``\texttt{\textcolor{myblue}{runs}}'' with ``\texttt{\textcolor{myred}{jumps}}''. 
    (C) Transferring motion style referring to two motions. From left to right, there are \textcolor{myblue}{motion style reference}, \textcolor{myblue}{motion texture reference}, and \textcolor{myred}{transferred motion}.
    (D) Shifting the order of ``\texttt{walking}'' and ``\texttt{sitting}'' actions in a motion.
    (E) Generating diverse motion with the same example motion, \aka example-based motion generation or crowd animation. The left crowd is boxing animation and the right crowd is kicking animation. 
    }
    \vspace{-0.8em}
    \label{fig:teaser}
\end{figure*}
\begin{abstract}
This research delves into the problem of interactive editing of human motion generation. Previous motion diffusion models lack explicit modeling of the word-level text-motion correspondence and good explainability, hence restricting their fine-grained editing ability. To address this issue, we propose an attention-based motion diffusion model, namely MotionCLR, with CLeaR modeling of attention mechanisms. Technically, MotionCLR models the in-modality and cross-modality interactions with self-attention and cross-attention, respectively. More specifically, the self-attention mechanism aims to measure the sequential similarity between frames and impacts the order of motion features. By contrast, the cross-attention mechanism works to find the fine-grained word-sequence correspondence and activate the corresponding timesteps in the motion sequence. Based on these key properties, we develop a versatile set of simple yet effective motion editing methods via manipulating attention maps, such as motion (de-)emphasizing, in-place motion replacement, and example-based motion generation, \etc. For further verification of the explainability of the attention mechanism, we additionally explore the potential of action-counting and grounded motion generation ability via attention maps. Our experimental results show that our method enjoys good generation and editing ability with good explainability. 
\end{abstract}

\section{Introduction}
\label{sec:intro}

Recently, text-driven human motion generation~\citep{text2action, temos, mdm, humantomato, momask, avatarclip, humanise, move} has attracted significant attention in the animation community for its great potential to benefit versatile downstream applications, such as games and embodied intelligence. As the generated motion quality in one inference might be misaligned with the users' intents, interactive motion editing plays a crucial role in the community by introducing humans into the loop of human-machine interaction.

To this end, some attempts have been made to edit or control the generated motion by specifying sparse motion signals, like 3D joint trajectories~\citep{omnicontrol,motionlcm,diffprior} or motion clips~\citep{tang2022real,robustbetweening}. Despite such progress, the constraints introduced in these works are mainly in-modality (motion) constraints, which \textit{require laborious efforts in the real animation creation pipeline}. Such interaction fashions strongly restrict the involving humans in the loop of creation. In this work, we aim to develop a more user-friendly editing fashion of introducing out-of-modality signals, such as editing texts. For example, when generating a motion with the prompt ``\texttt{a man jumps}.'', we can control the height or times of the ``\texttt{jump}'' action via adjusting the importance weight of the word ``\texttt{jump}''. Alternatively, we can also \textit{in-place} replace the word ``\texttt{jump}'' with other actions specified by users. The retrieval-based data annotation process of previous work~\citep{athanasiou2024motionfix} not only restricts the semantic editing applications, but also requires expendable labor. In contrast, in this work, we would like to equip the motion generation model with such abilities in a training-free style. 

\begin{figure}[!t]
    \centering
    \includegraphics[width=\linewidth]{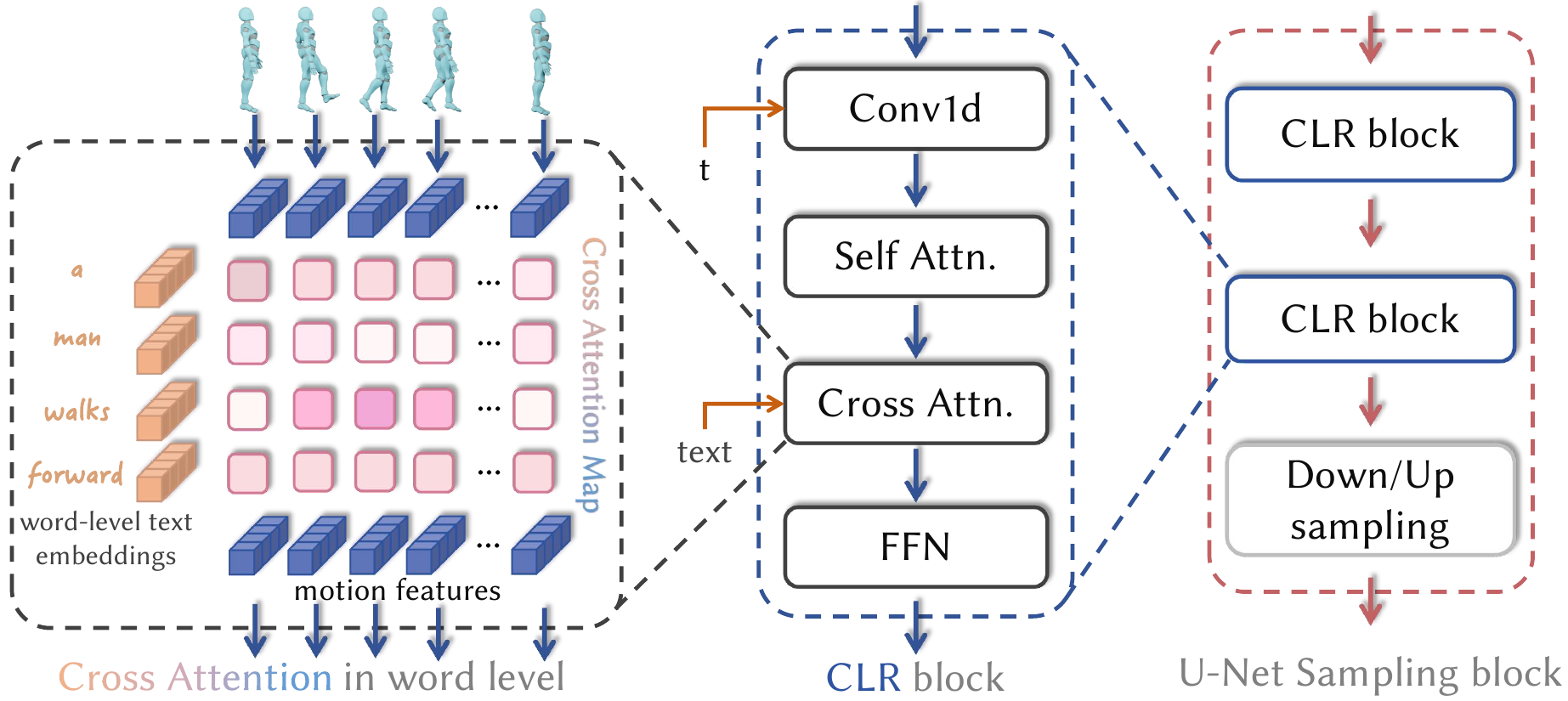}
    \caption{\textbf{System overview of MotionCLR architecture.} (a) The U-Net-like denoising network is with two CLR blocks before down/up-sampling. (b) The basic CLR block includes four layers, separating the timestep injection and the text condition. (c) The key component is the text-motion cross-attention at the word level. }
    \label{fig:enter-label}
\end{figure}

However, the key limitation of existing motion generation models is that the modeling of previous generative methods lacks \textbf{word-level} text-motion correspondence. This fine-grained cross-modality modeling not only plays a crucial role in text-motion alignment, but also makes it easier for fine-grained editing. To show the problem, we revisit previous transformer-based motion generation models~\citep{mdm, motiondiffuse, remodiffuse,mld}. The transformer-encoder-like methods~\citep{mdm,emdm} treat the textual input as one special embedding before the motion sequence. However, such integrated text embeddings and motion embeddings imply substantially different semantics, indicating unclear correspondence between the semantics of specific words and motions. Besides, this fashion over-compresses a sentence into one embedding, which compromises the fine-grained correspondence between each word and each motion frame, as evidenced in~\cref{fig:enter-label}. Although there are some methods~\citep{motiondiffuse, remodiffuse} to perform texts and motion interactions via linear cross-attention, they fuse the diffusion timestep embeddings with textual features together in the forward process. This operation undermines the structural text representations and weakens the input textual conditions (Sec.~\cref{fig:enter-label}). Through these observations, we argue that the fine-grained text-motion correspondence in these two motion diffusion fashions \textbf{is not well explored}. 

To resolve these issues, in this work, we propose a motion diffusion model, namely \underline{MotionCLR}, with a \underline{CL}ea\underline{R}\footnote{For clarification, the word ``Clear'' here means good explainability of the model.} modeling of the motion generation process and word-motion correspondence. The main component of MotionCLR is a CLR block, which is composed of a convolution layer, a self-attention layer, a cross-attention layer, and an FFN layer. In this basic block, the cross-attention layer is used to encode the text conditions \textbf{for each word}. More specifically, the cross-attention operation between each word and each motion frame models the text-motion correspondence at the word level. Meanwhile, the timestep injection of the diffusion process and the text encoding are modeled separately. Besides, the self-attention layer in this block is designed for modeling the interaction between different frames and the FFN is a common design for channel mixing.

Motivated by previous progress in the explainality of the attention mechanism~\citep{attention,ma2023visualizing,hao2021self,xu2015show,prompt-to-prompt,chefer2021transformer,chefer2021generic}, this work delves into the mathematical properties of the basic CLR block, especially the cross-attention and self-attention mechanisms. In the CLR block, the cross-attention value of each word along the time axis works as an activator to determine the execution time of each action. Besides, the self-attention mechanism in the CLR block mainly focuses on mining similar motion patterns between frames. Our empirical studies verify these properties. Although there are some early explanations~\citep{momo} about self-attention in motion generation, understanding both cross and self-attentions in one system is still unexplored, which mainly owes to the ``clear'' and separate modeling of both attention mechanisms. Based on these key observations, we show how we can achieve semantic motion editing tasks, \eg \textit{motion (de-)emphasis}, \textit{in-place motion replacement}, and \textit{motion erasing} by manipulating cross-attention. Additionally, our method can also be applied to other applications like \textit{style transfer}, \textit{sequence shifting}, and \textit{generating motions from an example} by manipulating self-attention calculations. We verify the effectiveness of these editing methods via both qualitative and quantitative experimental results. Additionally, we explore how our method can be applied to cope with the hallucination of generative models.

Before delving into the technical details of this work, we summarize our key contributions as follows. 
\begin{itemize}[left=0pt]
    \item {
        We propose an attention-based motion diffusion model, namely MotionCLR, with clear modeling of the text-aligned motion generation process. 
    }
    \item {
        For the first time in the human animation community, we clarify the roles that self- and cross-attention mechanisms play in \textit{one} attention-based motion diffusion model.
    }
    \item {
    
        Thanks to these observations, we propose a series of interactive motion editing tasks~(see~\cref{fig:teaser}) via manipulating attention layers. Importantly, both motion generation and editing are performed in \textit{one} model. We additionally explore the potential of \textit{grounded} motion generation when facing hallucination~(\cref{sec:failcase}). 
    }
    \item {
        We evaluate the generation quality of our method and achieve comparable generation performance with state-of-the-art methods. Besides, the training-free editing methods \textit{even outperform} baselines requiring specific training in some scenarios.
    }
\end{itemize}

\section{Related Work and Contribution}

\subsection{Text-driven Human Motion Generation}

Previous text-driven human motion generation~\citep{plappert2018learning, text2action, dvgans,  jl2p, t2g, motionclip, temos, avatarclip, tm2t, motiondiffuse, teach, mdm, humanise, mld, mofusion, physdiff, t2mgpt, diffprior, remodiffuse, gmd, motiongpt, omnicontrol, humantomato, tlcontrol, promotion, emdm, stmc, flowmdm, move} uses textual descriptions as input to synthesize human motions. One of the main generative fashions is a kind of GPT-like~\citep{t2mgpt, humantomato, momask, motiongpt} motion generation method, which compresses the text input into one conditional embedding and predicts motion in an auto-regressive fashion. Besides, the diffusion-based method~\citep{mdm, motiondiffuse, remodiffuse,emdm,mld,motionlcm} is another generative fashion in motion generation. Note that most work with this fashion also utilizes transformers~\citep{attention} as the basic network architecture. Although these previous attempts have achieved significant progress in the past years, the technical design of the explainability of the attention mechanism is still not well considered. 
We hope this work will provide a new understanding of these details. Besides, previous motion generation models also lack the capability of zero-shot motion editing, which is what we would like to explore in this work.

\subsection{Human Motion Editing}

The human motion editing task aims to edit a motion satisfying human demand. Previous works~\citep{motionlcm,mofusion,kim2023flame} attempt to edit a motion in a controlling fashion, like motion inbetweening and joint controlling. There are some other methods~\citep{modi,aberman2020unpaired,jang2022motion} trying to control the style of a motion. However, these works are either designed for a specific task or cannot edit fine-grained motion semantics, such as the height or times of a ``jump'' motion. \citet{momo} perform motion following via replacing the queries in the self-attention, which does not consider the semantic manipulations. \citet{goel2024iterative} propose to edit a motion with an instruction. However, the MEOs pipeline relies on the text-only LLM outputs, which will introduce hallucinations. MotionFix~\citep{athanasiou2024motionfix} proposes to use the language command to edit motions. However, it needs annotations on the editing text, additionally requiring more labor efforts. COMO~\citep{como} introduces editing motion via Large Language Models (LLMs) as a translator, which lacks the editing grounds of original motion content. The key reason why the existing method cannot achieve training-free motion editing is that the fine-grained text-motion correspondence in the cross-attention still lacks an in-depth understanding. There are also some methods designed for motion generation~\citep{motion_texture} or editing~\citep{hierarchical_motion,holden2016deep}, which are limited to adapt to diverse downstream tasks. 

Our key insights and contribution over previous attention-based motion diffusion models~\citep{mdm, motiondiffuse, remodiffuse,emdm,mld,motionlcm} lie in the clear explainability of the self-attention and cross-attention mechanisms in diffusion-based motion generation models. The cross-attention module in our method models the text-motion correspondence at the \textit{word level} explicitly. Besides, the self-attention mechanism models the motion coherence between frames. Therefore, we can easily clarify what roles self-attention and cross-attention mechanisms play in this framework, respectively. To the best of our knowledge, it is the first time in the human animation community to clarify these mechanisms in one system and explore how to perform training-free motion editing involving humans in the loop. 


\subsection{Visual Editing for Image Contents}

Image editing in diffusion models has been more explored than motion editing. Previous studies have achieved exceptional realism and diversity in image editing~\citep{prompt-to-prompt, han2023improving, parmar2023zero, masactrl, tumanyan2023plug, zhang2023real, mou2023dragondiffusion,ju2024pnp} by manipulating attention maps. Especially, although \citet{prompt-to-prompt} proposes to introduce cross-attention into image editing, these techniques and self-attention-based motion editing are still under-explored. However, relevant interactive editing techniques and observations are still unexplored in the human animation community. The basic reason for insufficient exploration for human animation is lacking a fine-grained modeling between words and motions.


\section{Base Motion Generation Model and Understanding Attention Mechanisms}

In this section, we will introduce the proposed motion diffusion model, MotionCLR, composed of several basic CLR modules. Specifically, we will analyze the technical details of the attention mechanism to obtain an in-depth understanding of this.  

\subsection{How Does MotionCLR Model Fine-grained Cross-modal Correspondence?}

Regarding the issues of the previous methods (see~\cref{sec:intro}), we carefully design a simple yet effective motion diffusion model, namely MotionCLR, with \textbf{fine-grained word-level text-motion correspondence}. The MotionCLR model is a U-Net-like architecture~\citep{unet}. Here, we name the down/up-sampling blocks in the MotionCLR as sampling blocks. Each sampling block includes two CLR blocks and one down/up-sampling operation. In MotionCLR, the atomic block is the CLR block, which is our key design. Specifically, a CLR block is composed of four modules, 

\begin{itemize}[left=0em]
    \item {
        \textbf{Convolution-1D module}, \aka \texttt{Conv1d}$(\cdot)$, is used for  timestep injection, which is disentangled with the text injection. The design principle here is to disentangle the text embeddings and the timestep embeddings for explicit modeling for both conditions. 
    }
    \item {
        \textbf{Self-attention module} is designed for learning temporal coherence between different motion frames. Notably, different from previous works~\citep{mdm,emdm,diffprior}, self-attention only models the correlation between motion frames and does not include any textual inputs. \textit{The key motivation here is to separate the motion-motion interaction from the text-motion interaction of traditional fashions~\citep{mdm}.}
    }
    \item {
        \textbf{Cross-attention module} plays a crucial role in learning text-motion correspondence in the CLR block. It takes word-level textual embeddings of a sentence for cross-modality interaction, aiming to obtain \textit{fine-grained} text-motion correspondence \textit{at the word level}. Specifically, \textit{the attention map models the relationship between each frame and each word, enabling more fine-grained cross-modality controlling}. 
    }
    \item {
        \textbf{FFN module} works as an additional feature transformation and extraction~\citep{dai2022knowledge,geva2021transformer}, which is a necessary component in transformer-based architectures. 
    }
\end{itemize}
\textit{In summary, in the basic CLR block, we model interactions between frames and word correspondence, separately in cross-attention.} We analyze both self-attention and cross-attention of MotionCLR in the following sections, which is useful for subsequent editing tasks.

\subsection{Mathematical Preliminaries of Attention Mechanism}

\textbf{The general attention mechanism} has three key components, query ($\mathbf{Q}$), key ($\mathbf{K}$), and value ($\mathbf{V}$), respectively. The output $\mathbf{X}'$ of the attention mechanism can be formulated as, 
\begin{equation}
    \mathbf{X}' = \texttt{softmax}(\mathbf{Q}\mathbf{K}^{\top}/\sqrt{d})\mathbf{V},
    \label{eq:attn}
\end{equation}
where $\mathbf{Q}\in \mathbb{R}^{N_1 \times d}$, $\mathbf{K}, \mathbf{V}\in \mathbb{R}^{N_2 \times d}$. Here, $d$ is the embedding dimension of the text or one-frame motion. In the following section, we take $t = 0, 1, \cdots, T$ as diffusion timesteps, and $f = 1, 2, \cdots, F$ as the frame number of motion embeddings $\mathbf{X}\in \mathbb{R}^{F\times d}$. For convenience, we name  $\mathbf{S}=\mathbf{Q}\mathbf{K}^{\top}$ as the similarity matrix and $\mathbf{A}=\texttt{softmax}(\mathbf{Q}\mathbf{K}^{\top}/\sqrt{d})$ as the attention map in the following sections.

\begin{figure*}[!t]
    \centering
    \includegraphics[width=\linewidth]{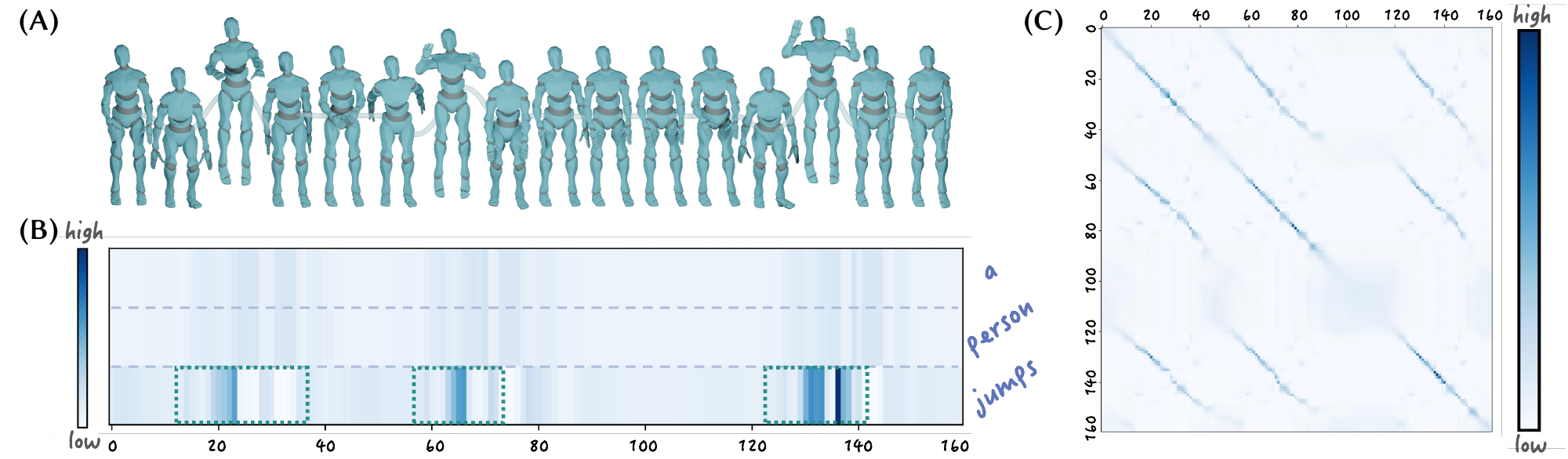}
    \caption{\textbf{Empirical study of attention mechanisms.} We use ``\texttt{\textcolor{myblue}{a person jumps.}}'' as an example. (\textbf{A}) Keyframes and the root trajectory of generated motion. The character jumps on $\sim 15-40$f, $\sim 60-80$f, and $\sim 125-145$f, respectively. (\textbf{B}) The \textbf{cross-attention} map between timesteps and words. The ``\textcolor{myblue}{jump}'' word is highly activated aligning with the ``\textcolor{myblue}{jump}'' action. (d) The \textbf{self-attention} map visualization. It is obvious that the character jumps three times, reflecting nine areas in the self-attention map. Different jumps share similar local motion patterns.}
    \label{fig:study_attn}
\end{figure*}


\subsection{Self and Cross Attention Mechanisms in MotionCLR}

\label{sec:clearattn}

\textbf{The self-attention} mechanism uses different transformations of motion features $\mathbf{X}$ as inputs,
\begin{equation}
    \mathbf{Q} = \mathbf{X}\mathbf{W}_{Q}, \ \ \mathbf{K} = \mathbf{X}\mathbf{W}_{K}, \ \ \mathbf{V} = \mathbf{X}\mathbf{W}_{V},
    \label{eq:selfattn}
\end{equation}
where $\mathbf{Q}, \mathbf{K}, \mathbf{V}\in \mathbb{R}^{F \times d}$, $F = N_1 = N_2$. We take a deep look at the formulation of the self-attention mechanism. As shown in~\cref{eq:attn}, the attention calculation begins with a matrix multiplication operation, meaning the similarity ($\mathbf{S} = \mathbf{QK}^{\top} \in \mathbb{R}^{F\times F}$) between $\mathbf{Q}$ and $\mathbf{K}$. Specifically, for each row $i$ of $\mathbf{S}$, it obtains the frame most similar to frame $i$. Here $\sqrt{d}$ is a normalization term. After obtaining the similarity for all frames, the $\mathtt{softmax}(\cdot)$ operation is not only a normalization function, but also works as a ``soft'' $\mathtt{max}(\cdot)$ function for selecting the frame most similar to frame $i$. Assuming the $j$-th frame is selected as the frame most similar to frame $i$ with the maximum activation, the final multiplication with $\mathbf{V}$ will approximately replace the motion feature $\mathbf{V}_j$ at the $i$-th row of $\mathbf{X}'$. Here, the output $\mathbf{X}'$ is the updated motion feature. In summary, we have the following remark.
\begin{table}[!ht]
\vspace{-0.5em}
\begin{minipage}{0.99\columnwidth}\vspace{0mm}    \centering
    \begin{tcolorbox}[left=1em, right=1em]
        \small
    \vspace{-0.6em}
        \begin{remark}
            The self-attention mechanism measures the motion similarity of all frames and aims to select the most similar frames in motion features at each place. 
        \end{remark}
    \vspace{-0.9em}
    \end{tcolorbox}
    \label{tab:remark1}
\end{minipage}
\vspace{-1em}
\end{table}

\textbf{The cross-attention} mechanism of MotionCLR uses the transformation of a motion as a query, and the transformation of textual words as keys and values, 
\begin{equation}
    \mathbf{Q} = \mathbf{X}\mathbf{W}_{Q}, \ \ \mathbf{K} = \mathbf{C}\mathbf{W}_{K}, \ \ \mathbf{V} = \mathbf{C}\mathbf{W}_{V},
    \label{eq:crossattn}
\end{equation}
where $\mathbf{C} \in \mathbb{R}^{L \times d}$ is the textual embeddings of $L$ word tokens, $\mathbf{Q} \in \mathbb{R}^{F \times d}$, $\mathbf{K}, \mathbf{V} \in \mathbb{R}^{L \times d}$. Note that $\mathbf{W}\hspace{-0.2em}\underline{{}_{\mathbf{\star}}}$ in~\cref{eq:selfattn} and~\cref{eq:crossattn} are not the same parameters, but are used for convenience. 
As shown in~\cref{eq:crossattn}, $\mathbf{K}$ and $\mathbf{V}$ are both the transformed text features. Recalling~\cref{eq:attn}, the matrix multiplication operation between $\mathbf{Q}$ and $\mathbf{K}$ measures the similarity ($\mathbf{S} = \mathbf{QK}^{\top}$) between motion frames and words in a sentence. Similar to that in self-attention, the $\mathtt{softmax}(\cdot)$ operation works as a ``soft'' $\mathtt{max}(\cdot)$ function to select which transformed word embedding in $\mathbf{V}$ should be selected at each frame. This operation models the motion-text correspondence explicitly. Therefore, we have the second remark.
\begin{table}[!h]
\vspace{-0.5em}
\begin{minipage}{0.99\columnwidth}\vspace{0mm}    \centering
    \begin{tcolorbox}[left=1em, right=1em]
        \small
    \vspace{-0.6em}
        \begin{remark}
            The cross-attention first calculates the similarity to determine which word (\ie value in cross attention) should be activated at the $i$-th frame. The final multiplication operation with values places the semantic features of their corresponding frames. 
        \end{remark}
    \vspace{-1.0em}
    \end{tcolorbox}
    \label{tab:remark2}
\end{minipage}
\vspace{-1em}
\end{table}

\subsection{Empirical Evidence on Understanding Attention Mechanisms}

To obtain a deeper understanding of the attention mechanism and verify the mathematical analysis of attention mechanisms, we provide some empirical studies on some cases.

As shown in~\cref{fig:study_attn}, we take the sentence ``\texttt{a person jumps.}'' as an example. We visualize the keyframe in \cref{fig:study_attn}(A), where we also visualize the root trajectory. As can be seen in~\cref{fig:study_attn}(A), the character jumps at $\sim 15-40$f, $\sim 60-80$f, and $\sim 125-145$f, respectively. Note that, as shown in~\cref{fig:study_attn}(B), the word ``\texttt{jump}'' is significantly activated aligning with the ``\texttt{jump}'' action in the cross-attention map. This not only verifies the soundness of the fine-grained text-motion correspondence modeling in MotionCLR, but also meets the theatrical analysis of motion-text ($\mathbf{Q}$-$\mathbf{K}$) similarity. This motivates us to manipulate the attention map to control when the action will be executed. The details will be introduced in~\cref{sec:applicaiton}.
We also visualize the self-attention map in~\cref{fig:study_attn}(C). As analyzed in~\cref{sec:clearattn}, the self-attention map evaluates the similarity between frames. As can be seen in~\cref{fig:study_attn}(C), the attention map highlights \textbf{nine} areas with similar motion patterns, indicating \textbf{three} jumping actions in total. Besides the temporal areas that the ``\texttt{jmup}'' word is activated are aligned with the jumping actions. The highlighted areas in the self-attention map are line areas, not square areas, indicating the taking-off, in-the-air, and landing actions of a jump with different detailed movement patterns. Due to the page limits, we leave more visualization for empirical evidence in~\cref{sec:moreemp}.

\section{Motion Editing Applications via Attention Manipulations}
\label{sec:applicaiton}

\begin{figure}[!t]
\centering
\captionsetup[subfigure]{aboveskip=0pt, belowskip=0pt}

\begin{subfigure}[b]{\linewidth}
    \centering
    \includegraphics[width=0.7\textwidth]{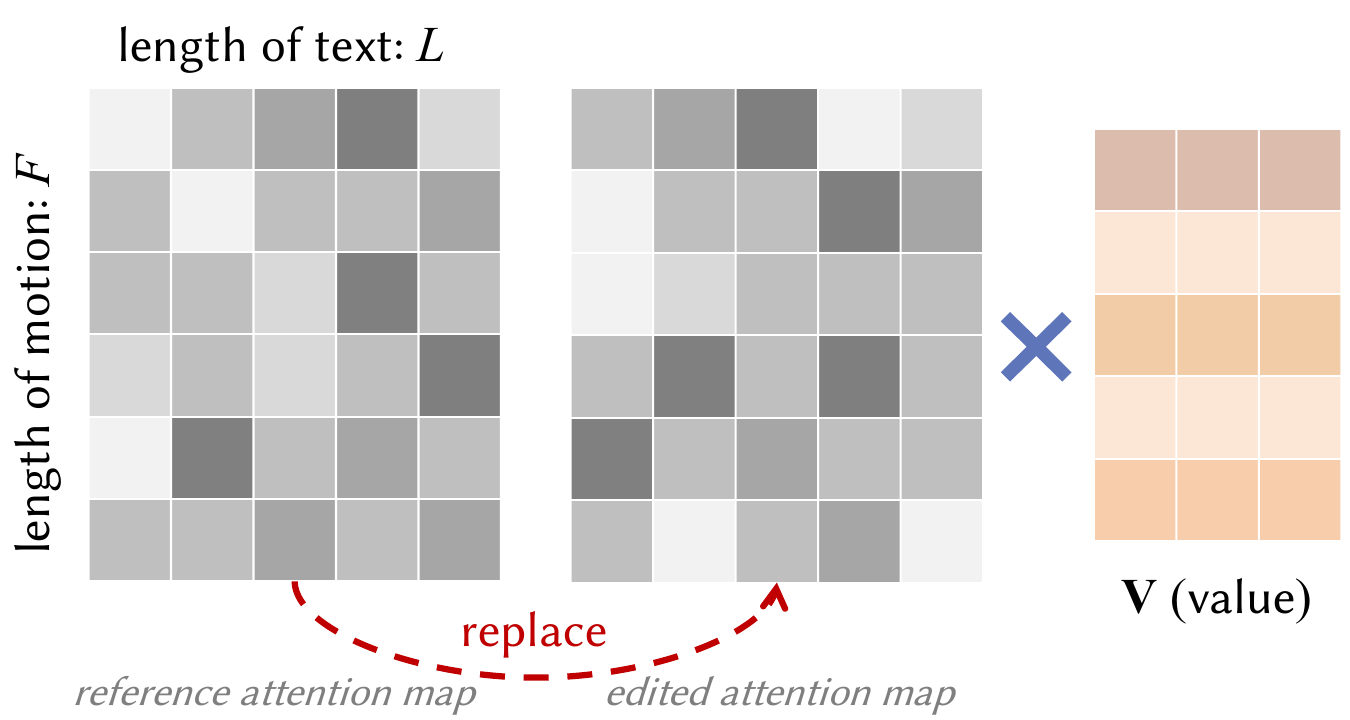}
    \caption{ In-place motion replacement via replacing cross-attention map (\cref{sec:replace_app}). The batch size of inference examples is two during the inference stage (reference and edited motion respectively). We replace the cross-attention map of the edited motion as the one of the reference motion. }
    \label{fig:replace}
\end{subfigure}
\begin{subfigure}[b]{\linewidth}
    \centering
    \begin{overpic}[width=0.75\textwidth]{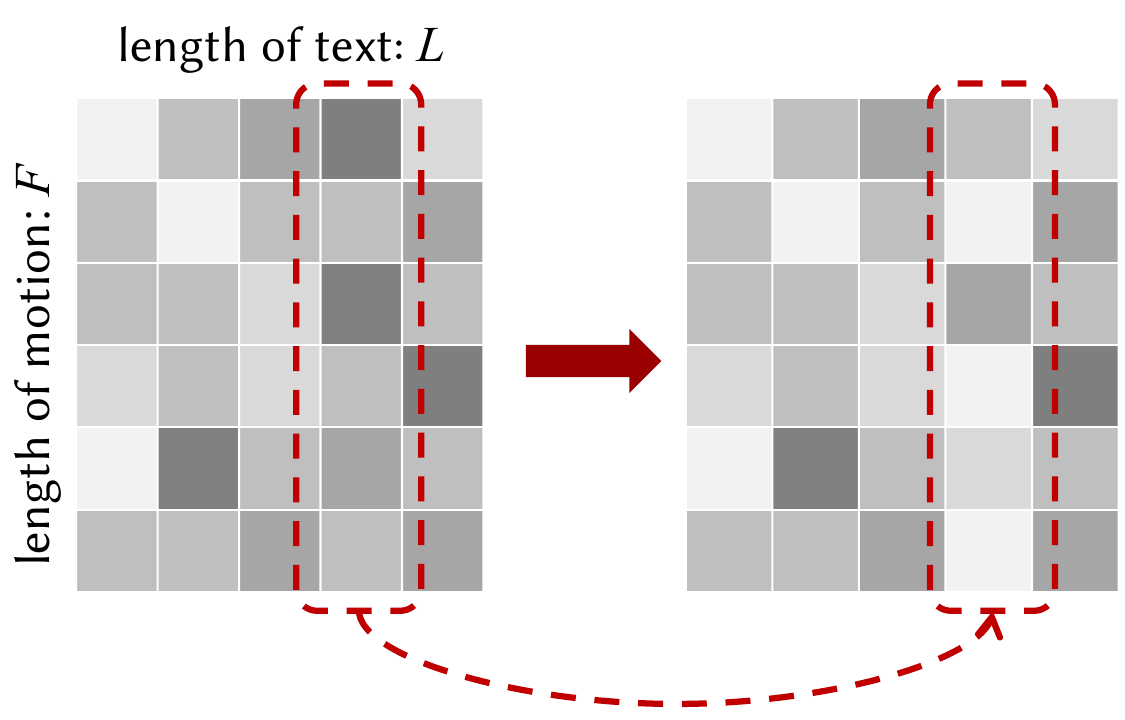}
        \put(47, 5){$\times (1+\alpha)$}
    \end{overpic}

    \caption{Motion (de-)emphasis via in/de-creasing cross-attention value. We can emphasize or de-emphasize a word in the prompt by adjusting the weight of the word. $\alpha \textless 0$ and $\alpha \textgreater 0$ mean emphasis and de-emphasis respectively.}
    \label{fig:emph}
\end{subfigure}
\begin{subfigure}[b]{\linewidth}
    \centering
    \includegraphics[width=0.75\textwidth]{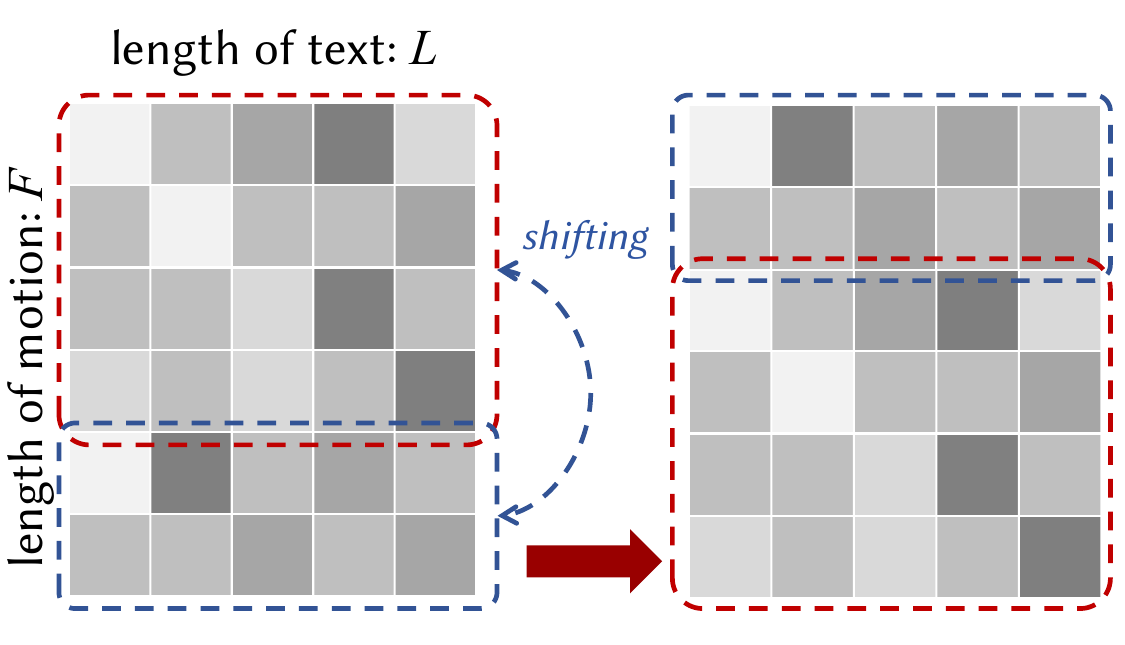}
    \vspace{-0.3em}
    \caption{Motion sequence shifting via shifting self-attention map. We adjust the sequence order of the motion sequence via shifting the attention map along the temporal axis.}
    \label{fig:shift}
\end{subfigure}
\vspace{-2em}

\caption{{\bf Diagram of motion editing via manipulating attention maps.} }
\vspace{-1em}
\label{fig:edit}
\end{figure}

Analysis in~\cref{sec:clearattn} has revealed the roles that attention mechanisms play in MotionCLR. In this section, we will show versatile downstream tasks of MotionCLR via manipulating attention maps.

\subsection{In-place Motion Replacement} 

\label{sec:replace_app}
In real scenarios, we would like to edit some local motion contents of the generated result. The key challenge is to keep the original composition and replace it with the new motion contents. Assuming we generate a reference motion at first, we would like to replace one action in the reference motion with another in place. Therefore, the batch size of inference examples is two during the inference stage, where the first is the reference motion and the other is the edited motion. As discussed in~\cref{sec:clearattn}, the cross-attention map determines when an action happens. Motivated by this, we replace the cross-attention map of the edited motion as the one of the reference motion. As shown in~\cref{fig:replace}, we use the replaced attention map to multiply the value matrix (text features) to obtain the output. As a result, we will obtain the motion of edited semantics with referenced temporal composition.

\subsection{Motion Emphasis and De-emphasis} 

In the text-driven motion generation framework, the process is driven by the input text. As discussed in~\cref{sec:clearattn}, the verb of the action will be significantly activated in the cross-attention map when the action is executed. As shown in~\cref{fig:emph}, if we increase/decrease the attention value of a verb in the cross-attention map, the corresponding action will be emphasized/de-emphasized, which can be implemented by multiplication ($\mathbf{A}_{:, i} \leftarrow \mathbf{A}_{:, i} \times (1+\alpha)$). Here, positive and negative values of $\alpha$ represent the motion emphasis and de-emphasis, respectively. Besides, this method can also be extended to the \textbf{grounded motion generation} application, which will be introduced in~\cref{sec:failcase}.

\subsection{Motion Erasing} 

Motion erasing is a special case of motion de-emphasis. We treat it as a special case of motion de-emphasis. When the decreased (de-emphasized) cross-attention value of an action is small enough, the corresponding action will be erased. The difference with motion de-emphasis is that the motion erasing is applied in a sub-temporal area specified by users in the whole sequence, and the de-emphasis is applied in the whole sequence. \\

\noindent Despite these semantic editing applications, MotionCLR also supports other editing tasks by manipulating self-attention.

\subsection{Motion Sequence Shifting} 

It is obvious that the generated motion is a combination of different actions along the time axis. Sometimes, users would like to shift a part of the motion along the time axis to satisfy the customized requirements. As shown in~\cref{fig:shift}, we can shift the motion sequentiality by shifting the self-attention map. As discussed in~\cref{sec:clearattn}, self-attention is only related to the motion feature without related to the semantic condition, which is our motivation on manipulating the self-attention map. Thanks to the denoising process, the final output sequence should be a natural and continuous sequence.

\subsection{Example-based Motion Generation} 

As defined by~\citet{weiyu23GenMM,motion_texture}, example-based motion generation aims to generate novel motions referring to an example motion. In MotionCLR system, this task is a special case of the motion sequence shifting. That is to say, we can shuffle the self-attention map along the temporal axis to obtain the diverse motions referring to the example. As analyzed in~\cref{sec:clearattn}, the ``value''s ($\mathbf{V}$) in self-attention means the texture of the motion. Therefore, shuffling the self-attention map along the temporal axis without manipulating ``value''s compromises the similar motion texture to the original one.

\begin{table*}[!t]
\centering
\setlength{\tabcolsep}{6pt}
\caption{\textbf{Comparison with different methods on the HumanML3D dataset.} The baselines include diffusion-based methods and state-of-the-art methods. The ``$^\dag$'' notation denotes the DPM-solver sampling inference design choice and ``$^*$'' is the DDIM sampling choice. As DPM-solver and DDIM present comparable performance, without specification, we set the DDIM sampling as our default choice. The comparison shows that MotionCLR is with comparable performance with state-of-the-art methods.}
    \vspace{-0.8em}
\resizebox{0.85\linewidth}{!}{
\begin{tabular}{lccccccc}
\toprule
\multirow{2}{*}{Methods} & \multicolumn{3}{c}{R-Precision$\uparrow$} & \multirow{2}{*}{FID$\downarrow$} & \multirow{2}{*}{MM-Dist$\downarrow$} & \multirow{2}{*}{Multi-Modality$\uparrow$} \\
 \cmidrule(lr){2-4}
 & Top 1 & Top 2 & Top 3 &  &  &  \\
\midrule
MDM~\citeyearpar{mdm}   & - & - & 0.611$^{\pm0.007}$ & 0.544$^{\pm0.044}$ & 5.566$^{\pm0.027}$ & \textbf{2.799$^{\pm0.072}$} \\
MLD~\citeyearpar{mld}  & 0.481$^{\pm0.003}$ & 0.673$^{\pm0.003}$ & 0.772$^{\pm0.002}$ & 0.473$^{\pm0.013}$ & 3.196$^{\pm0.010}$ & 2.413$^{\pm0.079}$ \\
MotionDiffuse~\citeyearpar{motiondiffuse} & 0.491$^{\pm0.001}$ & 0.681$^{\pm0.001}$ & 0.782$^{\pm0.001}$ & 0.630$^{\pm0.001}$ & 3.113$^{\pm0.001}$ & 1.553$^{\pm0.042}$ \\
ReMoDiffuse~\citeyearpar{remodiffuse}   & 0.510$^{\pm0.005}$ & 0.698$^{\pm0.006}$ & 0.795$^{\pm0.004}$ & 0.103$^{\pm0.004}$ & 2.974$^{\pm0.016}$ & 1.795$^{\pm0.043}$ \\
MoMask~\citeyearpar{momask} & 0.521$^{\pm0.002}$ & 0.713$^{\pm0.002}$ & 0.807$^{\pm0.002}$ & \textbf{0.045$^{\pm0.002}$} & 2.958$^{\pm0.008}$ & 1.241$^{\pm0.040}$ \\
\hline
MotionCLR$^\dag$  & {0.542}$^{\pm0.001}$ & \textbf{0.733}$^{\pm0.002}$ & {0.827}$^{\pm0.003}$ & 0.099$^{\pm0.003}$ & {2.981}$^{\pm0.011}$ & 2.145$^{\pm0.043}$ \\ 
MotionCLR$^*$ & \textbf{0.544}$^{\pm0.001}$ & 0.732$^{\pm0.001}$ & \textbf{0.831}$^{\pm0.002}$ & 0.269$^{\pm0.001}$ & \textbf{2.806}$^{\pm0.014}$ & 1.985$^{\pm0.044}$  \\
\bottomrule
\end{tabular}
}
\vspace{-0.8em}
\label{tab:humanml3d_main}
\end{table*}

\subsection{Motion Style Transfer} 

As discussed in the technical details of the self-attention mechanism, the values mainly contribute to the contents of motion and the attention map determines the selected indices of motion frames. Following~\citet{momo}, when synthesizing two motion sequences ($\mathbf{M}_1$ and $\mathbf{M}_2$ respectively), we only need to replace $\mathbf{Q}$s in $\mathbf{M}_2$ with that in $\mathbf{M}_1$ to achieve the style of $\mathbf{M}_2$ into $\mathbf{M}_1$'s. Specifically, queries ($\mathbf{Q}$s) in $\mathbf{M}_2$ determine which motion feature in $\mathbf{M}_2$ is the most similar to that in $\mathbf{M}_1$ at each timestep. Accordingly, these most similar motion features are selected to compose the edited motion. Besides, the edited motion is with the motion content of $\mathbf{M}_2$ while imitating the motion style of $\mathbf{M}_1$.


\section{Experiments}

\subsection{Implementation Details}

\label{sec:app_implementation}

The MotionCLR model is trained on the HumanML3D dataset with one NVIDIA A-100 GPU based on PyTorch~\citep{torch}. The latent dimension of the motion embedding is $512$. We take the CLIP-ViT-B model to encoder text as word-level embeddings. The training process utilizes a batch size of 64, with a learning rate initialized at $2e-4$ and decaying at a rate of $0.9$ every $5,000$ steps. Additionally, a weight decay of $1e-2$ is employed to regularize the model parameters. For the diffusion process, the model is trained over $1,000$ diffusion steps. We incorporate a probability of $0.1$ for condition masking to facilitate classifier-free guidance learning. During training, dropout is set at $0.1$ to prevent overfitting, and all networks in the architecture follow an $8$-layer Transformer design. For motion representation, we follow the setting in~\citet{humanml3d}. 

In the inference stage, all steps of the denoising sampling are set as $10$ consistently. For the motion erasing application, we set the erasing weight as 0.1 by default. MotionCLR supports both DDIM-sampling~\citep{ddim} and DPM-soler-sampling~\citep{dpm1} methods, with $1,000$ as full diffusion steps. For the in-placement motion replacement and the motion style transfer application, as the motion semantics mainly depend on the initial denoising steps, we set the manipulating steps until 5 as default. For the example-based motion generation, the minimum manipulating time of a motion zone is $1$s (\ie chunk size=20 for the 20 FPS setting). At each step, all attention maps at all layers will be manipulated at $1 \sim 9$ denoising timestep. For editing the ground truth motions, we directly use the DDIM inversion~\cite{ddim} to edit the motion. Users can adjust the parameters freely to achieve interactive motion generation and editing (more details of user interface in~\cref{sec:addon}).

\subsection{Motion Generation Evaluation for MotionCLR} 

The implementation details of the MotionCLR are in~\cref{sec:app_implementation}. 
We first evaluate the generation performance of the MotionCLR on HumanML3D~\cite{humanml3d}. We extend the evaluation metrics of previous works~\citep{humanml3d}, including FID, R-Precision, MM-Dist, and, Muiti-Modality. 
The results are shown in~\cref{tab:humanml3d_main}, indicating a comparable performance with the state-of-the-art method. Especially, our result has a higher text-motion alignment over baselines, owing to the explicit fine-grained cross-modality modeling. As shown in~\cref{tab:humanml3d_main}, both DDIM and DPM-solver sampling work consistently well compared with baselines. We leave more visualization and qualitative results in~\cref{sec:main_vis}.

\subsection{Constructing Evaluation Set for Motion Editing}

To evaluate the semantic correspondence between cross-attention and the generated motion, we construct an evaluation set to verify the observation qualitatively. We label verbs in HumanML3D texts in each sentence. Here, only a single verb in the sentence will be labeled and the sentences without any verbs will be filtered out. As a result, we construct 19,492 texts for evaluation, namely the HVerb test set. For the convenience of validating the in-placement replacement task, we also assign a new verb as the replacing verb of the original verb in the HVerb test set. To additionally test the explainability of the model in-the-wild, we construct 200 text prompts with verb labels via GPT-4o~\cite{gpt4}, namely HVerb-wild, whose results are checked by researchers.

\begin{table}[!t]
\centering
\caption{\textbf{IoU (\%) metrics on different settings.} High coherence among E1, E2, and E3 shows the fine-grained text-motion modeling in cross attention. }
\resizebox{0.65\linewidth}{!}{
\begin{tabular}{cccccc}
\toprule
Experiment ID & E1 & E2 & E3 & E4 & E5 \\ \hline
HVerb &\cellcolor{gray!30} 74.3\% & \cellcolor{gray!30} 72.9\% \cellcolor{gray!30} & \cellcolor{gray!30} 77.8\% & \cellcolor{gray!60} 17.5\% &\cellcolor{gray!60}  17.0\% \\
\hline 
\hline
Experiment ID & E1 & E2 & E3 & E4 & E5 \\ \hline
HVerb-wild &\cellcolor{gray!30} 73.5\% & \cellcolor{gray!30} 71.4\% \cellcolor{gray!30} & \cellcolor{gray!30} 74.5\% & \cellcolor{gray!60} 18.5\% &\cellcolor{gray!60}  19.4\% \\
\bottomrule
\end{tabular}
}
\vspace{-0.8em}
\label{tab:beforetest}
\end{table}

\subsection{Quantitative Results for Correspondence between Cross Attention and Generated Motion}
\label{sec:evidence}

Before evaluating the motion editing result, we initially validate the word-motion correspondence in the cross-attention map of MotionCLR. We set up 5 groups of experiments for comparison. (1) \textbf{E1}: IoU between cross attention \& root velocity. We treat the value in the cross-attention map larger than 80\% of the maximum value as an activated action. Accordingly, we treat the activated and unactivated parts as 1 and 0 respectively. Similarly, we also apply this to the root velocity to get another activation map. The IoU metric between these two activation maps means the coherence between the attention activation and the kinematic features. (2) \textbf{E2}: IoU between the cross attention \& moment retrieval value. We use the moment retrieval function of TMR to calculate the verb-motion similarity in 20-frame windows, obtaining the similarity between sub-motion and the verb along the whole motion sequence. Accordingly, we can also calculate a metric between attention and the moment retrieval value (also set 60\% as the threshold). (3) \textbf{E3}: IoU between the root velocity \& moment retrieval value. This setting is similar to that in \textbf{E2} and \textbf{E3}. Both root velocity and moment retrieval value are explicit kinematic metrics used to describe action execution. The motivation to set \textbf{E3} is for setting a comparison group to verify that the cross-attention correspondence is coherent with these two metrics. (4): \textbf{E4}: neg. cross-attention \& root velocity. We modify the setting in \textbf{E1} and replace the cross-attention map with a randomly sampled cross-attention map in the test in calculating IoU with the root velocity. (5): \textbf{E5}: neg. cross-attention \& moment-retrieval-value. Similarly, we modify the setting in \textbf{E2} and replace the cross-attention map with a randomly sampled cross-attention map in the test in calculating IoU with moment-retrieval-value.

As can be seen in~\cref{tab:beforetest}, compared with \textbf{E3} (both explicit action indicators: speed, moment retrieval of TMR), the values in \textbf{E1}/\textbf{E2} are similar to those in \textbf{E3}. Therefore, the cross-attention activation is well aligned with motion execution. Besides, as shown in random sampling comparison groups (\textbf{E1} \textit{v.s.} \textbf{E4}/\textbf{E5}), the cross attention is aligned with the action execution in the motion in the generation process of each motion. The result proves the word-motion correspondence in MotionCLR.

\subsection{Evaluation on Inference-only Motion Editing}

\begin{table}[!t]
\setlength{\tabcolsep}{3pt}
\caption{\textbf{In-place motion replacement.} $^*$The ``R'' and ``C'' settings represent ``Replace A as B'' and ``Change A with B'' prompts of MotionFix. The \textcolor{lightgray}{light gray} text is the comparison group, denoting the upper bound of the performance. The \textbf{bold} numbers are the best results excepting the comparison group. The significant metric margin over baselines shows the good performance of the method, even some methods requiring specific training.}
\resizebox{\linewidth}{!}{
\begin{tabular}{l|cccc}
\toprule
HVerb & \begin{tabular}[c]{@{}c@{}}training- \\ free\end{tabular} & \begin{tabular}[c]{@{}c@{}}align with \\ original text\end{tabular} (\%) $\downarrow$ & \begin{tabular}[c]{@{}c@{}}align with \\ edited text\end{tabular} (\%) $\uparrow$& \begin{tabular}[c]{@{}c@{}}unedited part \\ preserving\end{tabular} (mm) $\downarrow$\\ 
\hline
w/o editing         & -          & 76.205                   & 64.687                 & \textcolor{lightgray}{0.0}                      \\ 
editing text only        & \Checkmark          & \textcolor{lightgray}{62.324}                   & \textcolor{lightgray}{68.368}                 & 201.5                    \\
MotionFix (``R'')     & \ding{55}      & 74.526                   & 63.542                 & 130.2                    \\
MotionFix (``C'')     &  \ding{55}       & 75.112                   & 63.780                 & 142.1                    \\
Ours       & \Checkmark                 & \textbf{63.324}                   & \textbf{68.125}                 & \textbf{57.9} \\
\hline 
\end{tabular}
}
\resizebox{\linewidth}{!}{
\begin{tabular}{l|cccc}
\hline
HVerb-wild  & \begin{tabular}[c]{@{}c@{}}training- \\ free\end{tabular} & \begin{tabular}[c]{@{}c@{}}align with \\ original text\end{tabular} (\%) $\downarrow$ & \begin{tabular}[c]{@{}c@{}}align with \\ edited text\end{tabular} (\%) $\uparrow$& \begin{tabular}[c]{@{}c@{}}unedited part \\ preserving\end{tabular} (mm) $\downarrow$\\ 
\hline
w/o editing      & -                & 73.678                   & 59.229                 & \textcolor{lightgray}{0.0}                      \\ 
editing text only        & \Checkmark                & \textcolor{lightgray}{56.235}                   & \textcolor{lightgray}{66.701}                 & 235.0                    \\
MotionFix (``R'')    &  \ding{55}      & 70.125                   & 58.560                 & 151.1                    \\
MotionFix (``C'')     & \ding{55}     & 71.588                   & 60.009                 & 149.0                    \\
Ours           & \Checkmark           & \textbf{58.124}                   & \textbf{65.976}                 & \textbf{59.8} \\
\bottomrule
\end{tabular}
}
\label{tab:replacenew}
\end{table}

\begin{figure*}[!t]
    \centering
    \begin{subfigure}[b]{0.38\textwidth}
        \centering
        \includegraphics[width=\textwidth]{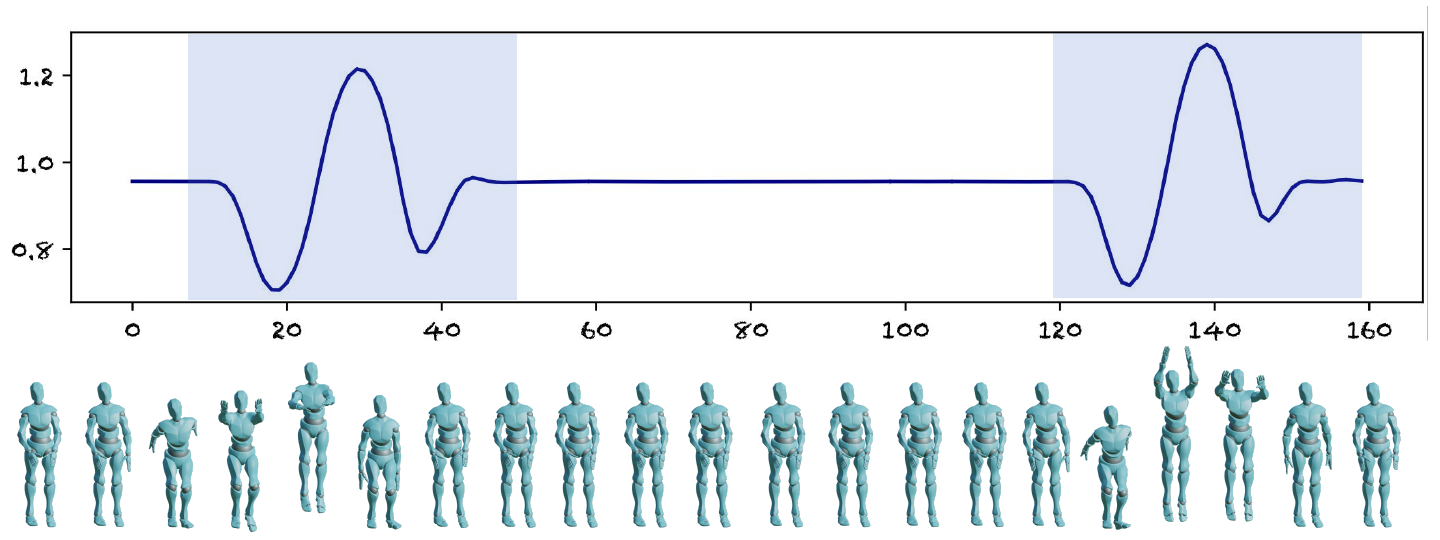}
        \caption{Root \textbf{height} visualization and the motion of ``\texttt{a man jumps.}'' before applying editing.}
    \label{fig:replace-result-a}
    \end{subfigure}
    \hspace{1em}
    \begin{subfigure}[b]{0.57\textwidth}
        \centering
        \includegraphics[width=\textwidth]{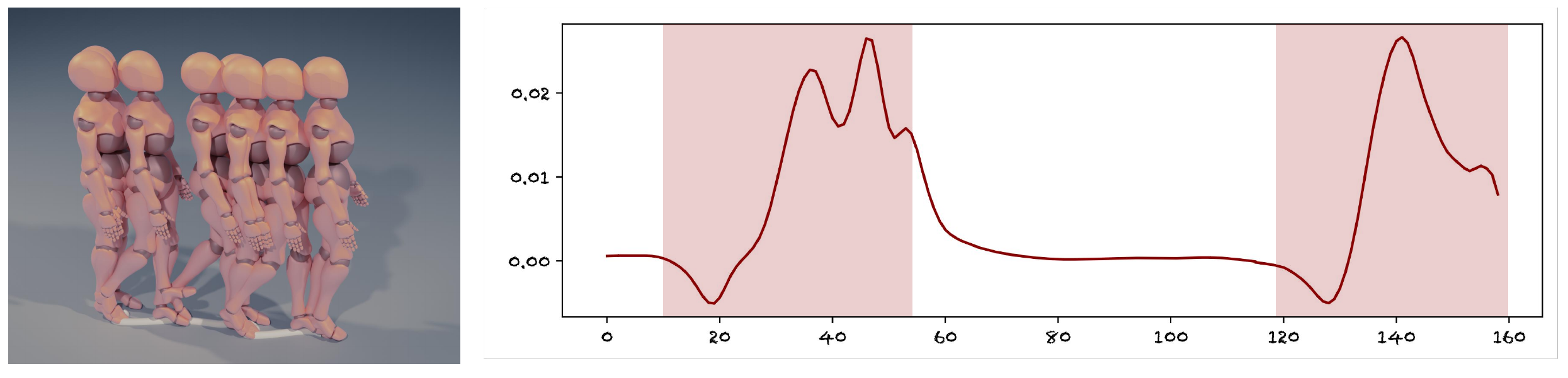}
        \caption{Motion visualization and the root \textbf{velocity} of editing the prompt ``\texttt{a man \textcolor{myblue}{jumps}.}''$\rightarrow$``\texttt{a man \textcolor{myred}{walks}.}''.}
    \label{fig:replace-result-b}
    \end{subfigure}
    \begin{subfigure}[b]{0.95\textwidth}
        \centering
        \includegraphics[width=0.9\textwidth]{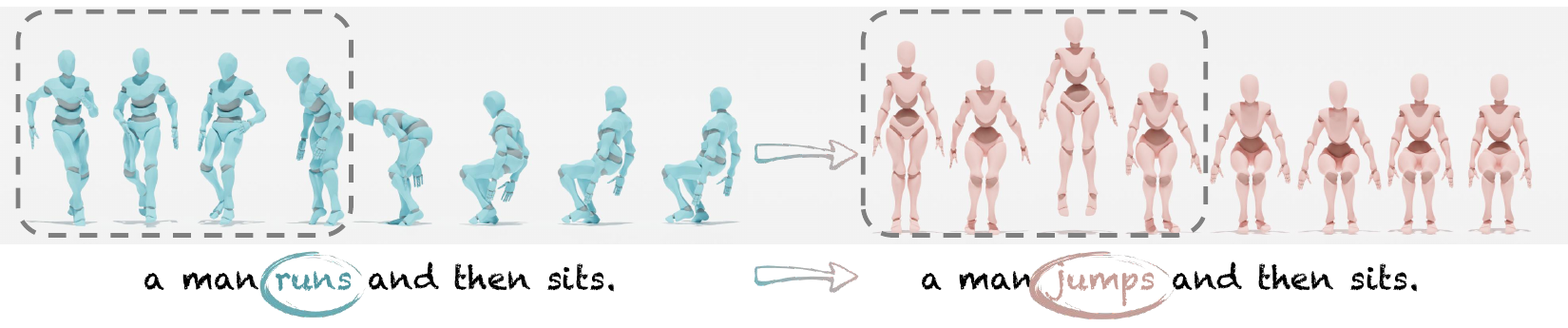}
        \vspace{-0.7em}
        \caption{Motion in-place replacement results of a motion including multiple actions. In this case, we replace the ``\texttt{\textcolor{myblue}{jumps}}'' in ``\texttt{a man \textcolor{myblue}{jumps}.}'' as ``\texttt{\textcolor{myred}{walks}}'' without editing the sitting action. The edited motion and the vanilla motion share the same temporal area of the action execution.}
    \label{fig:replace-result-c}
    \end{subfigure}
    \vspace{-0.8em}
    \caption{\textbf{In-place motion replacement.} (a) and (b) are a pair of motions before and after editing. (c) is a comparison of original and edited motions.}
    \label{fig:replace-result}
    \vspace{-0.5em}
\end{figure*}

\subsubsection{In-place motion replacement.} Different from na\"ive replacing prompts for motion replacement, in-place motion replacement not only needs to replace the original motion at the semantic level, but also needs to replace motions at the exact temporal place. \cref{fig:replace-result-a} and \cref{fig:replace-result-b} show the root height trajectory and the root horizontal velocity, respectively. In this case, the edited and original motion share the same time zone to execute the action. Besides, the edited motion is semantically aligned with the word ``\texttt{walk}''. \cref{fig:replace-result-c} also shows results of replacing ``\texttt{runs}'' as``\texttt{jumps}'' without changing the sitting action. These quantitative results show the effectiveness of the in-place motion replacement application.

\begin{figure}
    \centering
    \includegraphics[width=\linewidth]{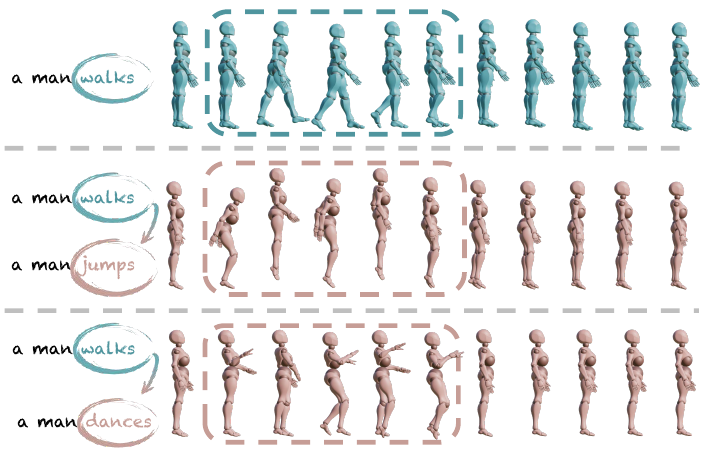}
    \vspace{-1em}
    \caption{{\bf Results of editing multiple semantics from the same motion.} All edited motions with different semantics share the same temporal location as the original motion.}
    \vspace{-1em}
    \label{fig:replace-multi}
\end{figure}

For the in-place motion replacement application, we additionally provide replacement results in~\cref{fig:replace-multi}. As shown in~\cref{fig:replace-multi}, MotionCLR changes the original ``walk'' motion into ``jumps'' and ``dances'' at the corresponding place, serving as a diverse semantic editing function. All edited motion with different semantics share the same action temporal location with the original motion. This function shows the robustness of the method for diverse editing requirements.

We also test our method on the HVerb and the HVerb-wild test sets quantitatively, which are shown in~\cref{tab:replacenew}. We take the latest motion editing method, MotionFix, as a baseline, whose prompts are set as `Replace A as B'' or ``Change A with B''. For example, the prompt can be ``Replace jump as walk''. Two examples are set to enhance prompt diversity of the baseline. As the editing process of COMO~\cite{como} and \citet{goel2024iterative} are not fully open-sourced before writing this paper, we do not take these as our baselines. For the application of in-place motion replacement, the editing goal comes up with two aspects. (1) \textit{The semantics of the edited motion should be aligned with the replaced text.} Here, we take the TMR~\cite{tmr} similarity (0\%$\sim$100\%) to evaluate the text-motion similarity. As shown in~\cref{tab:replacenew}, the former two columns indicate that the edited motion is more semantically aligned with the edited text, and less aligned with the original text. (2) \textit{The unedited part of the motion should be reserved.} We use the TMR moment retrieval function and the annotated verb to filter out the editing area of the motion, similar to~\cref{sec:evidence}. For the filtered motion part, we take the MPJPE (mm) metric to evaluate the motion-preserving ability of the unedited motion. As can be seen in the last column of~\cref{tab:replacenew}, MotionCLR shows a stable motion-preserving ability on unedited areas. As the dataset construction process of MotionFix is based on a similar motion retrieval process, the semantic changes in the editing process are unsatisfactory.

\begin{figure*}[!t]
    \centering
    \setlength{\abovecaptionskip}{0pt}
    \setlength{\belowcaptionskip}{0pt}
    \begin{subfigure}[b]{0.49\textwidth}
    \includegraphics[width=\linewidth]{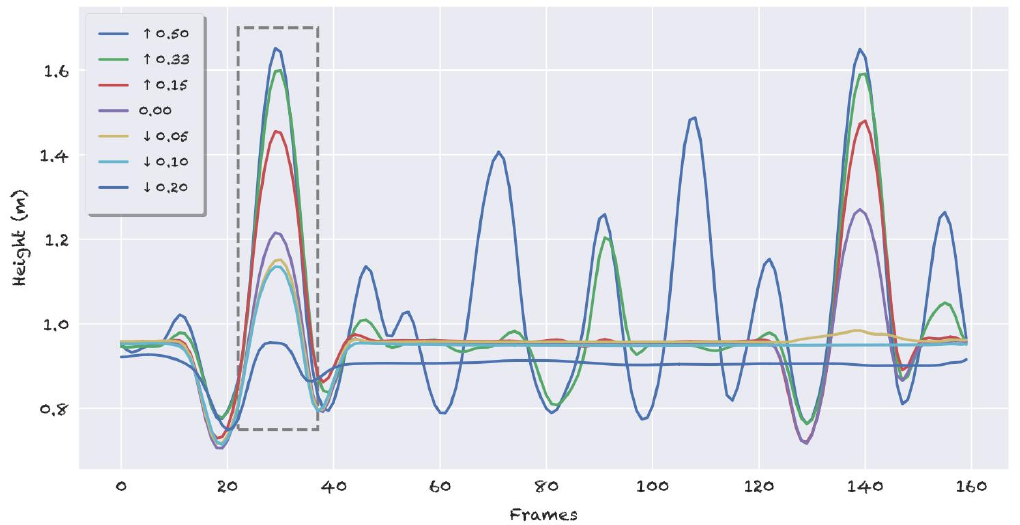}
    \caption{The height of the character's root. The highlighted area is obvious when comparing different weights.}
    \label{fig:emphasis-1}
    \end{subfigure}
    \hfill
    \begin{subfigure}[b]{0.46\textwidth}
    \centering
    \includegraphics[width=\linewidth]{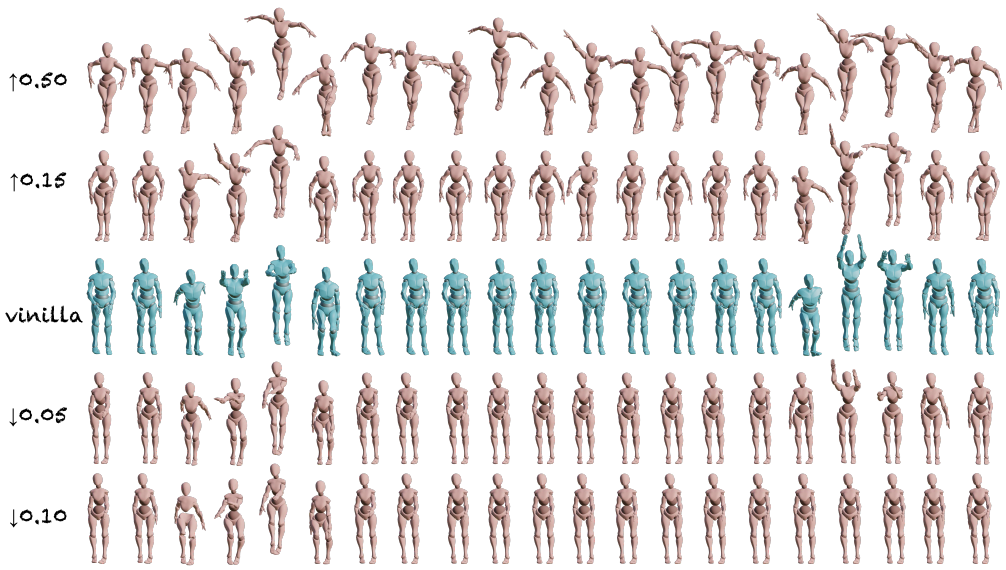}
    \caption{Motion visualization of the edited motions on different (de-)emphasis weight settings.}
    \label{fig:emphasis-2}
    \end{subfigure}
    \captionsetup{font=small} .
    \caption{\textbf{Motion (de-)emphasis.} Different weights of ``\texttt{jump}'' ($\uparrow$ or $\downarrow$) in ``\texttt{a man jumps.}''.}
    \label{fig:emphasis}
    \vspace{-0.4em}
\end{figure*}

\begin{figure*}[!t]
    \centering
    \captionsetup[subfigure]{aboveskip=0pt, belowskip=0pt, font=small}
    \captionsetup[figure]{aboveskip=0pt, belowskip=0pt}
    \begin{subfigure}[b]{0.97\textwidth}
    \includegraphics[width=\linewidth]{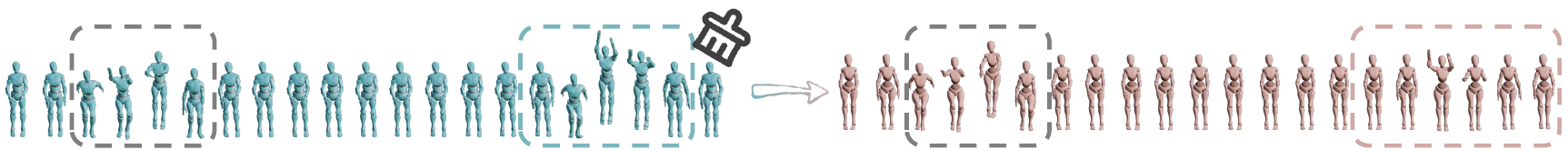}
    \caption{``\texttt{a man jumps.}'' case. The second jumping action is erased.}
    \label{fig:erase-1}
    \end{subfigure}
    \begin{subfigure}[b]{0.97\textwidth}
    \centering
    \includegraphics[width=\linewidth]{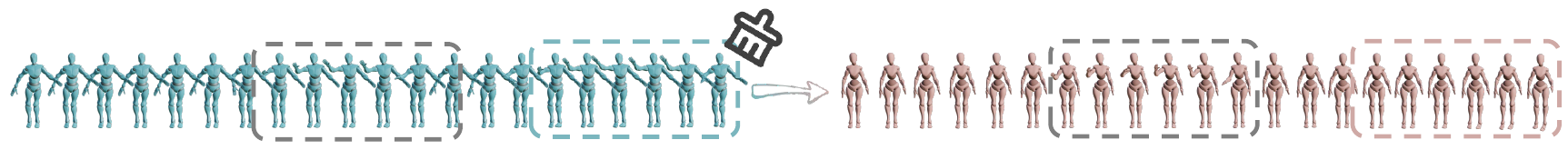}
    \caption{``\texttt{waving hand}'' case. The final 1/3 waving action is removed.}
    \label{fig:erase-2}
    \end{subfigure}
    \captionsetup{font=small}
    \caption{\textbf{Motion erasing results.} Case study of ``\texttt{a man jumps.}'' and ``\texttt{waving hand}'' cases.}
    \label{fig:erase}
    \vspace{-0.3em}
\end{figure*}

\begin{figure}[!t]
    \centering
    \includegraphics[width=0.8\linewidth]{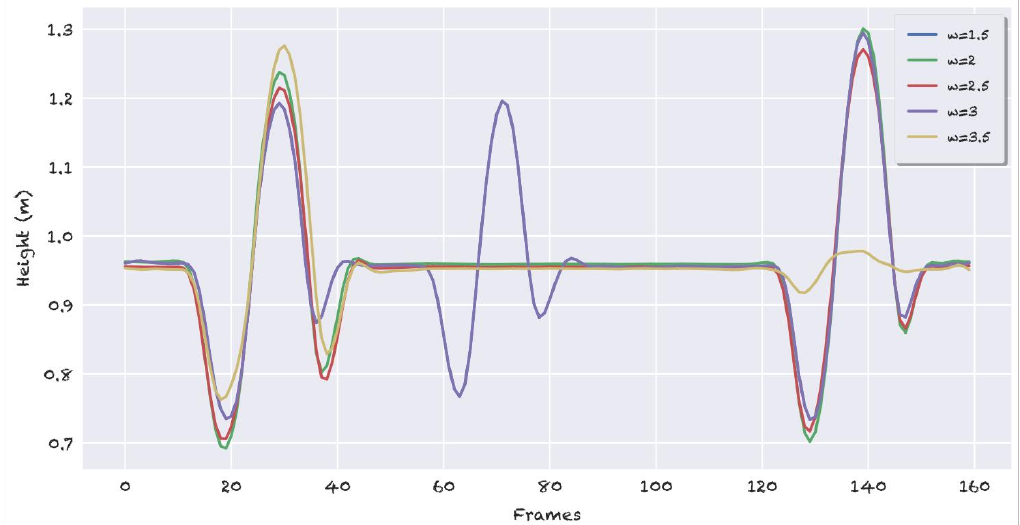}
    \caption{\textbf{The effect of varying $w$ in classifier-free guidance on generated motions.} While changing $w$ influences the general alignment between the text ``\texttt{a man jumps.}'' and the generated motion, it does not provide precise control over finer details like jump height and frequency. }
    \vspace{-1em}
    \label{fig:cfg_ablation}
\end{figure}

\subsubsection{Motion (de-)emphasis and motion erasing.} 
We mainly provide the visualization results of motion (de-)emphasis in~\cref{fig:emphasis}. As shown in~\cref{fig:emphasis}, the edited results are aligned with the manipulated attention weights. Especially, as can be seen, in~\cref{fig:emphasis-1}, the height of the ``\texttt{jump}'' action is accurately controlled by the cross-attention weight of the word ``\texttt{jump}''. For an extremely large adjusting weight, \eg $\uparrow$1.0, the times of the jumping action also increase. This is because the low-activated timesteps of the originally generated motion might have a larger cross-attention value to activate the ``\texttt{jump}'' action.

Additionally, we would like to discuss the difference between reweighting the cross-attention map and adjusting classifier-free guidance weights. For intuitive understanding, the classifier-free guidance weight also controls the semantics of the motion with the input text. \textbf{However}, as the classifier-free guidance mainly works for the semantic alignment between text and motion, it cannot control the weight of each word. We take the sentence ``\texttt{a man jumps.}'' as an example for a fair comparison, which is the case used in the main text (suggested to refer to \cref{fig:emphasis-1} for comparison). As shown in \cref{fig:cfg_ablation}, the generated motions with different $w$ values illustrate that $w$ \textbf{cannot} influences both the height and frequency of the jump. Nevertheless, the classifier-free guidance is limited in its ability to control more detailed aspects, such as the exact height and number of actions. Therefore, while $w$ improves text-motion alignment, it cannot achieve fine-grained adjustments. As there is no benchmark for such an application, we quantitatively evaluate this in the user study part (\cref{sec:user_study}).

As motion erasing is a special case of motion de-emphasis, we do not provide more quantitative on this application. We provide some visualization results in~\cref{fig:erase}. As can be seen in~\cref{fig:erase-1}, the second jumping action is erased. Besides, the ``waving hand'' case shown in~\cref{fig:erase-2} shows that the final 1/3 waving action is also removed. 


\definecolor{myorange1}{RGB}{252, 233, 217} 
\definecolor{myorange2}{RGB}{205, 107, 0} 
\definecolor{mygreen1}{RGB}{205, 233, 185} 
\definecolor{mygreen2}{RGB}{96, 153, 43}


\begin{figure*}[!t]
    \centering
    \begin{subfigure}[b]{0.48\textwidth}
        \centering
        \includegraphics[width=0.95\textwidth]{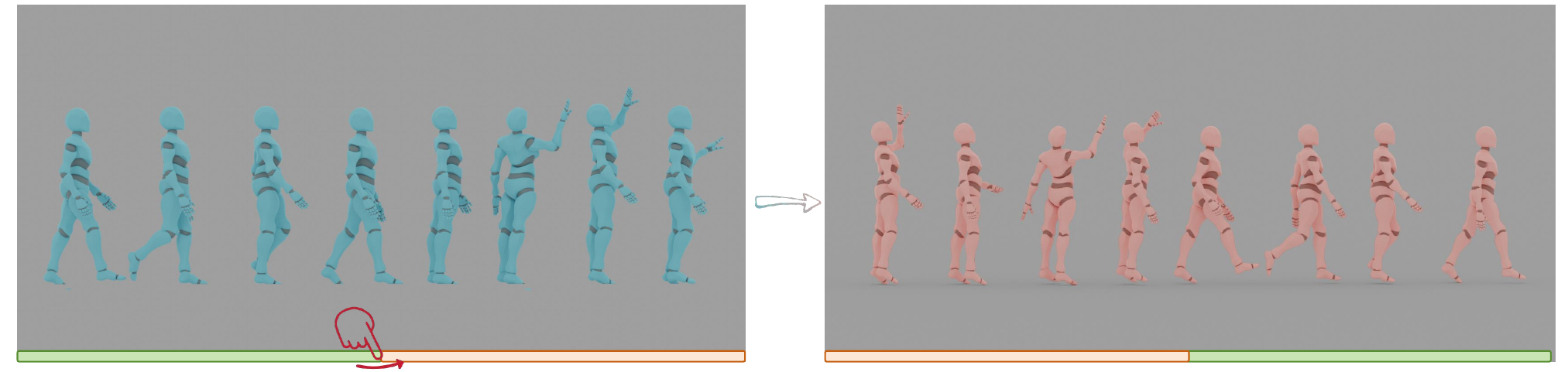}
        \caption{Prompt: ``\texttt{a person walks straight and then waves.}'' Original (blue) \vds shifted (red).}
    \label{fig:shift-result-a}
    \end{subfigure}
    \hfill
    \begin{subfigure}[b]{0.48\textwidth}
        \centering
        \includegraphics[width=0.95\textwidth]{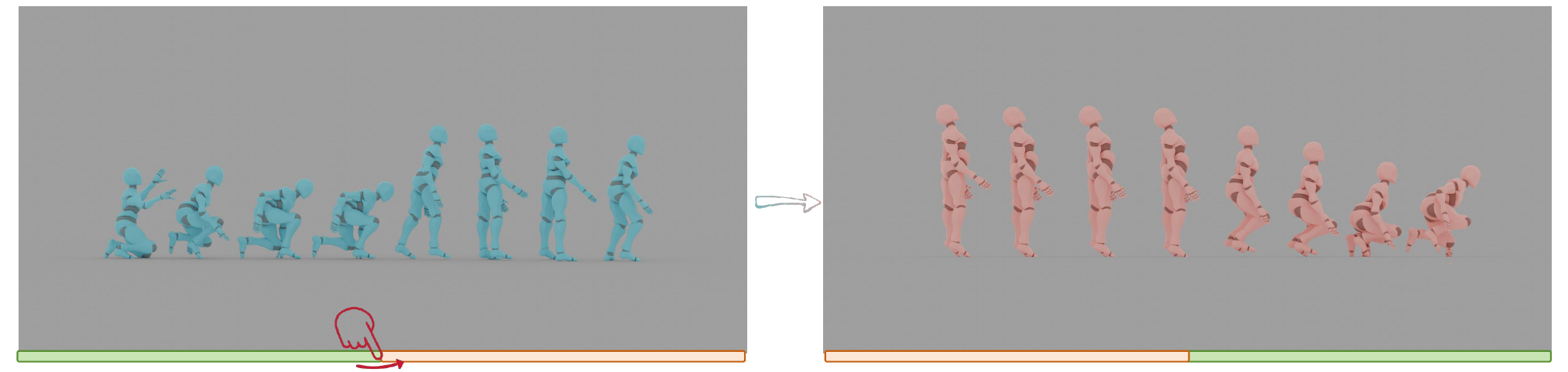}
        \caption{Prompt: ``\texttt{a man gets up from the ground.}'' Original (blue) \vds shifted (red).}
    \label{fig:shift-result-b}
    \end{subfigure}
    \vspace{-0.8em}
    \caption{\small \textbf{Comparison between original and shifted motions.} Time bars are shown in different colors. (a) The original figure raises hands after the walking action. The shifted one has the opposite sequentiality. (b) The squatting action is shifted to the end of the sequence, and the standing-by action is shifted to the beginning.}
    \label{fig:shift-result}
\end{figure*}

\subsubsection{Motion sequence shifting.} Here, we provide some comparisons between the original motion and the edited one. In~\cref{fig:shift-result}, we take ``\tikz[baseline]{\node[draw=myorange2, fill=myorange1, rounded corners=0.05cm, minimum width=0.8cm, minimum height=0.15cm, anchor=base, yshift=0.1cm] {}; }'' and ``\tikz[baseline]{\node[draw=mygreen2, fill=mygreen1, rounded corners=0.05cm, minimum width=0.8cm, minimum height=0.15cm, anchor=base, yshift=0.1cm] {}; }'' to represent different time bars, whose orders represent the sequentially. As can be seen in~\cref{fig:shift-result-a}, the execution of waving hands is shifted to the beginning of the motion. Besides, as shown in~\cref{fig:shift-result-b}, the squatting action has been moved to the end of the motion. These results show that the editing of the self-attention map sequentiality has an explicit correspondence with the editing motion sequentially. More results are in~\cref{sec:app_shift}.

\begin{figure*}[!t]
    \centering
    \captionsetup[subfigure]{aboveskip=0pt, belowskip=0pt}
    \begin{subfigure}[b]{0.65\textwidth}
        \centering
        \includegraphics[width=\textwidth]{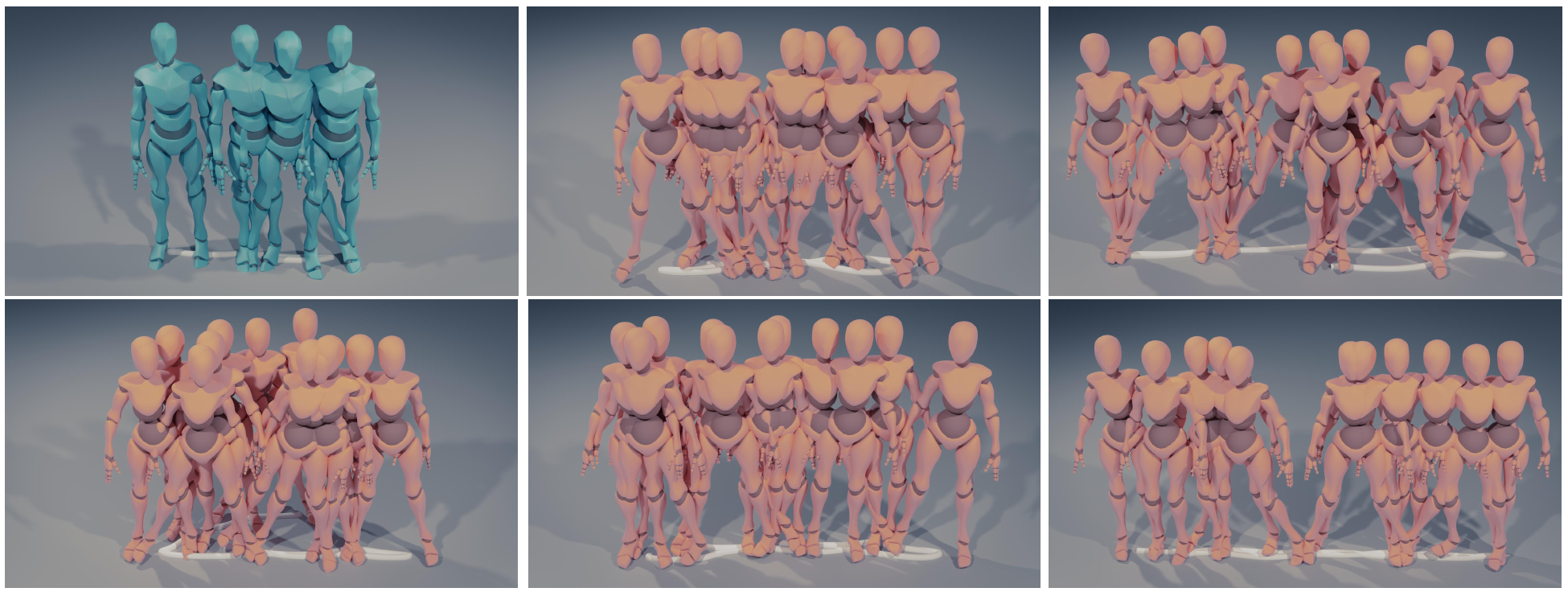}
        \caption{Examples (blue) and generated (red) motions.}
    \label{fig:example-based-result-a}
    \end{subfigure}
    \begin{subfigure}[b]{0.31\textwidth}
        \centering
        \includegraphics[width=0.88\textwidth]{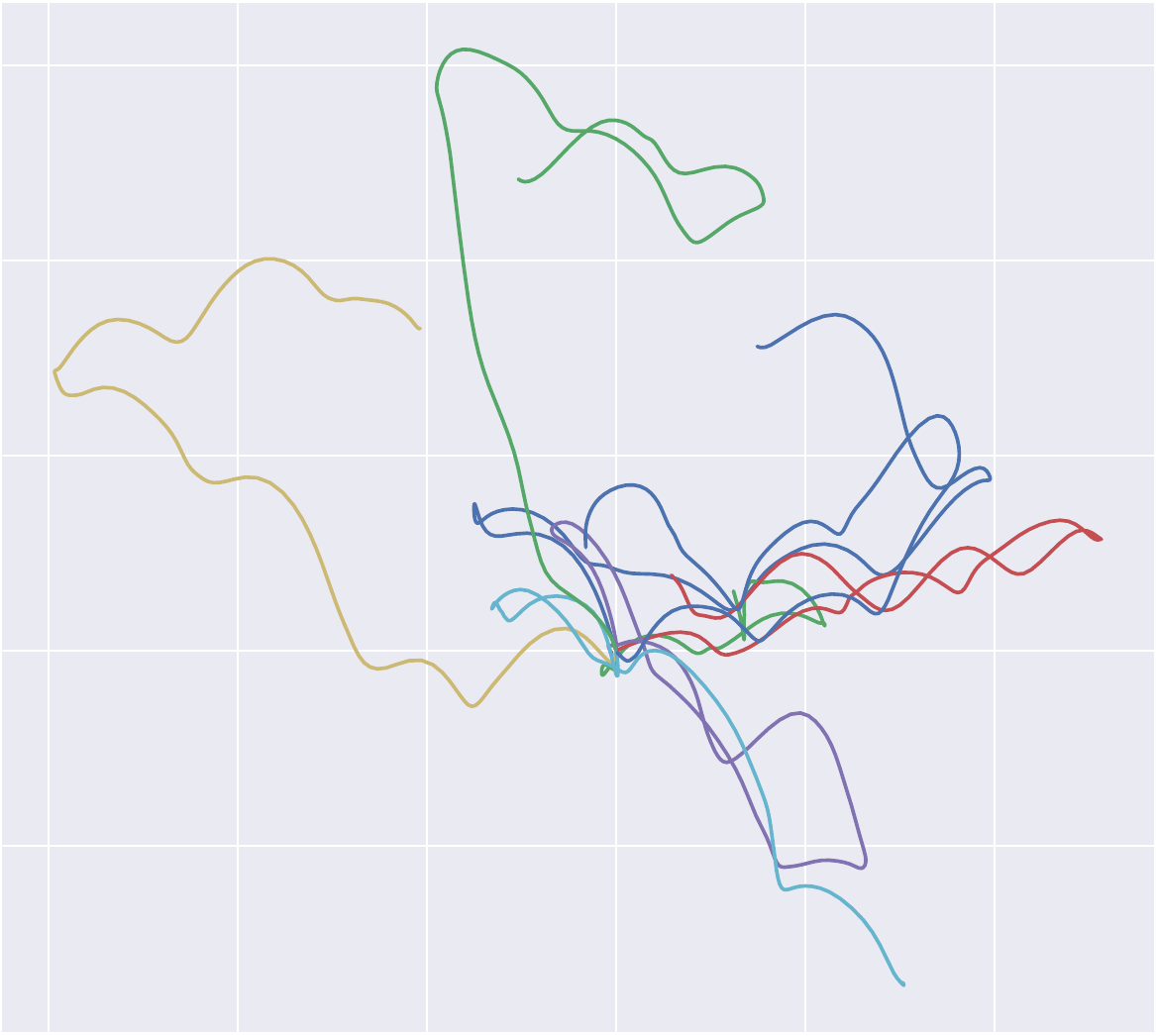}
        \caption{Root trajectory visualization.}
    \label{fig:example-based-result-b}
    \end{subfigure}
    \caption{\textbf{Diverse generated motions driven by the same example.} Prompt: ``\texttt{a person steps sideways to the left and then sideways to the right.}''. (a) The diverse generated motions driven by the same example motion share similar movement content. (b) The root trajectories of diverse motions are with similar global trajectories but not the same.}
    \label{fig:example-based-result}
\end{figure*}

\begin{figure}[!t]
    \centering
    \includegraphics[width=0.7\linewidth]{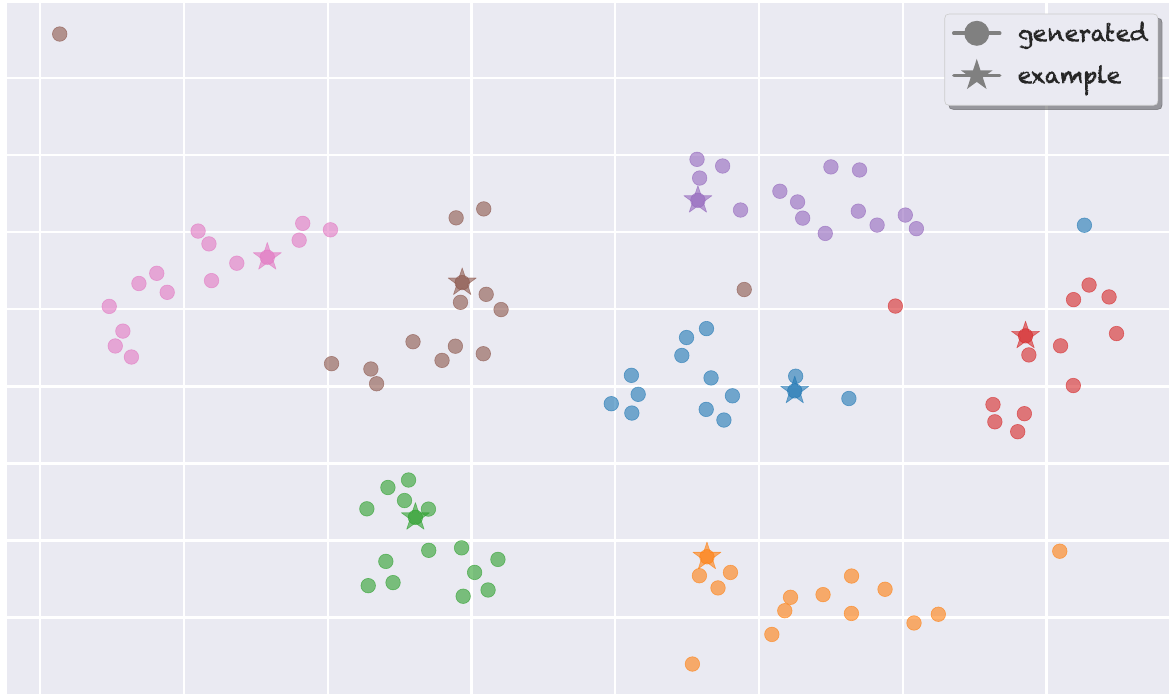}
    \captionsetup{font=small} 
    \caption{\textbf{t-SNE visualization of different example-based generated results.} Different colors imply different driven examples.}
    \vspace{-0.5em}
    \label{fig:tnse}
\end{figure}

\subsubsection{Example-based motion generation.} The example-based motion generation~\citep{weiyu23GenMM} task has two basic requirements. (1) \textit{The first one is the generated motions should share similar motion textures~\citep{motion_texture} with the example motion.} We observe the high-dimensional structure of motions via dimensionality reduction. As the t-SNE visualization results shown in~\cref{fig:tnse}, generated motions driven by the same example are similar to the given example (in the same color). (2) \textit{Besides, different generated motions driven by the same example should be diverse.} As shown in~\cref{fig:example-based-result}, these generated results are diverse not only in local motions (\cref{fig:example-based-result-a}) but also in the global trajectory (\cref{fig:example-based-result-b}). Furthermore, results in~\cref{fig:example-based-result} also share the similar motion textures. We leave more visualization results in~\cref{sec:app_example}.

Besides, we also compare our method with previous methods used for example-based motion generation on the HumanML3D test set. We use the FID and the diversity (Div.) to evaluate fidelity and generation diversity. Additionally, we also take the coverage (Cov.) metric (\cf~\citet{weiyu23GenMM}) and inference time for comparison. In this comparison, we take the motion texture~\cite{motion_texture}, GANimator~\cite{li2022ganimator}, and GenMM~\cite{weiyu23GenMM}. The comparison is shown in~\cref{tab:example_baseline}. As shown in~\cref{tab:example_baseline}, our method enjoys better editing quality and higher diversity than baselines. The coverage is comparable with the state-of-the-art methods, although our method is not designed for the specific task. The reason behind the competitive performance with baselines is that our model is pre-trained on a larger dataset than baselines, obtaining better human dynamic priors from the data distribution.

\begin{table}[!t]
\centering
\caption{{\bf Comparison with baselines.} Our method enjoys better editing quality and higher diversity than previous methods. The motion coverage is comparable with the state-of-the-art methods, although our method is not designed for the specific task.}
\resizebox{0.65\linewidth}{!}{
\begin{tabular}{lcccc}
\toprule
 & specific & FID $\downarrow$ & Diversity $\uparrow$  & Coverage (\%) $\uparrow$ \\
\hline
GANimator~\citeyearpar{li2022ganimator} & \Checkmark   & 0.505 & 2.012 & 84.1  \\
GenMM~\citeyearpar{weiyu23GenMM} & \Checkmark  & 0.514 & 2.478 & \textbf{97.5}  \\
MotionCLR & \ding{55}   & \textbf{0.427} & \textbf{2.567} & 97.0\\
\bottomrule
\end{tabular}
}
\vspace{-1em}
\label{tab:example_baseline}
\end{table}

\begin{table}[!t]
\centering
\caption{{\bf Comparison on FID and diversity values with manipulating self-attention in the motion space of the denoising process.} As can be seen in the table, our method enjoys better editing quality and higher diversity than editing at each diffusion step.}
\resizebox{0.5\linewidth}{!}{
\begin{tabular}{lcc}
\toprule
 & FID $\downarrow$ & Diversity $\uparrow$ \\
\hline
Diff. manipulation & 0.718 & 1.502 \\
MotionCLR manipulation & \textbf{0.427} & \textbf{2.567} \\
\bottomrule
\end{tabular}
}
\vspace{-1em}
\label{tab:diffusion_compare}
\end{table}

As the diffusion denoising process can manipulate the motion directly in the denoising process, we provide a baseline for comparison with our motion shifting and example-based motion generation applications. Here, \textit{for convenience}, we only take the example-based motion generation application as an example for discussion. In this section, we conduct a comparison between our proposed editing method and diffusion manipulation in the motion space, focusing on the FID and diversity metrics. The 200 samples used in this experiment were constructed by researchers. As depicted in~\cref{tab:diffusion_compare}, the ``Diff. manipulation'' serves for our comparison. Our method achieves an FID value of 0.427, indicating a relatively high generation quality, while the ``Diff. manipulation'' achieves a higher FID of 0.718, demonstrating worse fidelity. Conversely, in terms of diversity, the ``MotionCLR manipulation'' exhibits a higher diversity score of 2.567 compared to the 1.502 of the ``Diff. manipulation.'' These results verify our method is better than manipulating noisy motions in the denoising process. The main reason for the better quality and diversity mainly relies on the many times of manipulation of self-attention, but not the motion. Directly manipulating the motion results in some jitters, making more effort for models to smooth. Besides, the shuffling times of manipulating the self-attention maps are higher than the baseline, contributing to the better diversity.

\begin{figure}
    \centering
    \begin{subfigure}[b]{\linewidth}
        \centering
        \includegraphics[width=\linewidth]{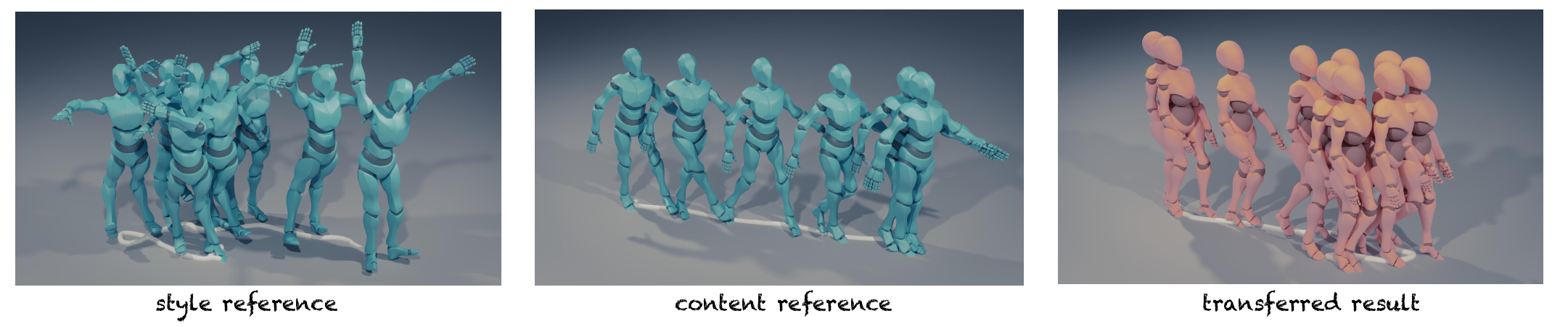}
        \caption{Style reference: ``\texttt{the person dances very happily}'', content reference: ``\texttt{the man is walking}''. The transferred result shows a figure walking in a back-and-forth happy pace.}
        \label{fig:style-res-2}
    \end{subfigure}
    \begin{subfigure}[b]{\linewidth}
        \centering
        \includegraphics[width=\linewidth]{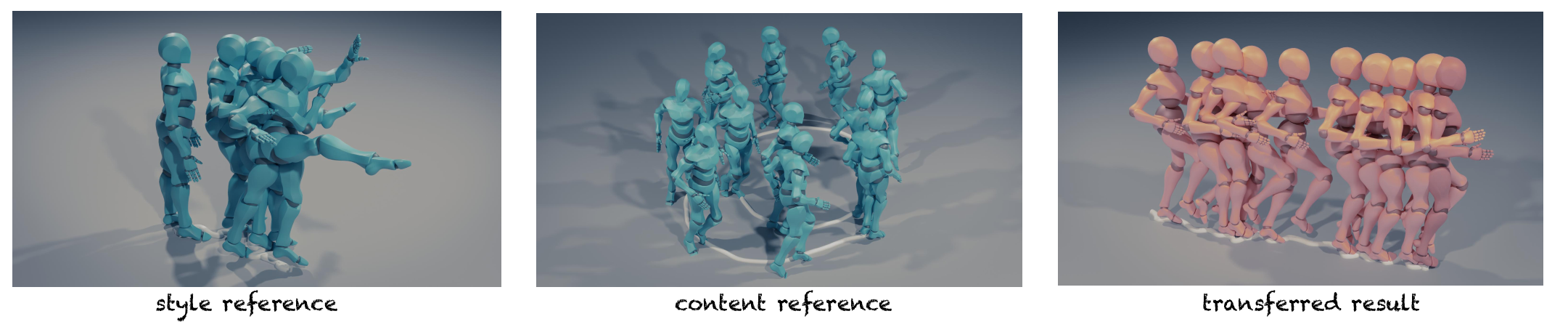}
        \caption{Style reference: ``\texttt{a man is doing hip-hop dance}'', Content reference: ``\texttt{a person runs around a circle}''. The stylized result shows a running motion with bent hands, shaking left and right.}
        \label{fig:style-res-3}
    \end{subfigure}
    \captionsetup{font=small} 
    \caption{\textbf{Motion style transfer results.} The style reference, content references, and the transferred results are shown from left to right for each case.}
    \label{fig:style-res}
\end{figure}

\begin{table}[!t]
    \centering
    \setlength{\tabcolsep}{3pt}
    \caption{{\bf Comparison of motion style transfer across baselines.} The ``Gen.'' and ``Inv.'' settings represent the editing during generation and DDIM inversion~\cite{ddim} (following~\citet{momo}) settings.}
    \resizebox{\linewidth}{!}{
    \begin{tabular}{l|ccccc}
        \toprule
         & {FID} $\downarrow$ & 
        \begin{tabular}[c]{@{}c@{}}{R Precision} \\ (Top 3)\end{tabular} $\uparrow$ & 
        \begin{tabular}[c]{@{}c@{}}{Texture Ref. Foot} \\ {Contact Sim.}\end{tabular} $\uparrow$ & 
        \begin{tabular}[c]{@{}c@{}}{Style Ref. } \\ {Loc.Sim.}\end{tabular} $\uparrow$ & AITS $\downarrow$ \\
        \hline
        MoMo (Gen.)~\citeyearpar{momo} & {2.33} & 0.439 & 0.816 & 0.972 & 0.544 \\
        MoMo (Inv.)~\citeyearpar{momo} & 2.50 & {0.490} & 0.793 & 0.856  & 0.691 \\
        Ours (Gen.)                     & \textbf{0.65} & 0.749 & \textbf{0.877}  & \textbf{0.989} & \textbf{0.345} \\
        Ours (Inv.)                     & 0.69 & \textbf{0.784} & 0.851  & 0.914 & 0.432 \\
        \bottomrule
    \end{tabular}
    }
    \label{tab:compare_momo}
\end{table}

\subsubsection{Motion style transfer.} As shown in~\cref{fig:style-res}, in the MotionCLR framework, the style reference motion provides style and the content reference motion provides keys and values. As can be seen in~\cref{fig:style-res}, all edited results are well-stylized with style motions and keep the main movement content with the content reference. 
To qualitatively evaluate the transferred result, we compare our method with the latest method, MoMo~\cite{momo}. We follow the evaluation protocol in~\citet{momo}, where the comparison results in~\cref{tab:compare_momo} indicate the better transfer result than MoMo. The quality gain mainly comes from the better base model of MotionCLR and the disentangled modeling of self-attention and cross-attention. Besides, we also compare the speed (AIST metric following~\citet{mld}) with baselines, indicating better efficiency of MotionCLR.

\subsection{Ablation Study}

\begin{table}[!t]
\centering
\setlength{\tabcolsep}{9pt}
\caption{Ablation studies between different technical design choices.}
\resizebox{0.5\textwidth}{!}{
\begin{tabular}{cccccc}
\toprule
\multirow{2}{*}{Ablation} & \multicolumn{3}{c}{R-Precision$\uparrow$} & \multirow{2}{*}{FID$\downarrow$} \\
 \cmidrule(lr){2-4}
 & Top 1 & Top 2 & Top 3 &    \\
\midrule
(1)  & 0.512 & 0.705 & 0.792  &  0.544 \\
(2)  & 0.509 & 0.703 & 0.788  &  0.550 \\ 
\hline \specialrule{0em}{1pt}{1pt}
MotionCLR & \textbf{0.544} & \textbf{0.732} & \textbf{0.831} & \textbf{0.269}   \\
\bottomrule
\end{tabular}
} 
\vspace{-0.5em}
\label{tab:ablation1}
\end{table}

\begin{table}[!t]
\centering
\caption{\textbf{The ablation study of manipulating different attention layers on the HVerb-wild test set.} The ``begin'' and ``end'' represent the beginning and the final layer/step for manipulation. The bottom row denotes our design choice for motion editing.}
\setlength{\tabcolsep}{3pt}
\resizebox{0.8\linewidth}{!}{
\begin{tabular}{ccccccc}
\toprule
\multicolumn{2}{c}{editing steps}  & \multicolumn{2}{c}{editing layers}  & \multirow{2}{*}{FID$\downarrow$}   & \multirow{2}{*}{\begin{tabular}[c]{@{}c@{}}align with \\ edited text\end{tabular} (\%) $\uparrow$}   & \multirow{2}{*}{\begin{tabular}[c]{@{}c@{}}align with \\ original text\end{tabular} (\%) $\downarrow$} \\
 \cmidrule(lr){1-2} \cmidrule(lr){3-4} begin & end &  begin & end &  \\
 \hline
 1 &  9 &  8     & 11  & 0.339 & 63.5  &  62.5  \\ 
 1 &  9 &  5     & 14  & 0.335 & 65.7  &  59.2\\ \hline
 3 &  7 &  1     & 18  & 0.399 & 64.7  &  59.3 \\  
 4 &  6 &  1     & 18  & 0.455 & 62.8  &  59.0 \\ \hline
 1 &  9 &  1     & 18  & \textbf{0.330} & \textbf{66.0} &  \textbf{58.1} \\ \bottomrule
\end{tabular}
}
\label{tab:ablation_replace}
\end{table}

In this section, we provide ablation studies for the generation function and the editing functions, respectively.

\myPara{Ablation study of the generation ability in MotionCLR.} We provide some ablation studies on some technical designs in~\cref{tab:ablation1}. (1) The setting \textit{w/o separate word modeling} shows poorer qualitative results with the w/ separate word setting. The separate word-level cross-attention correspondence benefits better text-to-motion controlling, which is critical for motion fine-grained generation. (2) The setting of \textit{injecting text tokens before motion tokens} performs worse than the MotionCLR. This validates the effectiveness of introducing the cross-attention for cross-modal correspondence. The ablation studies additionally verify the basic motivation of modeling word-level correspondence in MotionCLR.

\myPara{Ablation study of editing different attention layers.} To further explore the impact of attention manipulation, we conduct an ablation study by varying the layers in MotionCLR for manipulation, shown in~\cref{tab:ablation_replace}. Without losing generalization, we test this on the in-place motion replacement application. The table lists the results for different ranges of manipulated attention layers. It can be observed that manipulating different attention layers influences the editing quality and the semantic similarity (TMR-sim.). In particular, manipulating the layers from 1 to 18 achieves the best semantic consistency, demonstrating the effectiveness of editing across multiple attention layers for maintaining semantic alignment in the edited motion. The less effectiveness of manipulating middle layers is mainly due to the fuzzy semantics present in the middle layers of the U-Net. As these layers capture more abstract with reduced temporal resolution, the precise details and localized information become less distinct. Consequently, manipulating these layers has a limited impact on the final output, as they contribute less directly to the fine-grained details of the task.

\myPara{Ablation study of editing at different diffusion steps.} We conducted an ablation study to evaluate the impact of editing across different ranges of diffusion steps and compare various design choices (as shown in Table~\cref{tab:ablation_replace}). Here, without losing generalization, we also test this on the in-place motion replacement application. The results show that editing across a broader range of diffusion steps (\eg, steps 1 to 9) achieves the best balance between semantic consistency with the edited text and the quality of the generated motion (FID). This is because the early steps focus on establishing the global semantic structure, while the later steps refine the fine-grained details. Covering the entire range effectively combines the strengths of both stages. In contrast, limiting edits to narrower ranges, such as the middle or late steps (e.g., steps 4 to 6), results in lower semantic alignment with the edited text, as these steps mainly refine details and have limited influence on the global structure. This highlights the importance of carefully designing the range of diffusion steps to achieve high-quality, semantically consistent motion editing.

\begin{figure*}[!t]
    \vspace{-0.5em}
    \captionsetup[figure]{aboveskip=0pt, belowskip=0pt, font=small}
    \centering
    \includegraphics[width=\linewidth]{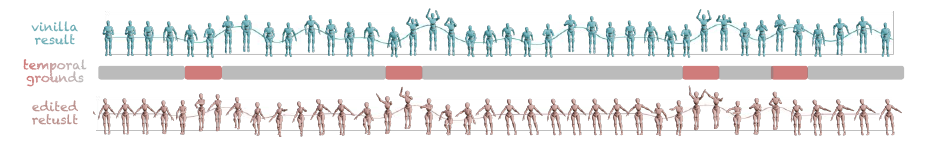}
    \captionsetup{font=small}
    \caption{\textbf{Comparison between w/ \vds w/o grounded motion generation settings.} The root height and motion visualization of the textual prompt ``a person jumps four times''. }
    \label{fig:failure-case}
\end{figure*}

\subsection{User Study}
\label{sec:user_study}

\begin{table}[!t]
\setlength{\tabcolsep}{9pt}
\caption{{\bf User study results.} The study evaluates the motion (de-)emphasis and in-place replacement quality, respectively. The results show that our method outperforms other baselines across fidelity, user requirement satisfaction, and content similarity with the unedited motion.}
\centering
\resizebox{0.8\linewidth}{!}{
\begin{tabular}{l|ccc}
\toprule
(de-)emphasis \ \ \ \   & fidelity $\uparrow$ & satisfactory $\uparrow$ & similarity $\uparrow$ \\ 
\hline
MotionFix   &   3.51      &    3.80        &   4.07     \\
Ours      &    4.02         &    3.95      &   4.25      \\
\hline 
\end{tabular}
}
\resizebox{0.8\linewidth}{!}{
\begin{tabular}{l|ccc}
\hline
replacement  & fidelity $\uparrow$ & satisfactory $\uparrow$ & similarity $\uparrow$ \\ 
\hline
editing text only   &    3.04     &   3.55      &    3.01      \\   
MotionFix   &     3.10     &     2.78       &     3.22     \\
Ours              &   4.05     &    4.26     & 3.99 \\
\bottomrule
\end{tabular}
}
\label{tab:userstudy}
\vspace{-1em}
\end{table}

To additionally verify the effectiveness of the semantic editing result, we also compare our method with baselines via human evaluation. Here, for the semantic motion editing method, we mainly focus on the motion (de-)emphasis and the im-place motion replacement task, both of which are based on editing cross-attention. 

Similar to~\citet{goel2024iterative}, we set up the user study for 30 participants to evaluate 10 groups of motion editing results. For motion emphasis and de-emphasis applications, we evaluate the result based on 1) the edited motion quality (\aka fidelity), 2) whether the edited result satisfies the requirement, and 3) content similarity between edited and unedited motions. For the in-place motion replacement application, the evaluation is based on 1) the edited motion quality (fidelity), 2) whether the edited result satisfies the requirement, and 3) the motion reserving of the unedited part. We set the latest method MotionFix as our baseline, and include directly editing text as another baseline for the in-place motion replacement application. For MotionFix in motion (de-)emphasis, we annotate the prompts by human researchers, like using ``jump higher'' to emphasize ``jump''. For the in-place replacement application, we keep the prompts usage in~\cref{tab:replacenew}.
As shown in~\cref{tab:userstudy}, our method outperforms baselines across motion fidelity, user requirement satisfaction, and motion content preserving. These gains mainly come from our method not relying on any specifically designed data construction process, like motion retrieval in~\citet{athanasiou2024motionfix}.

\subsection{Potential Action Counting Ability from Attention Map} 
\label{sef:actioncount}

\begin{figure}[!t]
    \centering
    \includegraphics[width=0.7\linewidth]{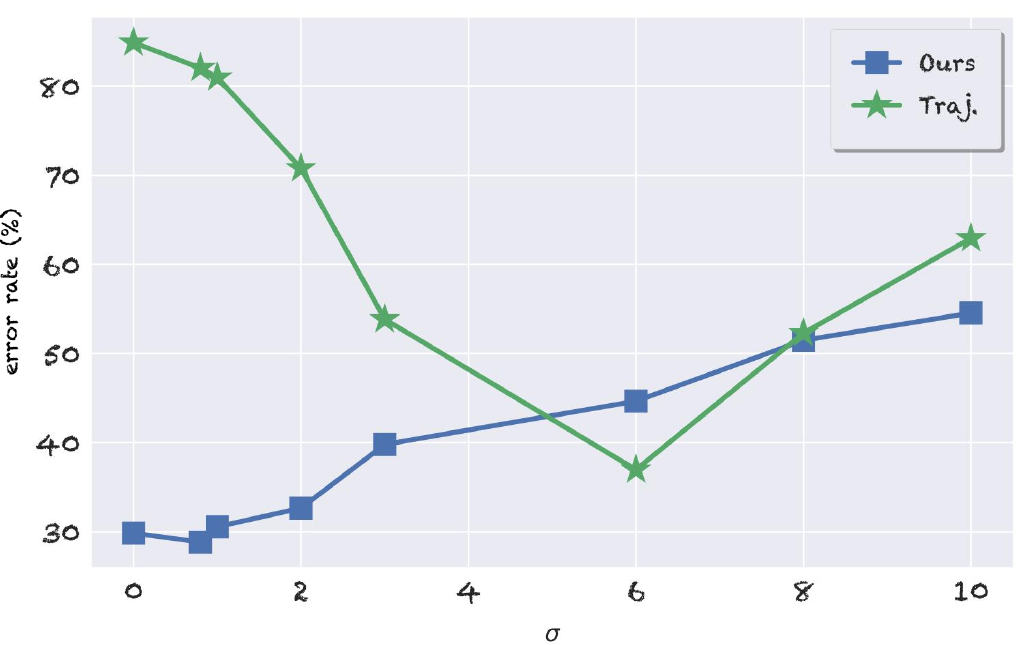}
    \caption{{\bf Action counting error rate comparison.} Root trajectory (Traj.) \vds attention map (Ours). ``$\sigma$'' is the smoothing parameter.}
    \label{fig:counting_main}
    \vspace{-1em}
\end{figure}

As shown in~\cref{fig:study_attn}, the number of executed actions in a generated motion sequence can be accurately calculated via the self-attention map. We directly detect the number of peaks in each row of the self-attention map and finally average this of each row. In the technical implementation, to avoid sudden peaks from being detected, we apply average downsampling and Gaussian smoothing (parameterized by standard deviation $\sigma$). We leave more technical details in~\cref{sec:app_detial_count}. 

We construct a set of text prompts corresponding to different actions to perform the counting capability via the self-attention map. The counting number of actions is labeled by professional researchers. The details of the evaluation set are detailed in~\cref{sec:app_evaluation}. As the ``walking'' action is composed of sub-actions of two legs, the atomic unit of this action counting is set as 0.5. We compare our method to counting with the vertical root trajectory (Traj.) peak detection. As shown in~\cref{fig:counting_main}, counting with the self-attention map mostly works better than counting with root trajectory. Both settings use Gaussian smoothing to blur some jitters. Our method does not require too much smoothing regularization due to the smoothness of the attention map, while counting with root trajectory needs this operation. This case study reveals the effectiveness of understanding the self-attention map in MotionCLR.

\section{Case Study: Failure Cases Analysis and Correction}
\label{sec:failcase}

There are few generative methods that can escape the curse of hallucination. 
In this section, we will discuss some failure cases of our method and analyze how we can refine these results. The hallucination of counting is a notoriously tricky problem for generative models, attracting significant attention in the community and lacking a unified technical solution. Considering that this problem cannot be thoroughly resolved, we try to partially reveal this issue by additionally providing temporal grounds. For example, if the counting number of an action is not aligned with the textual prompt, we can correct this by specifying the temporal grounds of actions. Technically, the temporal mask can be treated as a sequence of weights to perform the motion emphasis and de-emphasis. Therefore, grounded motion generation can be easily achieved by adjusting the weights of words. 

Specifically, we show some failure cases of our method. As shown in~\cref{fig:failure-case}, the generated result of ``\texttt{a person jumps four times}'' fails to show \textit{four} times of jumping actions, but \textit{seven} times. To meet the requirement of counting numbers in the textual prompts, we additionally input a temporal mask, including \textit{four} jumping timesteps, to provide temporal grounds. Technically, for the desired four jumping areas, we set $\alpha=0.15$ as emphasis. For the reserving area, we set $\alpha=-0.1$ as de-emphasis. From the root height visualization and the motion visualization, the times of the jumping action have been successfully corrected from \textit{seven} to \textit{four}. Therefore, our method is promising for \textit{grounded motion generation} to reveal the hallucination of deep models. 
Moreover, other editing fashions are also potential ways to correct hallucinations of generated results. For example, the motion sequence shifting and in-place motion replacement functions can be used for correcting sequential errors and semantic misalignments, respectively.

\begin{figure}[!t]
    \centering

    \begin{subfigure}[b]{0.45\textwidth}
        \centering
        \includegraphics[width=\textwidth]{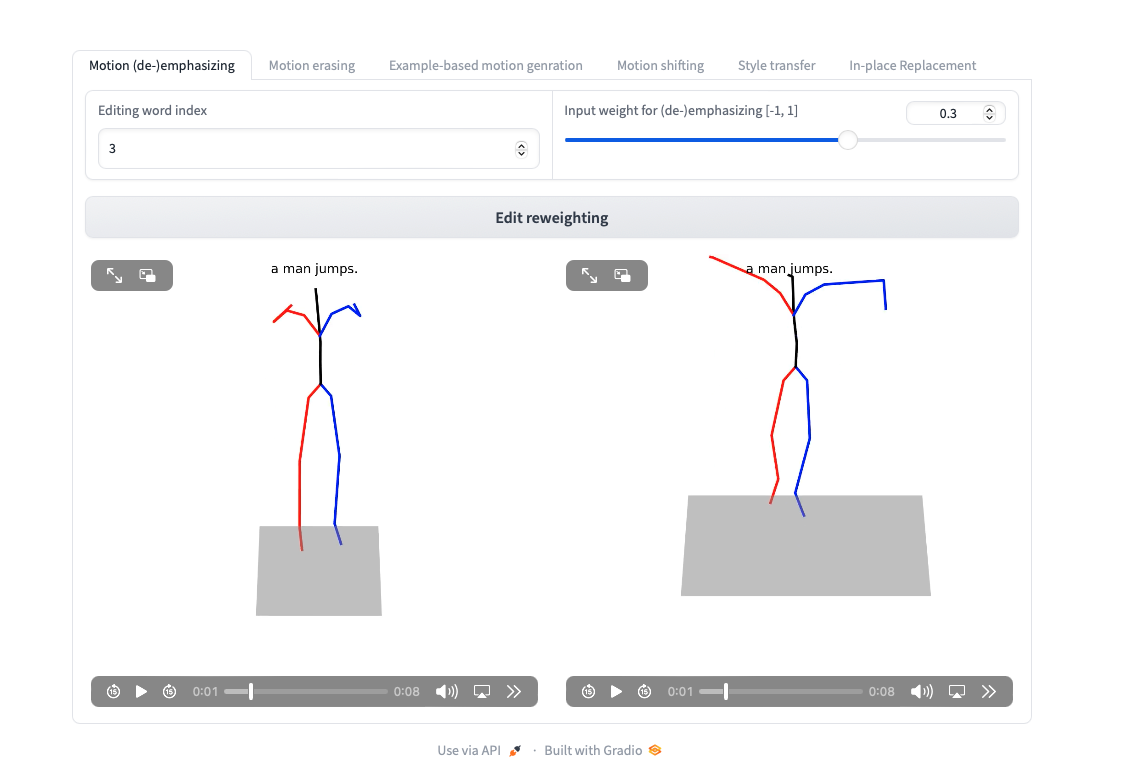}
        \caption{Web interface for MotionCLR.}
        \label{fig:interface_1}
    \end{subfigure}
    \begin{subfigure}[b]{0.45\textwidth}
        \centering
        \includegraphics[width=0.99\linewidth]{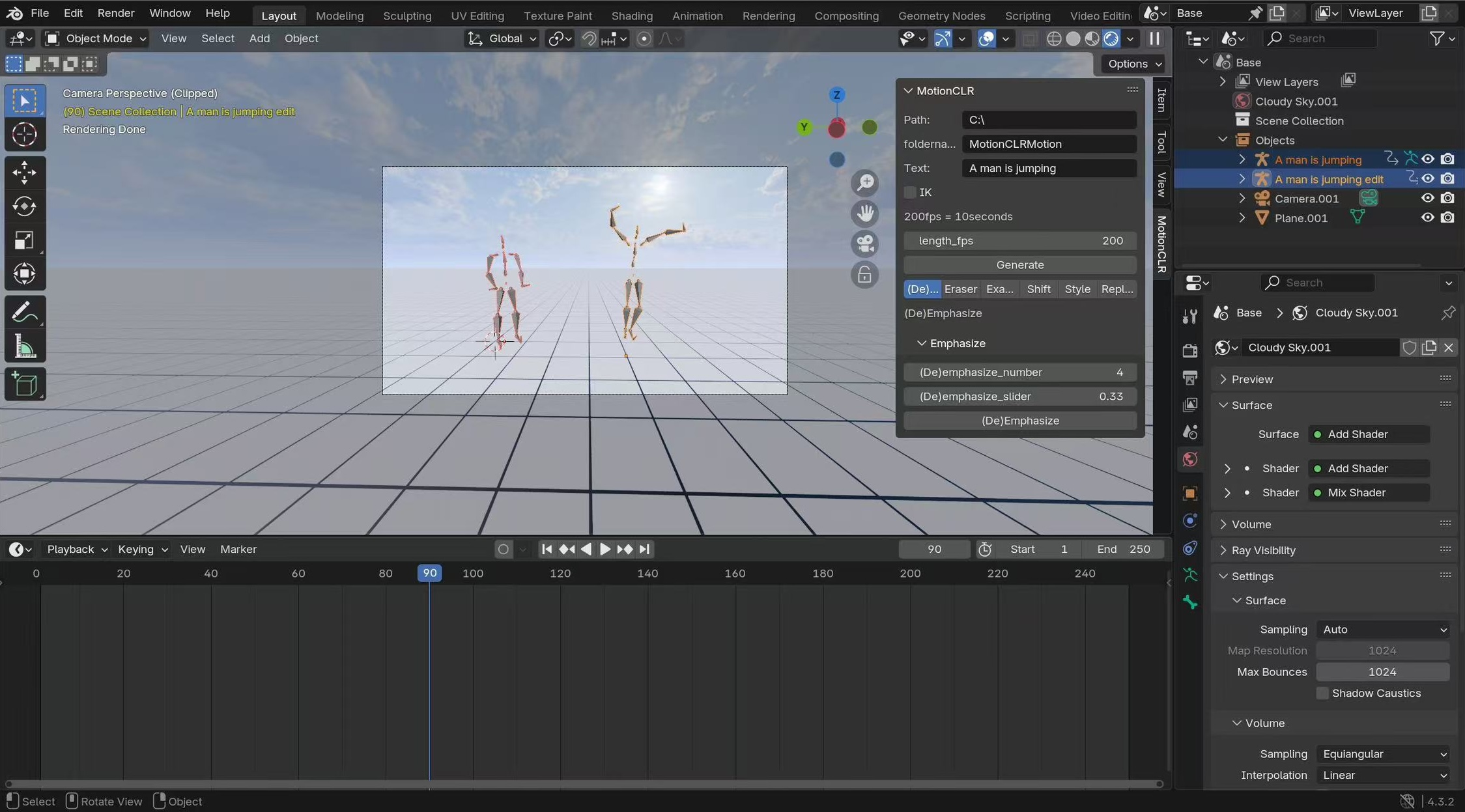}
        \caption{Blender add-on for MotionCLR. }
        \label{fig:interface_2}
    \end{subfigure}
    \caption{{\bf Web interface and the Blender add-on of MotionCLR, supporting proposed applications.} In these two interfaces, we take the motion (de-)emphasis application as an example. Both interfaces support adjusting the weight of parameters to refine the result, until reaching satisfaction.}
    \label{fig:addon}
\end{figure}

\section{User Interface}
\label{sec:addon}

MotionCLR is a simple yet effective framework that can be integrated into industrial tools. We develop a web interface of the MotionCLR model for interactive motion generation and editing. The web interface provides a quick view for new customers without installing any dependencies. For experienced users, like animators, we also support a Blender add-on of MotionCLR similar to the web demo, supporting the real creation loop of motion synthesis. Two interfaces are shown in~\cref{fig:interface_1} and \cref{fig:interface_2}, respectively. In~\cref{fig:addon}, both the web interface and add-on support all applications of MotionCLR. Besides, users are allowed to \textit{interactively} adjust the weight of parameters to refine the result, until reaching their satisfaction.

\section{Discussion and Conclusion}

In this work, we carefully clarify what roles cross-attention plays in motion generation, enhancing the explainability of attention mechanisms in text-to-motion generation. Except for the comparable generative performance with SOTA methods, our proposed MotionCLR model supports diverse interactive motion generation and editing in one system for the first time. Importantly, in the editing process, MotionCLR supports editing a motion without any training, which even outperforms some methods requiring specific training. 

To have a thorough exploration of the proposed method, we construct an evaluation set to validate the theoretical analysis of the attention mechanism. Based on these preliminary experiments and new understanding, we evaluate our applications with specific baselines. Our method even outperforms some specifically designed methods in some scenarios. Besides, we also explore the boundaries of our method in action counting and grounded motion generation. Considering real-world applications in the animation creation, we develop a web interface and a Blender add-on for users. 

This work explores the new interaction fashion of motion synthesis. We are still facing some limitations, driving for developing future methods.
1) {\bf \textit{Different interaction fashions.}} This work introduces a new interaction fashion for motion generation, relying on multiple interactions with the machine. However, the language-based chatting interaction is also useful to users, which is leaving as our future work. Besides, MotionCLR can also serve as a data generator to synthesize a dataset for such applications. 2) {\bf \textit{ Robustness of the method.}} As shown in~\cref{sec:failcase}, our model can also not escape the hallucination curse of generative models. Although the proposed grounded motion generation method can relieve the phenomena, it is still worth enhancing the capability of the base model. To this end, the grounded synthesis method could provide additional supervision in this process. We will reach this in the future. Besides, our method will also benefit from a larger dataset with good quality, which is already in the collecting process before submission.

\newpage

\bibliographystyle{plainnat}
\bibliography{main}

\clearpage

\appendix
\section*{Appendix}



\section{Supplemental Experiments}

\label{sec:main_vis}

\subsection{What is the Self-attention Map like in Motion (De-)emphasis?}
\label{sec:emph_supp}

\begin{figure}[h]
    \centering
    \includegraphics[width=\linewidth]{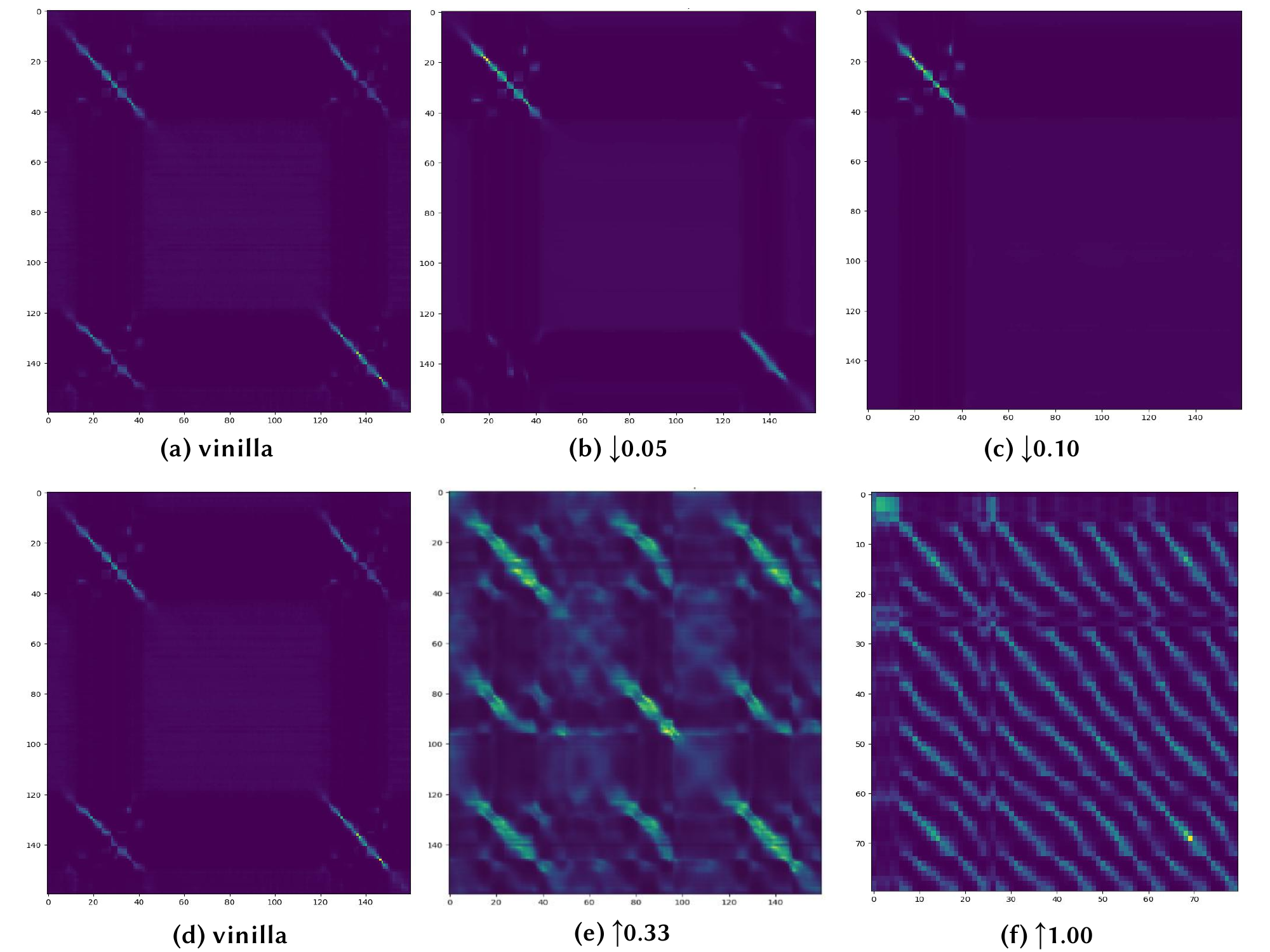}
    \caption{{\bf Additional visualization results for different (de-)emphasis weights.} The self-attention maps show how varying the different weights (\eg, $\downarrow 0.05$, $\downarrow 0.10$, $\uparrow 0.33$, and $\uparrow 1.00$) affect the emphasis on motion.} 
    \vspace{-1em}
    \label{fig:emph_supp}
\end{figure}

This experiment is an extension of the experiment shown in~\cref{fig:emphasis}.
We provide more examples of how increasing or decreasing weights impact motion (de-)emphasis and erasing. As seen in \cref{fig:emph_supp}, the attention maps illustrate that reducing the weights (\eg, $\downarrow 0.05$ and $\downarrow 0.10$) results in less activations, while increasing weights (\eg, $\uparrow 0.33$ and $\uparrow 1.00$) leads to more activations. The vanilla map serves as a reference without any adjustments. However, as indicated, excessively high weights such as $\uparrow 1.00$ introduce some artifacts, emphasis the need for careful tuning of weights to maintain the integrity of the generated motion outputs.
This demonstrates the importance of careful weight tuning to achieve the desired motion emphasis or erasure.
Compared to \cref{fig:emph_supp}a, \cref{fig:emph_supp}b shows two fewer trajectories. This reduction is due to the de-emphasis effect, where the character's second jump was not fully executed, resulting in just an arm motion. Consequently, the two actions became distinguishable, leading to fewer detected two trajectories. In \cref{fig:emph_supp}c, the second jumping has been completely erased, resulting in only one trajectory, further demonstrating how de-emphasis significantly affects motion execution.

\subsection{Motion Generation Result Visualization}
\label{sec:main_vis_motion}

\begin{figure*}[!t]
    \centering
    \includegraphics[width=0.95\linewidth]{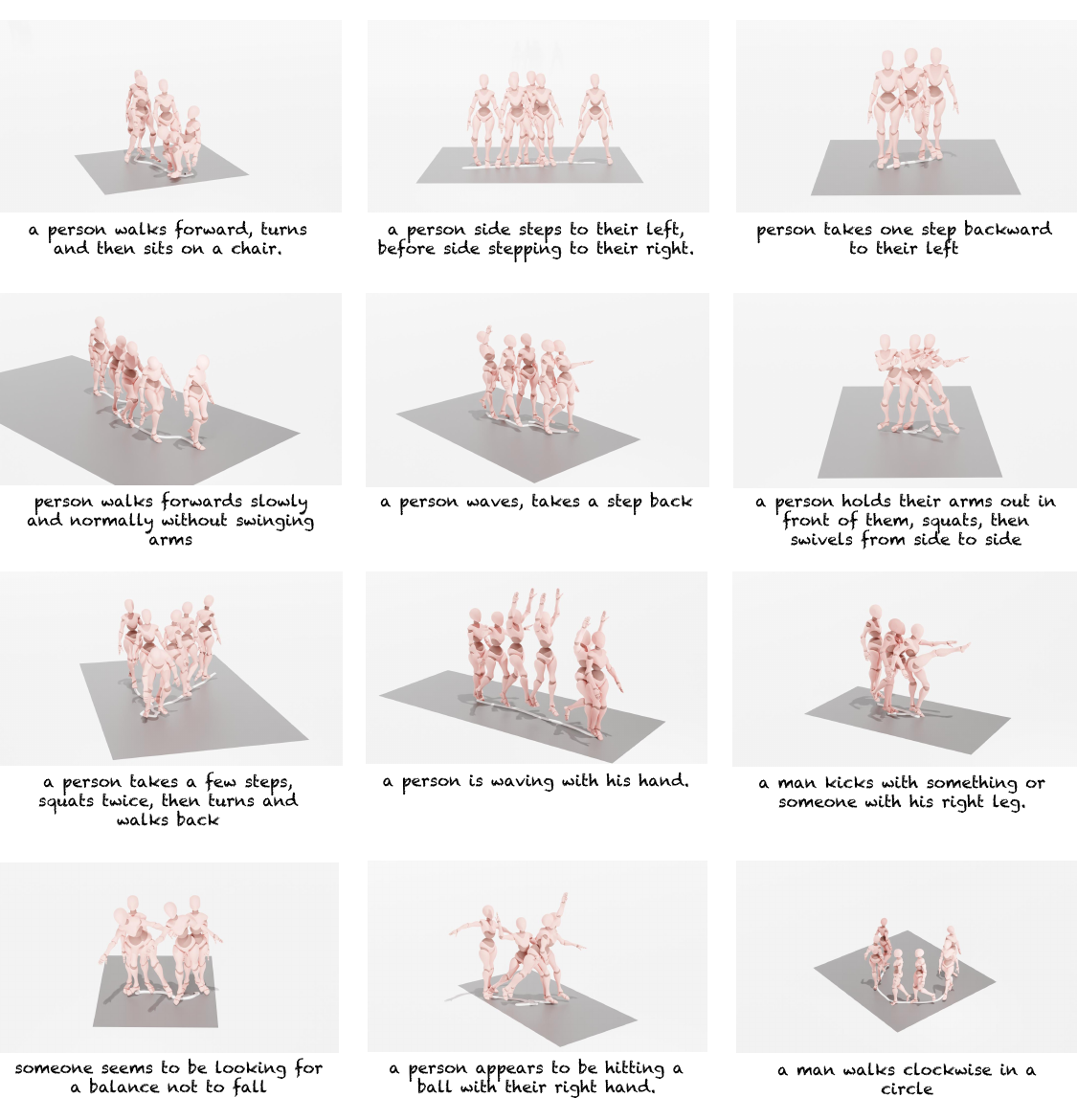}
    \caption{\bf More visualization of human motion generation result by MotionCLR.}
    \vspace{1.5em}
    \label{fig:main-vis-gen}
\end{figure*}

We randomly chose some examples of the motion generation results in~\cref{fig:main-vis-gen}. The visualization results demonstrate that MotionCLR can generate coherent and realistic human motions based on diverse textual descriptions. The generated sequences capture various actions ranging from simple gestures to more complex movements, indicating the capability to handle a wide range of human behaviors.
Overall, the qualitative results suggest that MotionCLR effectively translates textual prompts into human-like motions with a clear understanding of texts. This demonstrates the potential for applications in scenarios requiring accurate motion generation based on natural language inputs.

\subsection{More Visualization Results of Motion Sequence Shifting}
\label{sec:app_shift}

\begin{figure*}[!ht]
    \centering
    \captionsetup[subfigure]{aboveskip=0pt, belowskip=0pt}
    \begin{subfigure}[b]{0.47\textwidth}
        \centering
        \includegraphics[width=0.85\textwidth]{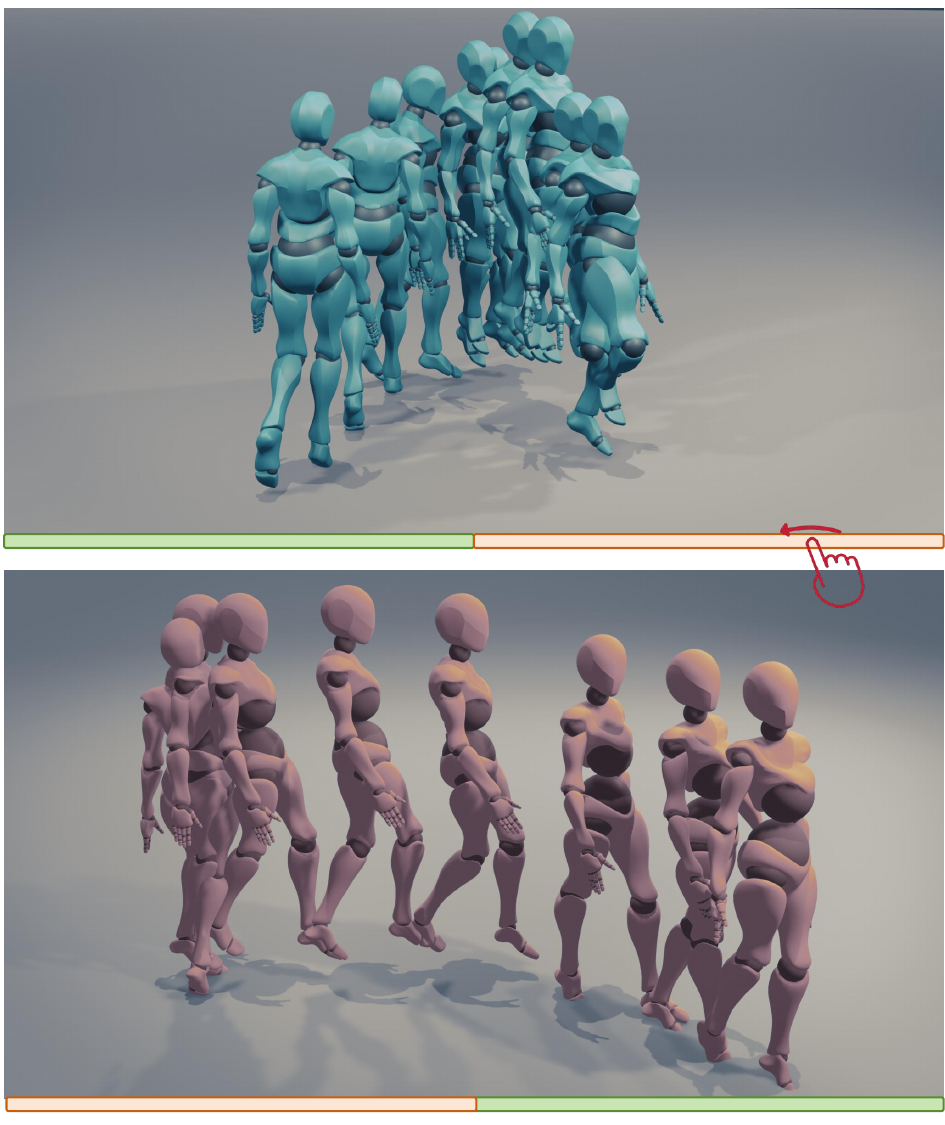}
        \caption{Prompt: ``\texttt{ the person is walking forward on uneven terrain.}'' Original (blue) \vds shifted (red) motion.}
    \label{fig:shift-result-a-appn}
    \end{subfigure}
    \hspace{1.5em}
    \begin{subfigure}[b]{0.47\textwidth}
        \centering
        \includegraphics[width=0.85\textwidth]{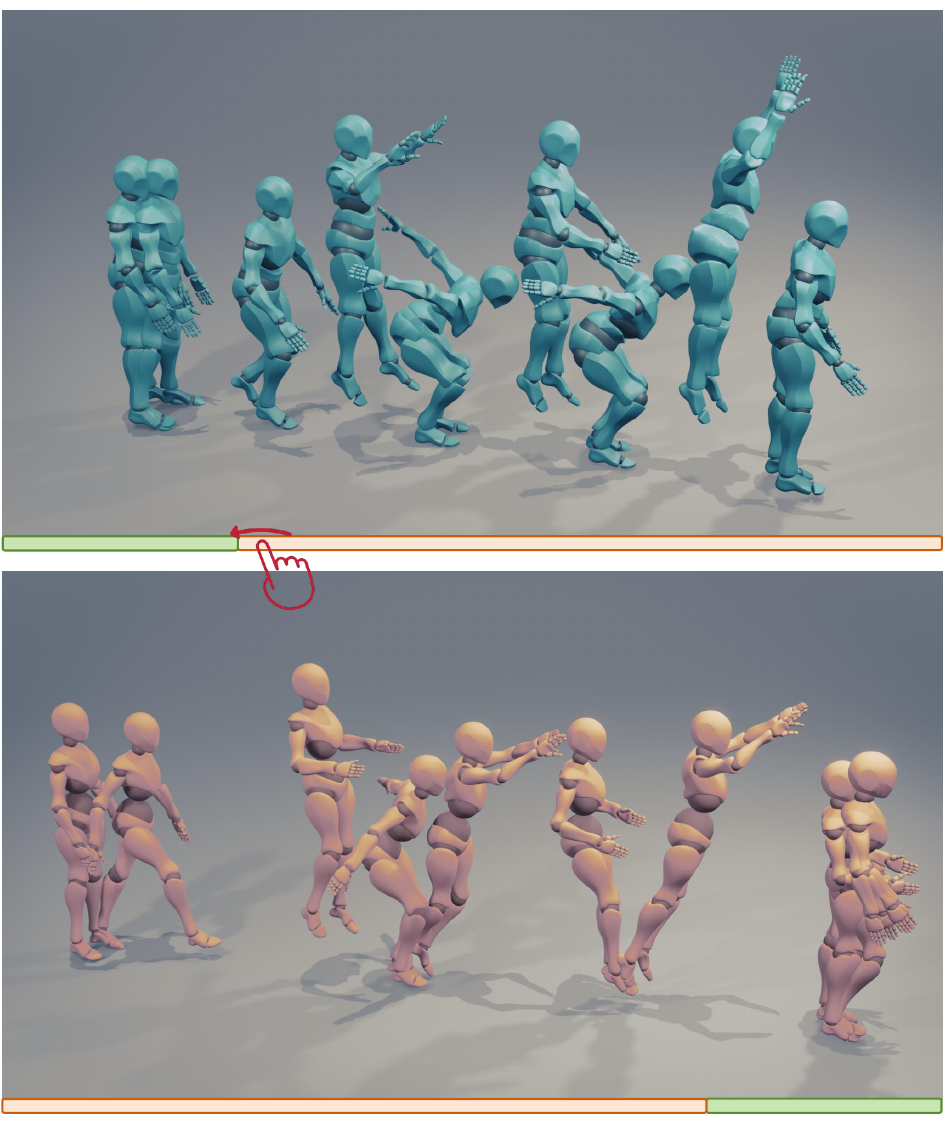}
        \caption{Prompt: ``\texttt{a person walks then jumps.}'' Original (blue) \vds shifted (red) motion.}
    \label{fig:shift-result-b-appn}
    \end{subfigure}
    \caption{\small \textbf{Comparison between original motion and the shifted motion.} The shifted time bars are shown in different colors. (a) The original figure crosses the obstacle after the walking action. The shifted motion has the opposite sequentiality. (b) The key walking and jumping actions are shifted to the beginning of the sequence, and the standing-by action is shifted to the end.}
    \label{fig:shift-result-appn}
\end{figure*}

We present further comparisons between the original and edited motions in \cref{fig:shift-result-appn}. The time bars, indicated by ``\tikz[baseline]{\node[draw=myorange2, fill=myorange1, rounded corners=0.05cm, minimum width=0.8cm, minimum height=0.15cm, anchor=base, yshift=0.1cm] {}; }'' and ``\tikz[baseline]{\node[draw=mygreen2, fill=mygreen1, rounded corners=0.05cm, minimum width=0.8cm, minimum height=0.15cm, anchor=base, yshift=0.1cm] {}; },’’ represent distinct phases of the motion, with their sequential arrangement reflecting the progression of the motion over time.

In \cref{fig:shift-result-a-appn}, we observe that the action of crossing the obstacle, originally positioned earlier in the sequence, is shifted towards the end in the edited version. This adjustment demonstrates the model’s capacity to rearrange complex motions effectively while maintaining coherence. Similarly, \cref{fig:shift-result-b-appn} shows the standing-by action being relocated to the end of the motion sequence. This change emphasizes the model’s ability to handle significant alterations in the temporal arrangement of actions. These results collectively indicate that our editing process, driven by the attention map sequentiality, exhibits a high level of correspondence with the intended edits to the motion’s sequence. The model accurately captures and replicates the desired modifications, ensuring that the restructured motion retains a natural and logical flow, thereby validating the effectiveness of our motion editing approach.

\subsection{More Visualization Results on Exampel-based Motion Generation}
\label{sec:app_example}

\begin{figure*}[htp]
    \centering
    \begin{minipage}{0.48\textwidth}
    \begin{subfigure}[b]{\linewidth}
    \centering
    \includegraphics[width=0.9\linewidth]{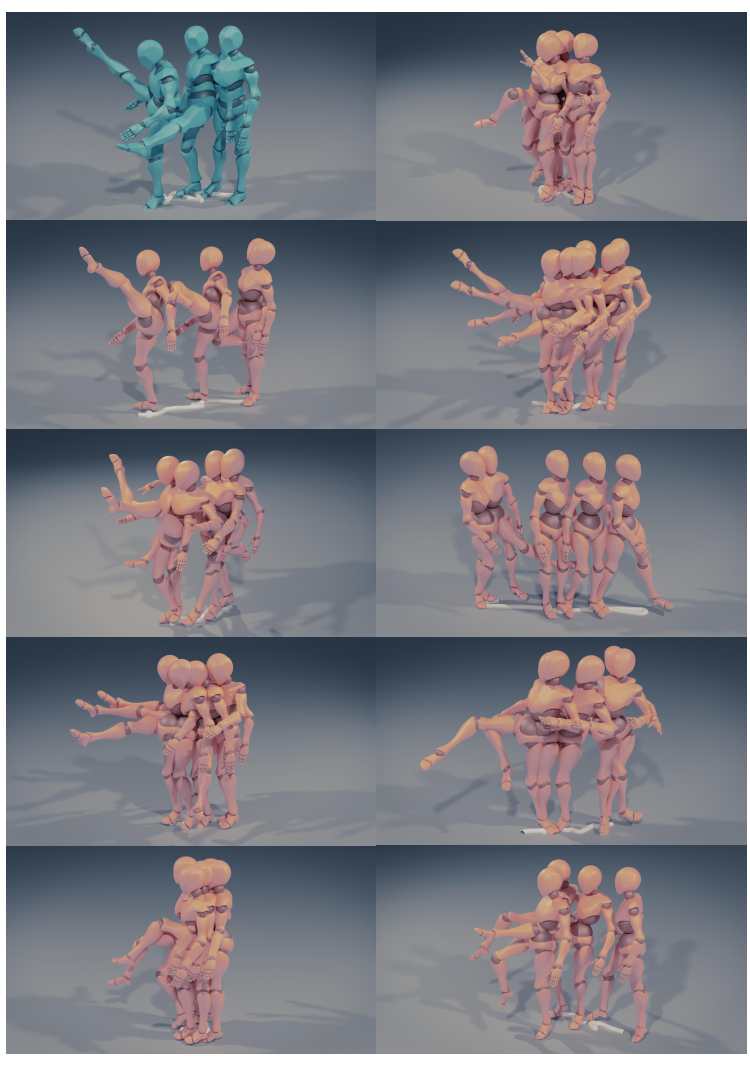}
    \caption{The example motion (blue) and the generated diverse motion (red).}
    \label{fig:diverse_case5_sub1}
    \end{subfigure}
    \begin{subfigure}[b]{\linewidth}
    \centering
    \includegraphics[width=0.9\linewidth]{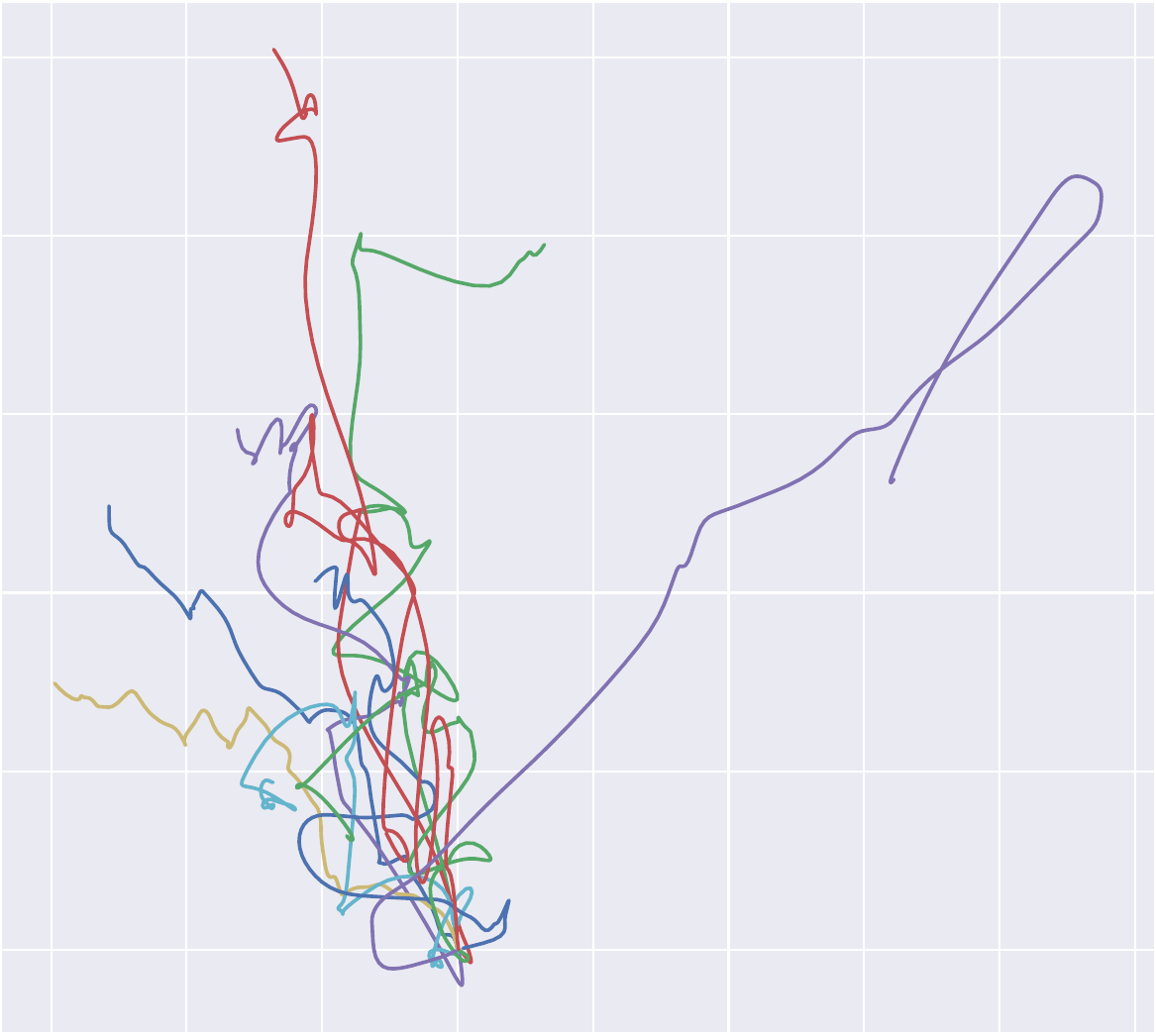}
    \caption{The trajectory visualizations of the example motion and diverse motions.}
    \label{fig:diverse_case5_sub2}
    \end{subfigure}
    \caption{\textbf{Diverse generated results of blue example generated by the prompt ``\texttt{a person kicks their feet.}''.} The example-based generated kick motions are diverse and similar to the source example.}
    \label{fig:diverse_case5}

    \end{minipage} \hfill
    \begin{minipage}{0.48\textwidth} 

    \begin{subfigure}[b]{\linewidth}
    \centering
    \includegraphics[width=0.9\linewidth]{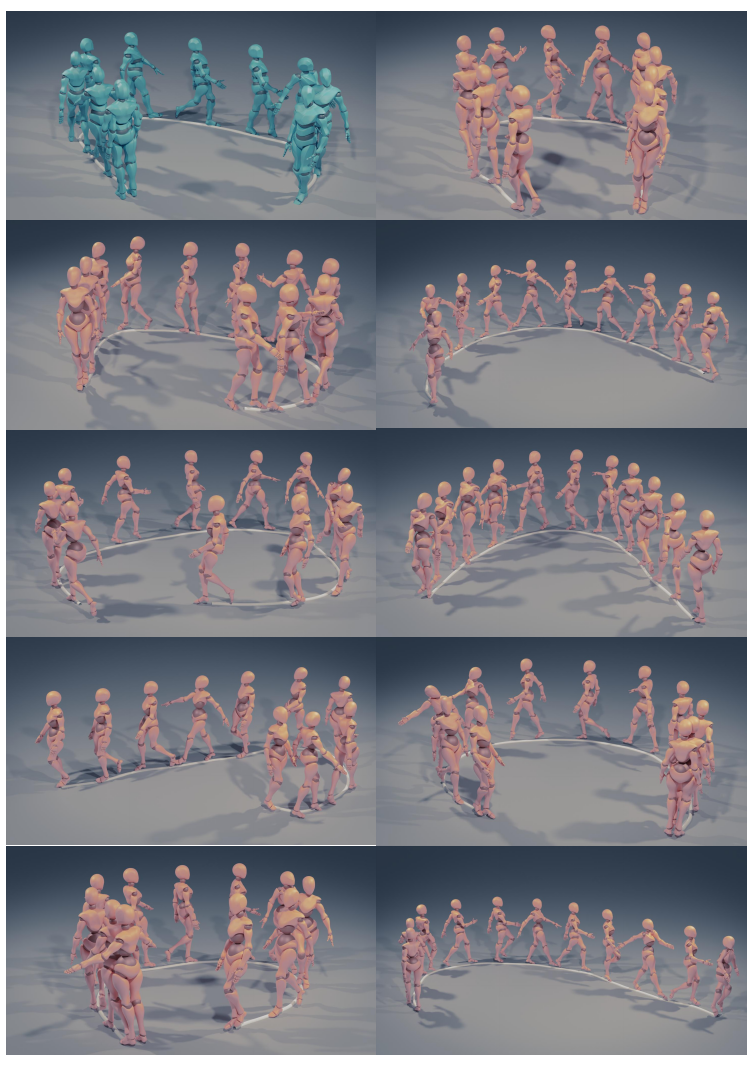}
    \caption{The example motion (blue) and the generated diverse motion (red).}
    \label{fig:diverse_case7_sub1}
    \end{subfigure}
    \begin{subfigure}[b]{\linewidth}
    \centering
    \includegraphics[width=0.9\linewidth]{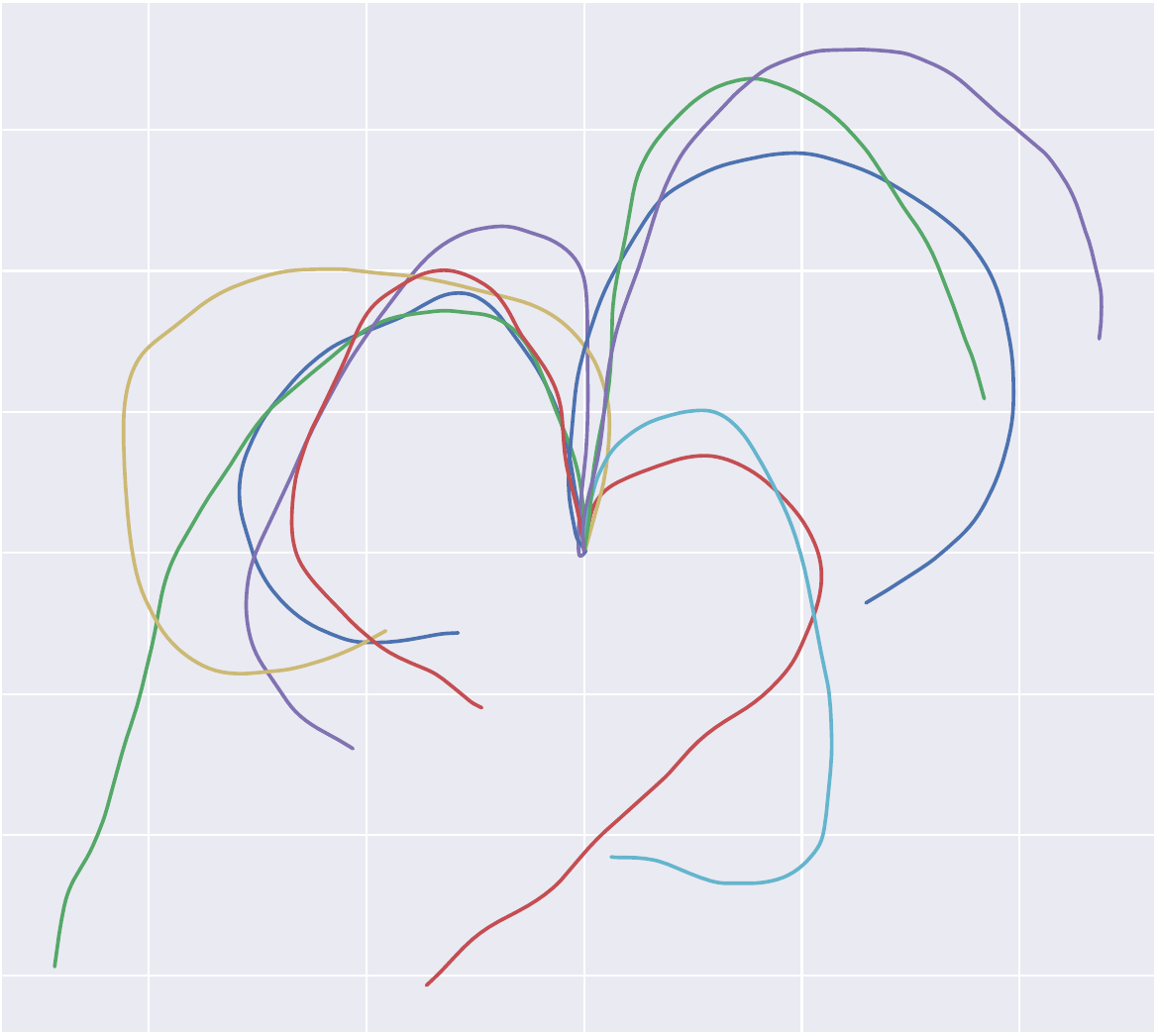}
    \caption{The trajectory visualizations of the example motion and diverse motions.}
    \label{fig:diverse_case7_sub2}
    \end{subfigure}
    \caption{\textbf{Diverse generated results of blue example generated by the prompt ``\texttt{person walks in a semi-circular shape while swinging arms slightly.}''.} The example-based generated walking motions are diverse and similar to the source walking example.}
    \label{fig:diverse_case7}
\end{minipage}
\end{figure*}

We provide some visualization results to further illustrate the effectiveness of our approach in generating diverse motions that adhere closely to the given prompts. In \cref{fig:diverse_case5}, the example motion of ``\texttt{a person kicking their feet}'' is taken as the reference, and multiple diverse kick motions are generated. These generated motions not only exhibit variety but also maintain key characteristics of the original example.
Similarly, in \cref{fig:diverse_case7}, the example motion of ``\texttt{a person walking in a semi-circular shape while swinging arms slightly}'' demonstrates the capability to generate diverse walking motions that maintain the distinct features of the source motion. The generated trajectories, as visualized in \cref{fig:diverse_case5_sub2} and \cref{fig:diverse_case7_sub2}, show that the diverse motions follow different paths while retaining similarities with the original motion, confirming the effectiveness of our method.

\subsection{Inversion \textit{v.s.} Generation}

Our method can directly adopt DDIM inversion to edit the existing GT motions. Here, we compare the generation results with DDIM output. As shown in~\cref{tab:ddiminverse}, the inversion result of the generated motion is similar to the generated motion, supporting good performance of editing GT motions. 

\begin{table}[!h]
\centering
\setlength{\tabcolsep}{6pt}
\caption{Comparison between the inversed and generated motions.}
\resizebox{0.6\textwidth}{!}{
\begin{tabular}{cccccc}
\toprule
\multirow{2}{*}{Exp.} & \multicolumn{3}{c}{R-Precision$\uparrow$} & \multirow{2}{*}{FID$\downarrow$} \\
 \cmidrule(lr){2-4}
 & Top 1 & Top 2 & Top 3 &    \\
\midrule
generated & {0.544} & {0.732} & {0.831} & {0.269}   \\
generated + inversion & {0.535} & {0.724} & {0.818} & {0.299}   \\
\bottomrule
\end{tabular}
} 
\label{tab:ddiminverse}
\end{table}

\begin{table*}[b]

\begin{minipage}{0.48\textwidth}
\subsection{Results on Other Datasets}
We further evaluate our method on the KIT-ML dataset~\cite{kit}. As shown in \cref{tab:kit-results}, MotionCLR outperforms the state-of-the-art MoMask in key metrics. These results demonstrate the robustness and generalizability of MotionCLR, with consistent performance trends across datasets of different sizes.
\end{minipage}%
\hfill
\begin{minipage}[t]{0.48\textwidth}
\vspace{-4em}
\centering
\setlength{\tabcolsep}{3pt}
\captionof{table}{{\bf Comparison of results on KIT-ML.} MotionCLR demonstrates superior performance on key metrics.}
\resizebox{\textwidth}{!}{
\begin{tabular}{lcccccc}
\toprule
{Method} & {Top 1} & {Top 2} & {Top 3} & {FID} & {MM-Dist} & {Multi-Modality} \\
\midrule
MoMask (SOTA) & 0.433 & 0.656 & 0.781 & \textbf{0.204} & 2.779 & 1.131 \\
MotionCLR & \textbf{0.438} & \textbf{0.658} & \textbf{0.783} & 0.275 & \textbf{2.773} & \textbf{1.213} \\
\bottomrule
\end{tabular}
}
\label{tab:kit-results}
\end{minipage}
\end{table*}

\clearpage

\vspace{-2em}



\clearpage

\section{Web User Interface for Interactive Motion Generation and Editing}

\label{sec:interface}

\begin{figure}[!h]
    \centering
    \begin{subfigure}[b]{0.45\textwidth}
        \centering
        \includegraphics[width=0.9\linewidth]{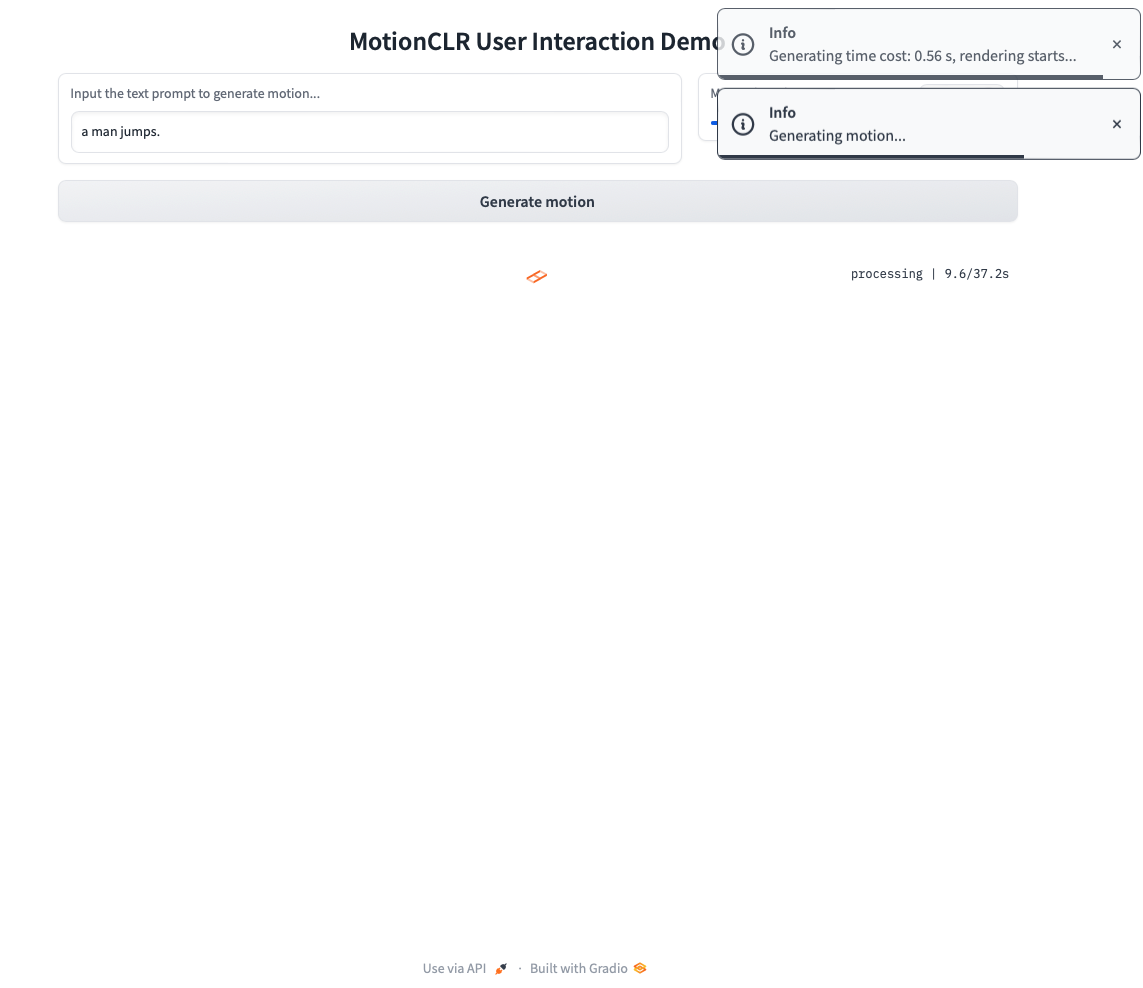}
        \caption{Motion generation interface example.}
        \label{fig:gen-1}
    \end{subfigure}
    \hfill
    \begin{subfigure}[b]{0.45\textwidth}
        \centering
        \includegraphics[width=0.9\linewidth]{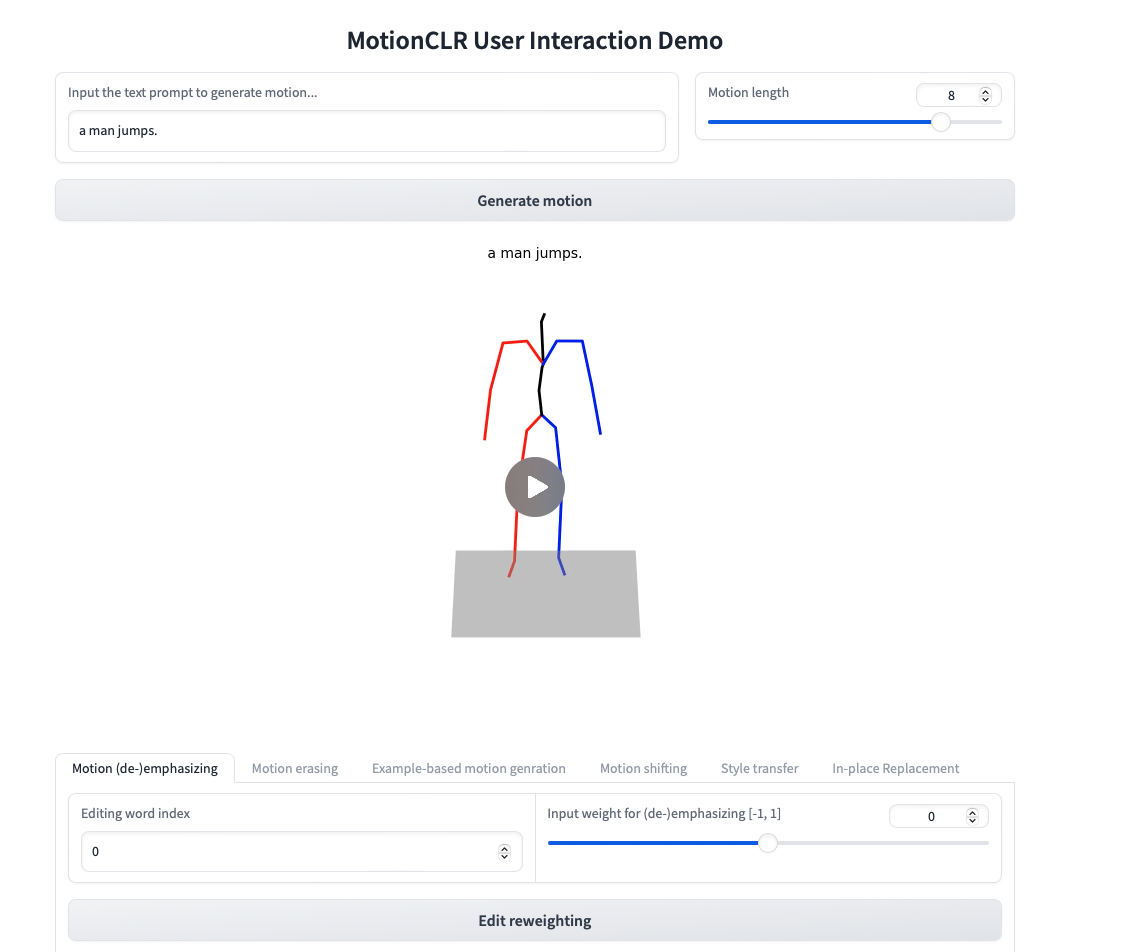}
        \caption{Generated Motion Example}
        \label{fig:gen-2}
    \end{subfigure}
    \caption{Motion generation and its output examples.}
    \label{fig:motion-generation}
\end{figure}

To have a better understanding of our task, we build a user interface with Gradio~\citep{gradio}. We introduce the demo as follows.
In~\cref{fig:motion-generation}, we illustrate the steps involved in generating and visualizing motions using the interactive interface. \cref{fig:gen-1} displays the initial step where the user provides input text such as ``a man jumps'' and adjusts motion parameters. Once the settings are finalized, the system begins processing the motion based on these inputs, as seen in the left panel. 
\cref{fig:gen-2} showcases the generated motion based on the user’s input. The interface provides a rendered output of the skeleton performing the described motion. This presentation allows users to easily correlate the input parameters with the resulting animation. The generated motion can further be edited by adjusting parameters such as the length of the motion, emphasizing or de-emphasizing certain actions, or replacing actions altogether, depending on user requirements. This process demonstrates how the interface facilitates a workflow from input to motion visualization.

\begin{figure*}[!h]
    \centering
    \begin{subfigure}[b]{0.48\textwidth}
        \centering
        \includegraphics[width=\textwidth]{images/demo-figs/app1.jpg}
        \caption{Motion (de-)Emphasizing interface.}
        \label{fig:app1}
    \end{subfigure}
    \hfill
    \begin{subfigure}[b]{0.48\textwidth}
        \centering
        \includegraphics[width=\textwidth]{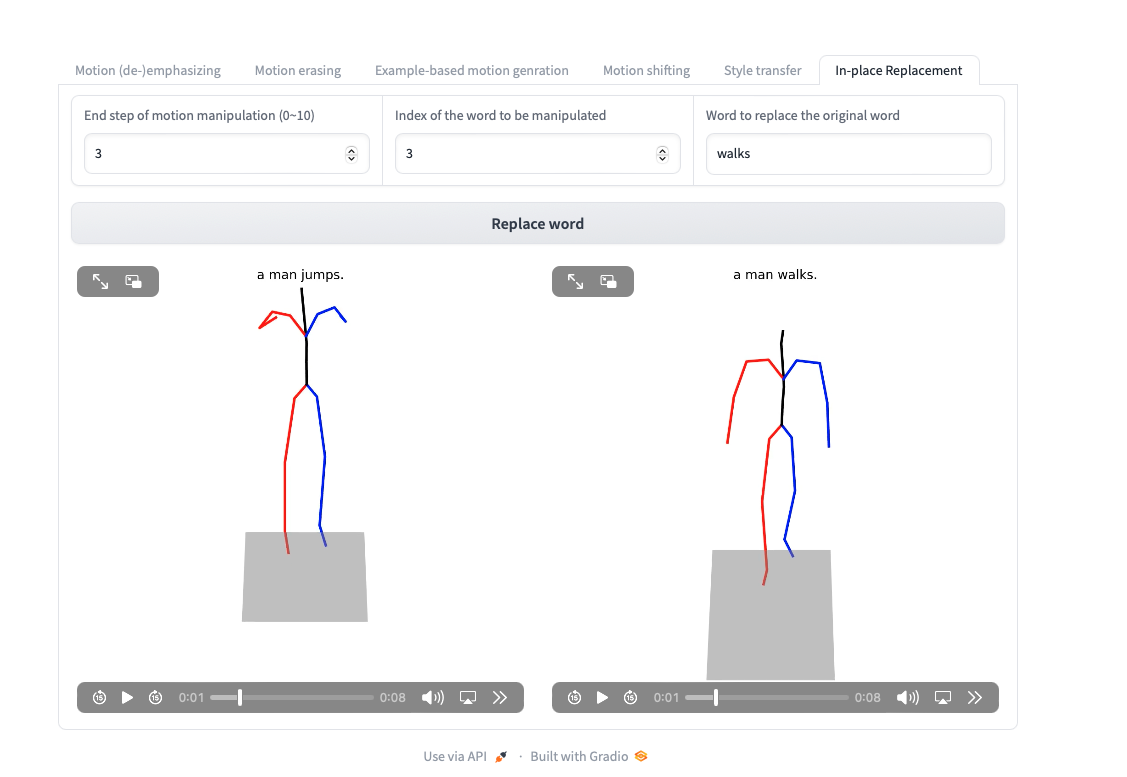}
        \caption{In-place replacement example.}
        \label{fig:app4}
    \end{subfigure}
    
    \vskip\baselineskip
    
    \begin{subfigure}[b]{0.48\textwidth}
        \centering
        \includegraphics[width=\textwidth]{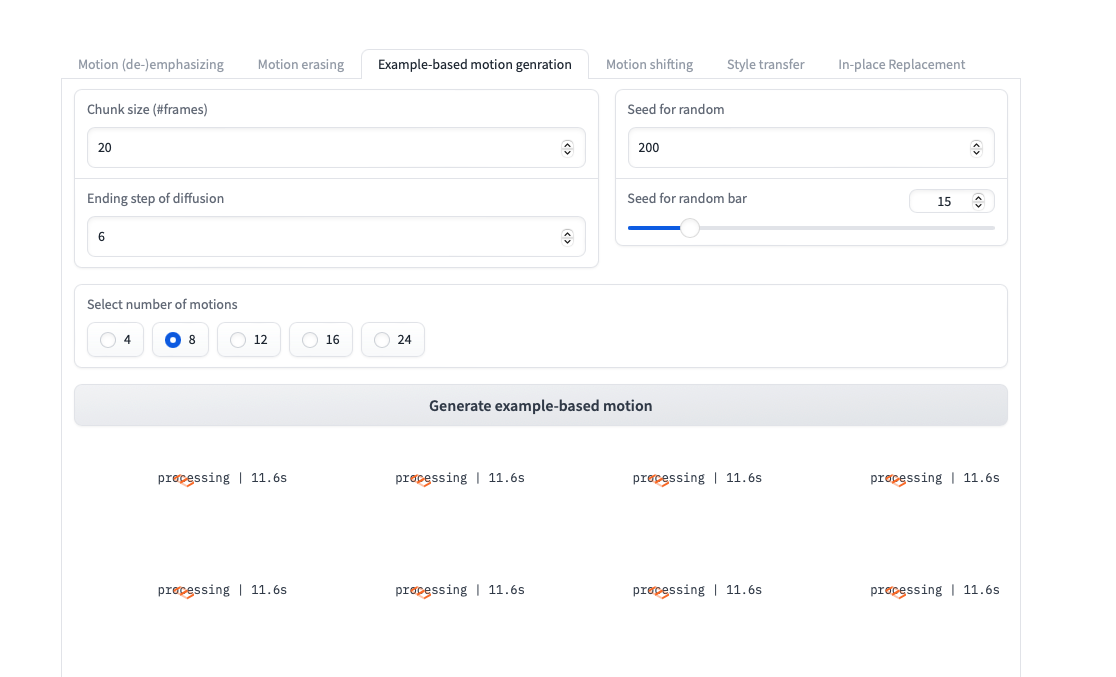}
        \caption{Example-based motion generation progress.}
        \label{fig:app2}
    \end{subfigure}
    \hfill
    \begin{subfigure}[b]{0.48\textwidth}
        \centering
        \includegraphics[width=\textwidth]{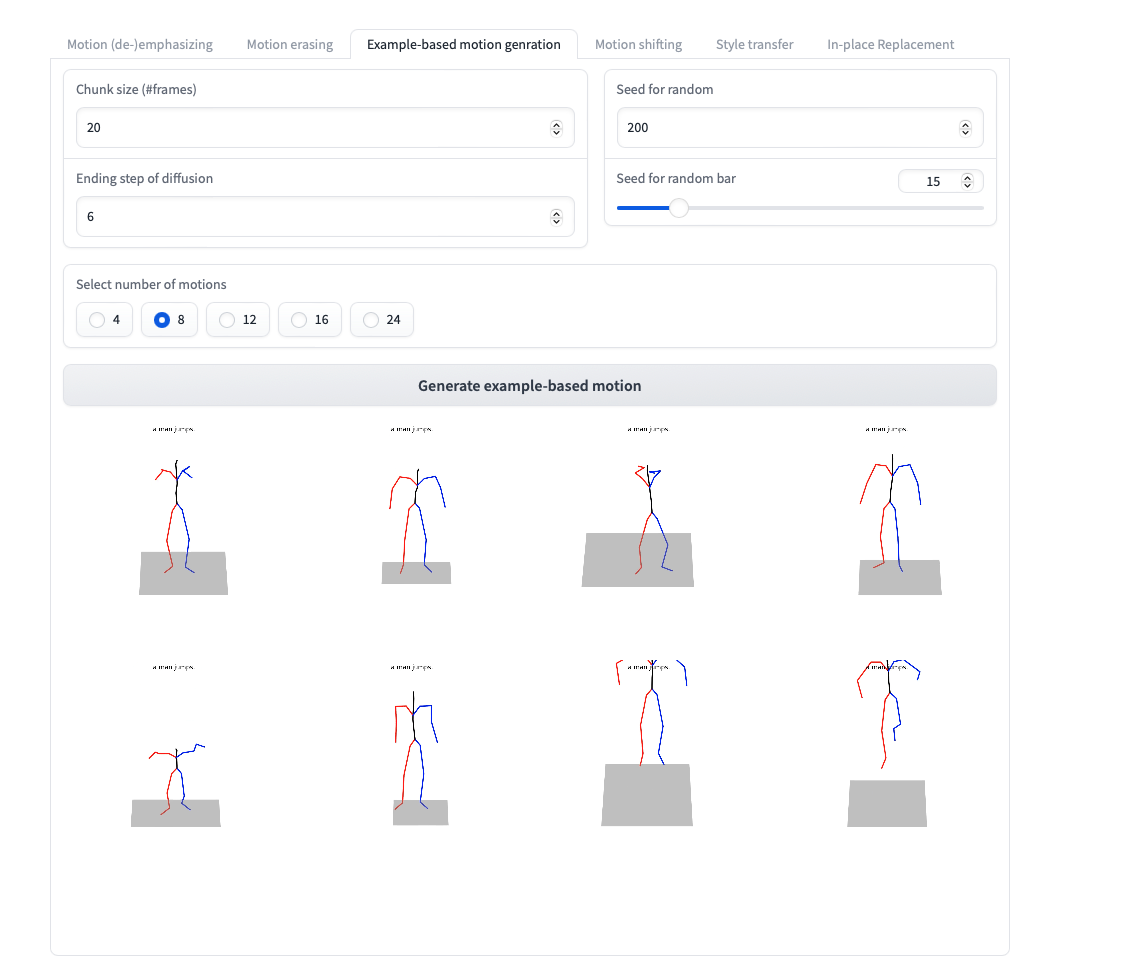}
        \caption{Example-based motion generation results. }
        \label{fig:app3}
    \end{subfigure}
    \caption{Different interfaces and supporting functions for interactive motion editing.}
    \label{fig:editing-demo}
\end{figure*}

The logical sequence of operations is as follows:
\begin{enumerate}[left=0pt]
    \item \textbf{Input the text:} Users start by entering text describing the motion (e.g., ``\texttt{a man jumps.}'') or set the frames of motions to generate (as shown in \cref{fig:gen-1}).
    
    \item \textbf{Generate the initial motion:} The system generates the corresponding skeleton motion sequence based on the input text (as shown in \cref{fig:gen-2}). 
    
    \item \textbf{Motion editing:} We show some downstream tasks of MotionCLR here. 
    \begin{itemize}[left=0pt]
        \item \textbf{Motion emphasizing/de-emphasizing:} Users can select a specific word from the text (e.g., ``\texttt{jumps}'') and adjust its emphasis using a weight slider (range [-1, 1]) (as seen in \cref{fig:app1}). For example, setting the weight to 0.3 will either increase the jump motion's intensity.
        \item \textbf{In-place replacement:} If users want to change the action, they can select the ``replace'' option. For example, replacing ``\texttt{jumps}'' with ``\texttt{walks}'' will regenerate the motion, showing a comparison between the original and new edited motions (as shown in \cref{fig:app4}).
        \item \textbf{Example-based motion generation:} Users can generate motion sequences based on predefined examples by setting parameters like chunk size and diffusion steps. After specifying the number of motions to generate, the system will create multiple variations of the input motion, providing diverse options for further refinement (as illustrated in~\cref{fig:app3}). The progress bars of the process are visualized in~\cref{fig:app2}.
    \end{itemize}
\end{enumerate}


\newpage

\clearpage

\section{Detailed Diagram of Attention Mechanisms}
\label{sec:detailedqkv}

\subsection{Mathematical Visualization of Self-attention Mechanism}

In the main text (\cref{eq:selfattn}), we introduced the self-attention mechanism of MotionCLR, which utilizes different transformations of motion as inputs. The motion embeddings serve as both the query ($\mathbf{\color[RGB]{246, 189, 137} Q}$), key ($\mathbf{\color[RGB]{249, 211, 179} K}$), and value ($\mathbf{\color[RGB]{246, 189, 137} V}$), capturing the internal relationships within the sequence of motion frames.
\cref{fig:appendix_selfattn} provides a detailed mathematical visualization of this process:
(1) \textbf{Similarity Calculation}. In the first step, the similarity between the motion embeddings at different frames is computed using the dot product, represented by $\mathbf{\color[RGB]{175, 175, 175} S} = \mathbf{{\color[RGB]{246, 189, 137} Q}{\color[RGB]{249, 211, 179} K}}^\top$. This measurement reflects the internal relationship/similarity between different motion frames within the sequence. \cref{fig:appendix_selfattn-a} illustrates how the $\mathtt{softmax}(\cdot)$ operation is applied to the similarity matrix to determine which motion feature should be selected at a given frame $f$.
(2) \textbf{Feature Updating}. Next, the similarity scores are used to weight the motion embeddings ($\mathbf{\color[RGB]{246, 189, 137} V}$) and generate updated features $\mathbf{\color[RGB]{236, 182, 131}X'}$, as shown by the equation $\mathbf{\color[RGB]{236, 182, 131}X'} = \mathtt{softmax}(\mathbf{{\color[RGB]{246, 189, 137} Q}{\color[RGB]{249, 211, 179} K}}^\top/\sqrt{d})\mathbf{\color[RGB]{246, 189, 137} V}$. Here, the similarity matrix applies its selection of values ($\mathbf{\color[RGB]{246, 189, 137} V}$) to update the motion features. This process allows the self-attention mechanism to dynamically adjust the representation of each motion frame based on its relevance to other frames in the sequence.

In summary, the self-attention mechanism aims to identify and emphasize the most relevant motion frames in the sequence, updating the features to enhance their representational capacity for downstream tasks. The most essential capability of cross-attention is to order the motion features. 


\subsection{Mathematical Visualization of Cross-attention Mechanism}

In the main text (\cref{eq:crossattn}), we introduced the cross-attention mechanism of MotionCLR, which utilizes the transformation of motion as a query ($\mathbf{{\color[RGB]{255, 153, 63} Q}}$) and the transformation of text as a key ($\mathbf{{\color[RGB]{242, 158, 168} K}}$) and value ($\mathbf{{\color[RGB]{179, 221, 148} V}}$) to explicitly model the correspondence between motion frames and words.

\begin{figure*}[t]
\begin{minipage}{\textwidth}
    \centering
    \begin{subfigure}[b]{0.48\textwidth}
    \centering
    \includegraphics[width=0.99\linewidth]{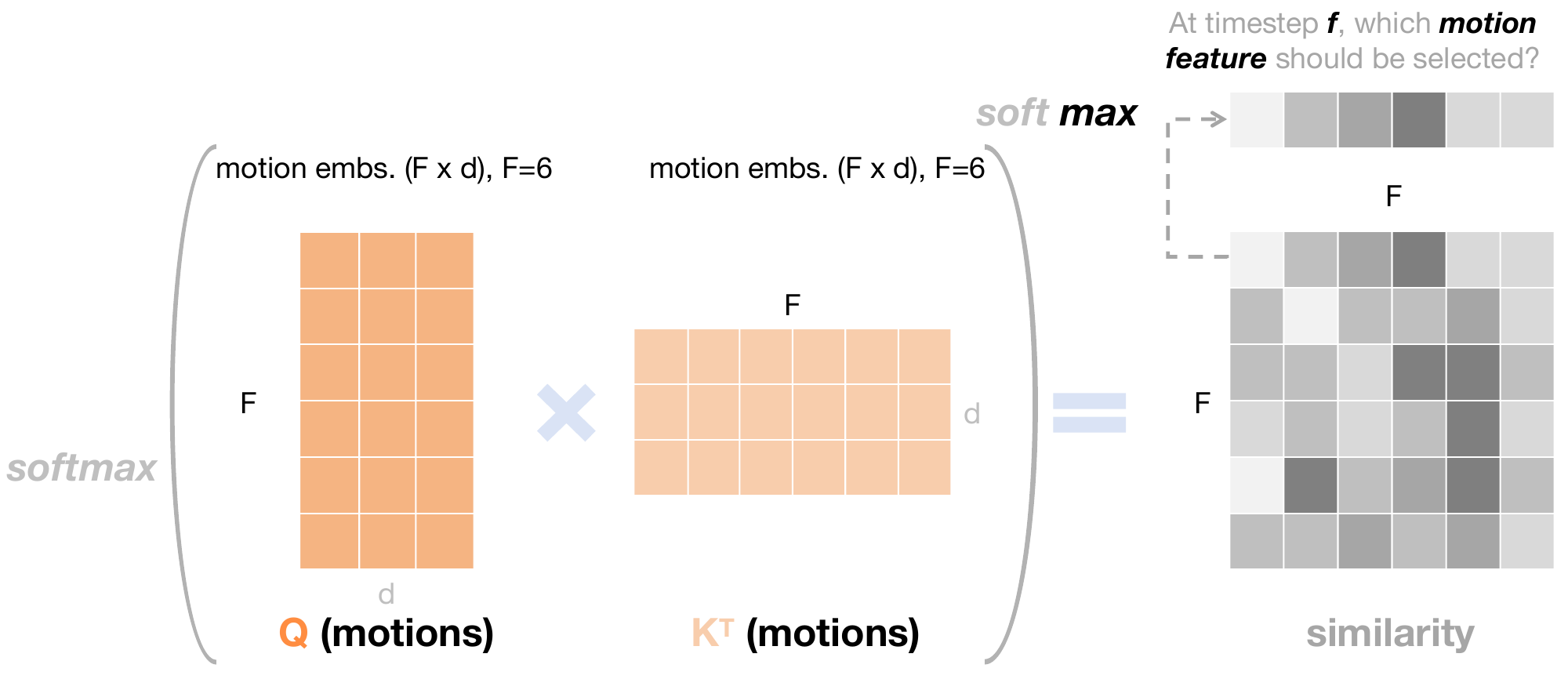}
    \caption{$\mathbf{\color[RGB]{175, 175, 175} S} = \mathbf{{\color[RGB]{246, 189, 137} Q}{\color[RGB]{249, 211, 179} K}}^\top $.}
    \label{fig:appendix_selfattn-a}
    \end{subfigure}
    \begin{subfigure}[b]{0.48\textwidth}
    \centering
    \includegraphics[width=0.99\linewidth]{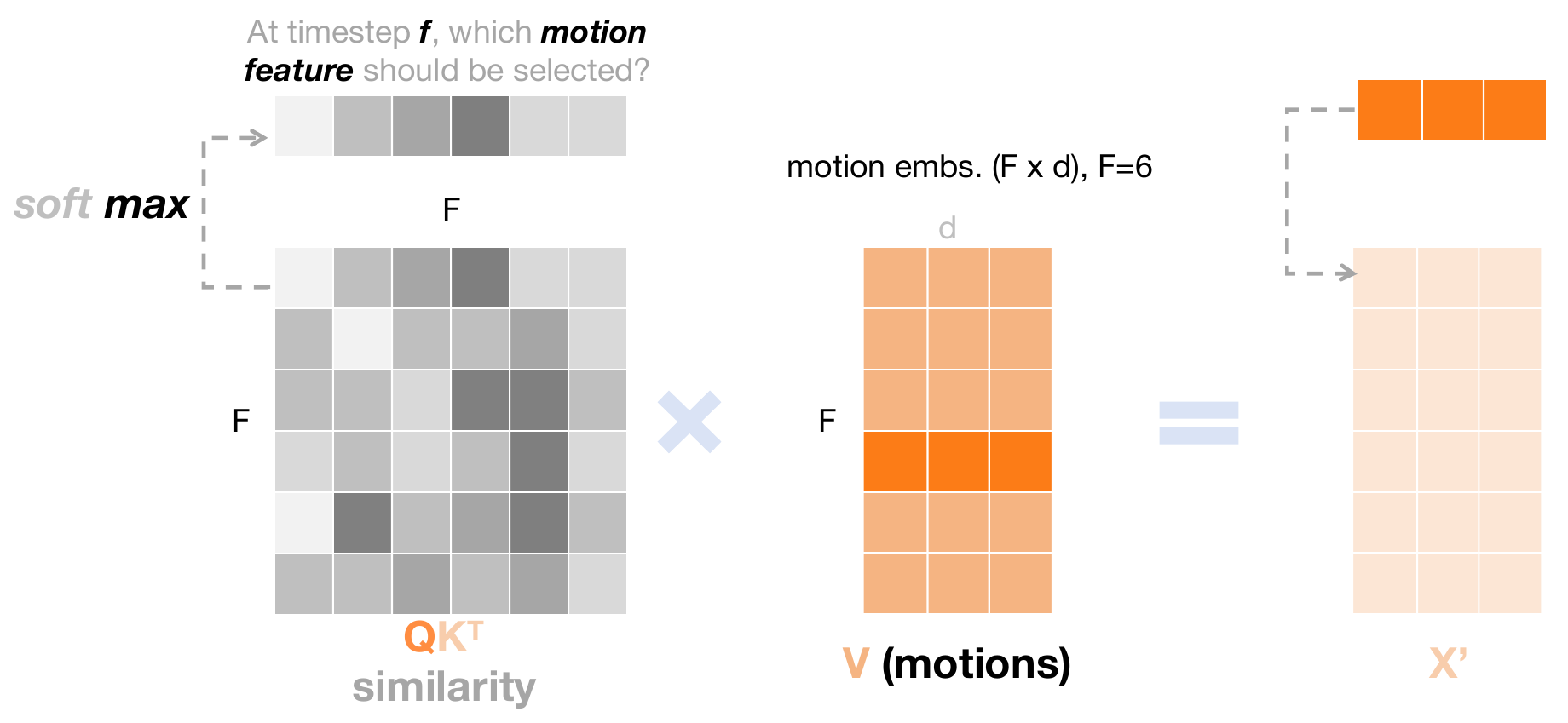}
    \caption{$\mathbf{\color[RGB]{236, 182, 131}X'} = \mathtt{softmax}(\mathbf{\color[RGB]{175, 175, 175} S} /\sqrt{d}) \mathbf{\color[RGB]{246, 189, 137} V}$.}
    \label{fig:appendix_selfattn-b}
    \end{subfigure}
    \caption{\textbf{Mathematical Visualization of Self-attention Mechanism.} This figure takes $F=6$ as an example. (a) The similarity calculation with queries and keys (different frames). (b) The similarity matrix picks ``value''s of the attention mechanism and updates motion features. }
    \vspace{2em}
    \label{fig:appendix_selfattn}
\end{minipage}
\begin{minipage}{\textwidth}
    \centering
    \begin{subfigure}[b]{0.48\textwidth}
    \centering
    \includegraphics[width=0.99\linewidth]{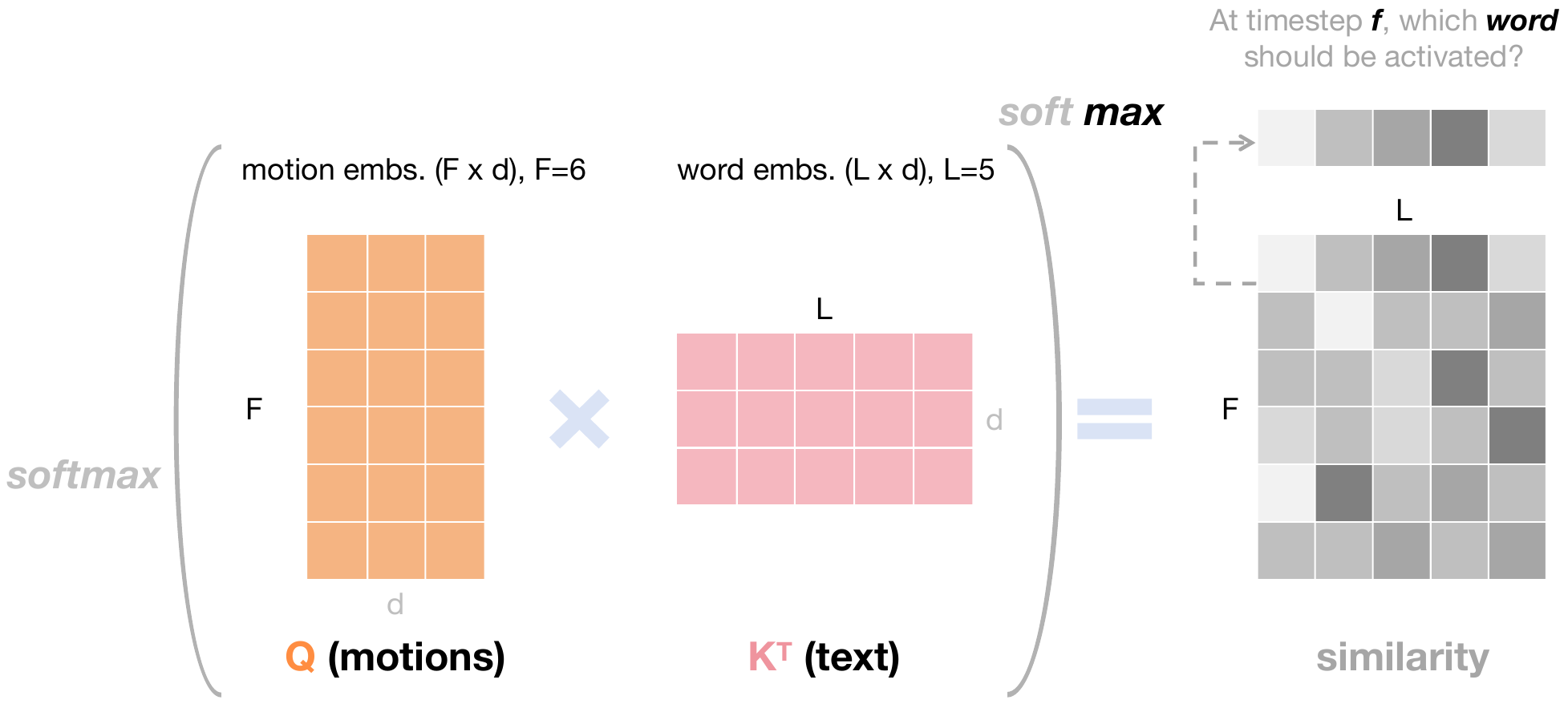}
    \caption{$\mathbf{\color[RGB]{175, 175, 175} S} = \mathbf{{\color[RGB]{255, 153, 63} Q}{\color[RGB]{242, 158, 168} K}}^\top $.}
    \label{fig:appendix_crossattn-a}
    \end{subfigure}
    \hspace{1em}
    \begin{subfigure}[b]{0.48\textwidth}
    \centering
    \includegraphics[width=0.99\linewidth]{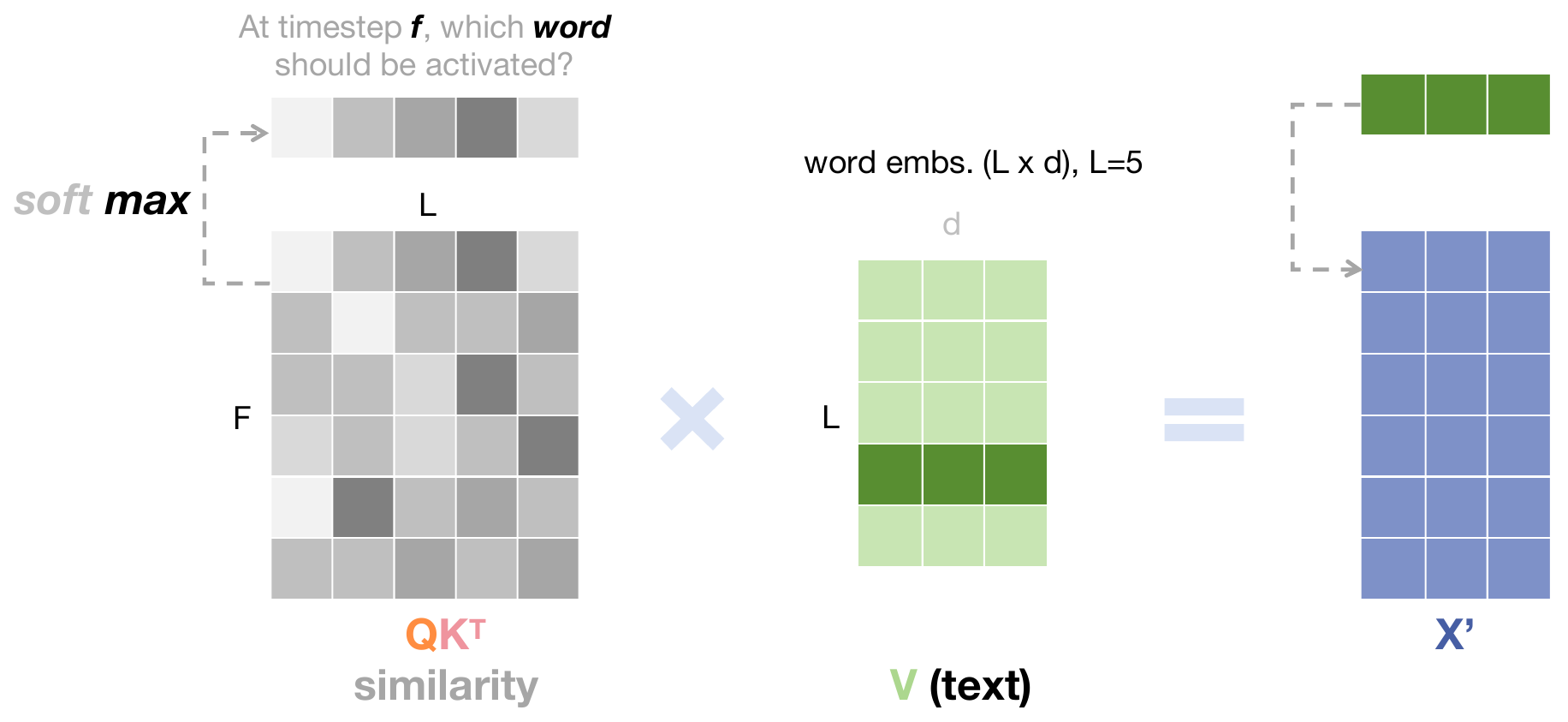}
    \caption{$\mathbf{\color[RGB]{83, 104, 177}X'} = \mathtt{softmax}(\mathbf{\color[RGB]{175, 175, 175} S}/\sqrt{d}) \mathbf{{\color[RGB]{179, 221, 148} V}}$.}
    \label{fig:appendix_crossattn-b}
    \end{subfigure}
    \caption{\textbf{Mathematical Visualization of Cross-attention Mechanism.} This figure takes $F=6$ and $N=5$ as an example. (a) The similarity calculation with queries and keys. (b) The similarity matrix picks ``value''s of the attention mechanism and updates features. }
    \label{fig:appendix_crossattn}
\end{minipage}
\end{figure*}

\cref{fig:appendix_crossattn} provides a detailed mathematical visualization of this process:

(1) \textbf{Similarity Calculation}. In the first step, the similarity between the motion embeddings ($\mathbf{{\color[RGB]{255, 153, 63} Q}}$) with $F$ frames and the text embeddings ($\mathbf{{\color[RGB]{242, 158, 168} K}}$) with $N$ words is computed through the dot product, represented by $\mathbf{\color[RGB]{175, 175, 175} S} = \mathbf{{\color[RGB]{255, 153, 63} Q}{\color[RGB]{242, 158, 168} K}}^\top$. This similarity measurement reflects the relationship between motion frames and words. \cref{fig:appendix_crossattn-a} shows how the $\mathtt{softmax}(\cdot)$ operation is applied to the similarity matrix to determine which word should be activated at a given frame $f$. 

(2) \textbf{Feature Updating}. Next, the similarity scores are used to weight the text embeddings ($\mathbf{{\color[RGB]{179, 221, 148} V}}$) and generate updated features $\mathbf{\color[RGB]{83, 104, 177}X'}$, as shown by the equation $\mathbf{\color[RGB]{83, 104, 177}X'} = \mathtt{softmax}(\mathbf{{\color[RGB]{255, 153, 63} Q}{\color[RGB]{242, 158, 168} K}}^\top/\sqrt{d})\mathbf{{\color[RGB]{179, 221, 148} V}}$. Here, the similarity matrix applies its selection of values ($\mathbf{{\color[RGB]{179, 221, 148} V}}$) to update the features. This process establishes an explicit correspondence between the frames and specific words. 

In summary, the similarity calculation process determines which frame(s) should be selected, and the feature updating process (multiplication with ${\bf V}$) is the execution of the frame(s) placement.

\subsection{The Basic Difference with Previous Diffusion-based Motion Generation Models in Cross-modal Modeling}
\label{sec:compare_other}

As discussed in the main text (see~\cref{sec:intro}), despite the progresses in human motion generation~\citep{zhang2024finemogen,cai2024digital,zhang2024large,guo2024crowdmogen,raabsingle,kapon2024mas,cohan2024flexible,fan2024freemotion,xu2023interdiff,xu2023stochastic,yao2022controlvae,feng2023robust,ao2023gesturediffuclip,yao2024moconvq,zhang2024semantic,liu2010sampling,aberman2020skeleton,karunratanakul2024optimizing,li2023controllable,li2023object,gong2023tm2d,zhou2023ude,zhong2023attt2m,athanasiou2023sinc,zhong2024smoodi,guo2024generative,zhang2024large,zhao2023synthesizing,zhang2022egobody,zhang2020perpetual,diomataris2024wandr,pinyoanuntapong2024mmm,diller2024cg,peng2023hoi,hou2023compositional,liu2023interactive,cong2024laserhuman,jiang2022chairs,kulkarni2024nifty,tessler2024maskedmimic,liang2024intergen,ghosh2023remos,wu2024thor}, there still lacks a explicit modeling of word-level cross-modal correspondence in previous work. To clarify this, our method models a fine-grained word-level cross-modal correspondence.

\begin{figure*}
    \centering
    \includegraphics[width=\textwidth]{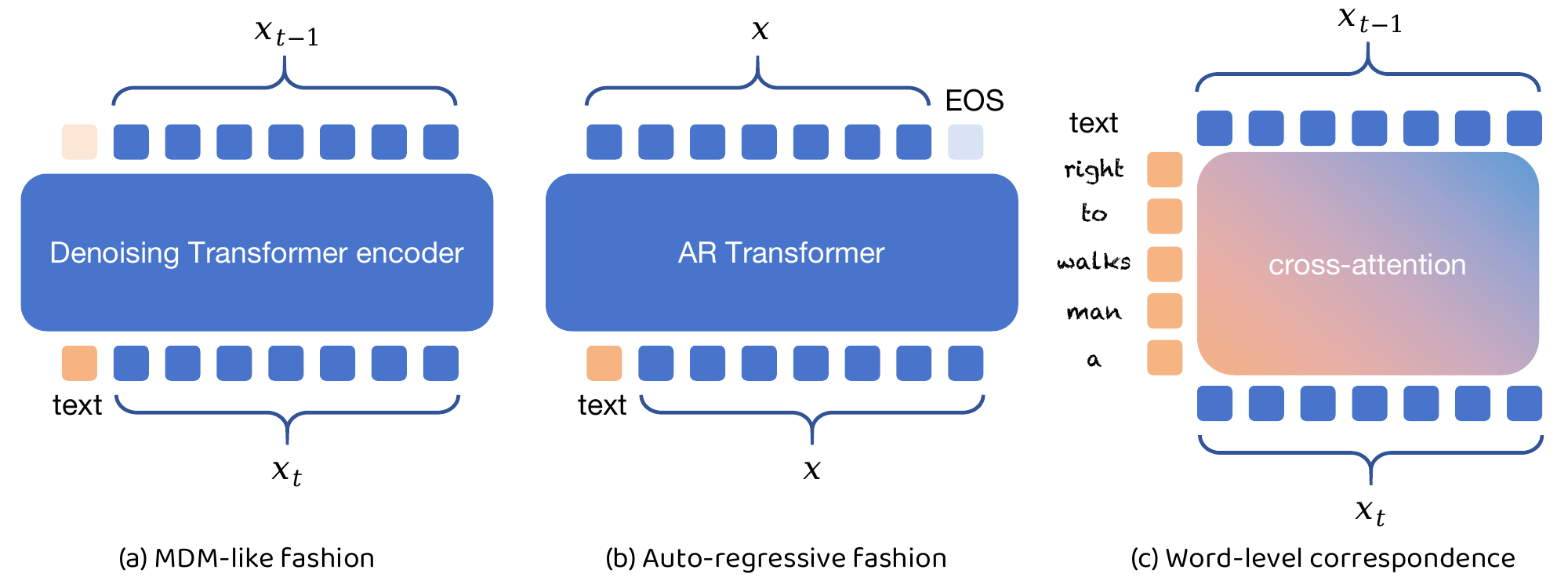}
    \vspace{-2em}
    \caption{\textbf{Comparison with previous diffusion-based motion generation models.} (a) MDM-like fashion: \citet{mdm} and its follow-up methods treat text embeddings as a whole and mix them with motion representations using a denoising Transformer. (b) Auto-regressive fashion: \citet{t2mgpt} and its follow-up methods concatenate the text with the motion sequence and feed them into an auto-regressive transformer without explicit correspondence modeling. (c) Our proposed method establishes fine-grained word-level correspondence using cross-attention mechanisms.}
    \label{fig:difference-cross}
\end{figure*}

As illustrated in \cref{fig:difference-cross}, the major distinction between our proposed method and previous diffusion-based motion generation models lies in the explicit modeling of word-level cross-modal correspondence. In the MDM-like fashion~\cite{mdm} (see \cref{fig:difference-cross}a), previous methods usually utilize a denoising transformer encoder that treats the entire text as a single embedding, mixing it with the motion sequence. This approach lacks the ability to capture the nuanced relationship between individual words and corresponding motion elements, resulting in an over-compressed representation. Although we witness that \cite{motiondiffuse} also introduces cross-attention in the motion generation process, it still faces two problems in restricting the fine-grained motion editing applications. First of all, the text embeddings are mixed with frame embeddings of diffusion, resulting in a loss of detailed semantic control. Our approach disentangles the diffusion timestep injection process in the convolution module to resolve this issue. Besides, the linear cross-attention in MotionDiffuse is different from the computation process of cross-attention, resulting in a lack of explanation of the word-level correspondence. The auto-regressive (AR) fashion~\citep{t2mgpt} (\cref{fig:difference-cross}b) adopts a simple concatenation of text and motion, where an AR transformer processes them together. However, this fashion also fails to explicitly establish a fine-grained correspondence between text and motion, as the AR transformer merely regards the text and motion embeddings as one unified sequence. 

Our approach (shown in \cref{fig:difference-cross}c) introduces a cross-attention mechanism that explicitly captures the word-level correspondence between the input text and generated motion sequences. This allows our model to maintain a clear and interpretable mapping between specific words and corresponding motion patterns, significantly improving the quality and alignment of generated motions with the textual descriptions.
By integrating such a word-level cross-modal modeling technique, our method not only achieves more accurate and realistic motion generation but also supports fine-grained word-level motion editing. This capability enables users to make precise adjustments to specific parts of the generated motion based on textual prompts, addressing the critical limitations present in previous diffusion-based motion generation models and allowing for more controllable and interpretable editing at the word level.

\clearpage

\newpage

\section{More Visualization Results of Empirical Evidence}
\label{sec:moreemp}

\begin{figure}[!ht]
    \centering
    \includegraphics[width=\linewidth]{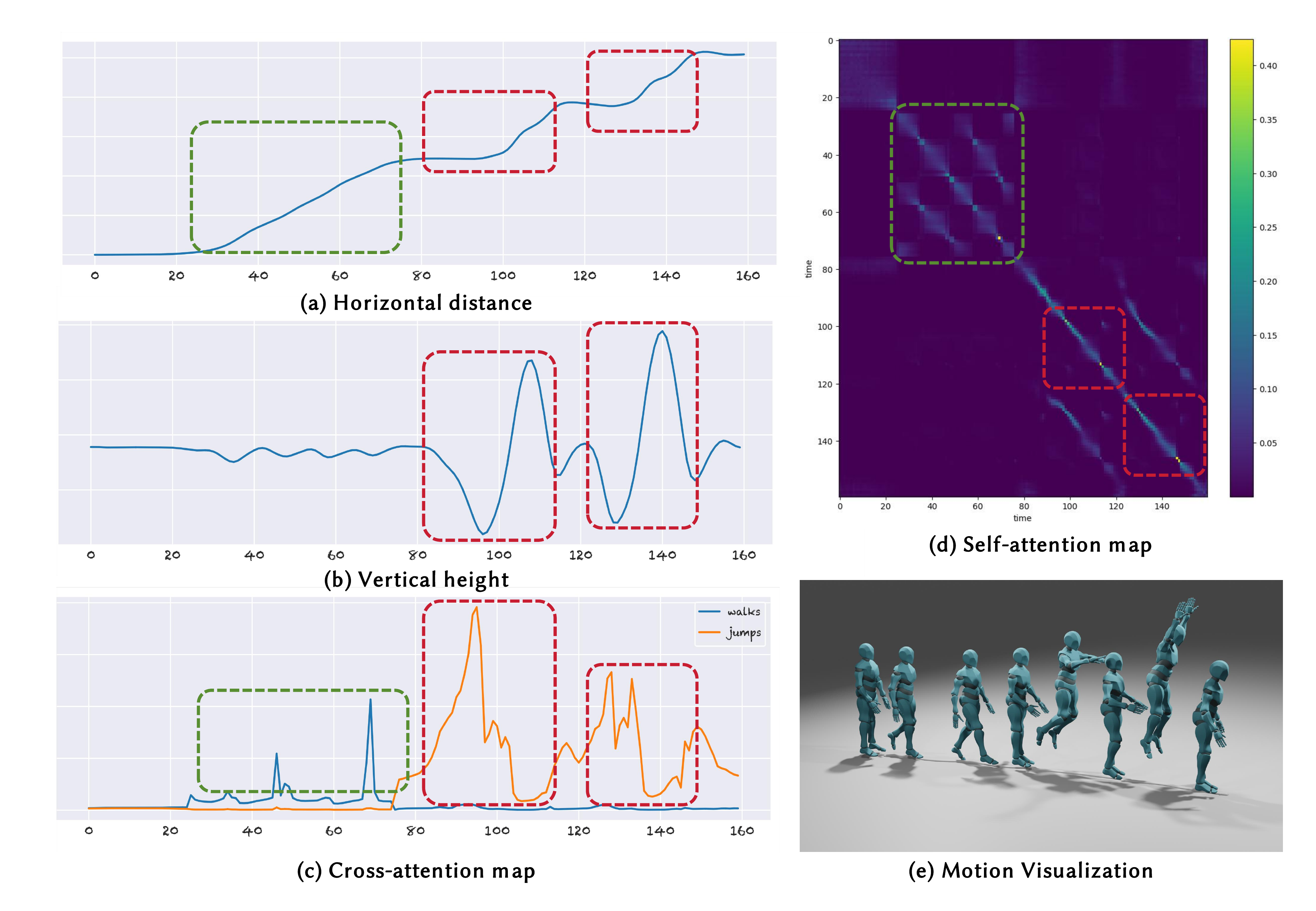}
    \captionsetup{font=small}
    \vspace{-1.9em}
    \caption{\textbf{Empirical study of attention patterns.} We use the example ``\texttt{a person walks stop and then jumps.}'' (a) Horizontal distance traveled by the person over time, highlighting distinct walking and jumping phases. (b) The vertical height changes of the person, indicating variations during walking and jumping actions. (c) The \textbf{cross-attention} map between timesteps and the described actions. Notice that ``\texttt{walk}'' and ``\texttt{jump}'' receive a stronger attention signal corresponding to the walk and jump segments. (d) The \textbf{self-attention} map, which clearly identifies repeated walking and jumping cycles, shows similar patterns in the sub-actions. (e) Visualization of the motion sequences, demonstrating the walking and jumping actions.}
    \label{fig:study_attn_app1}
\end{figure}

In the main text, we introduced the foundational understanding of both cross-attention and self-attention mechanisms, emphasizing their ability to capture temporal relationships and dependencies across motion sequences. As a supplement, we provide a new, more detailed example here. As shown in~\cref{fig:study_attn_app1}, this visualization illustrates how different attention mechanisms respond to a complex sequence involving both walking and jumping actions. Specifically, we use green dashed boxes to highlight the ``\texttt{walk}'' phases and red dashed boxes to indicate the ``\texttt{jump}'' phases. This allows us to clearly distinguish the temporal patterns associated with each action. Besides, we observed that the word ``\texttt{jump}'' reaches its highest activation during the crouching phase, which likely correlates with this moment being both the start of the jumping action and the ``power accumulation phase''. This suggests that the attention mechanism accurately captures the preparatory stage of the movement, highlighting its capability to recognize the nuances of motion initiation within complex sequences. The cross-attention map effectively aligns key action words like ``\texttt{walk}'' and ``\texttt{jump}'' with their respective motion segments, while the self-attention map reveals repeated motion patterns and similarities between the walking and jumping cycles.

\begin{figure}[!ht]
\centering
\includegraphics[width=\linewidth]{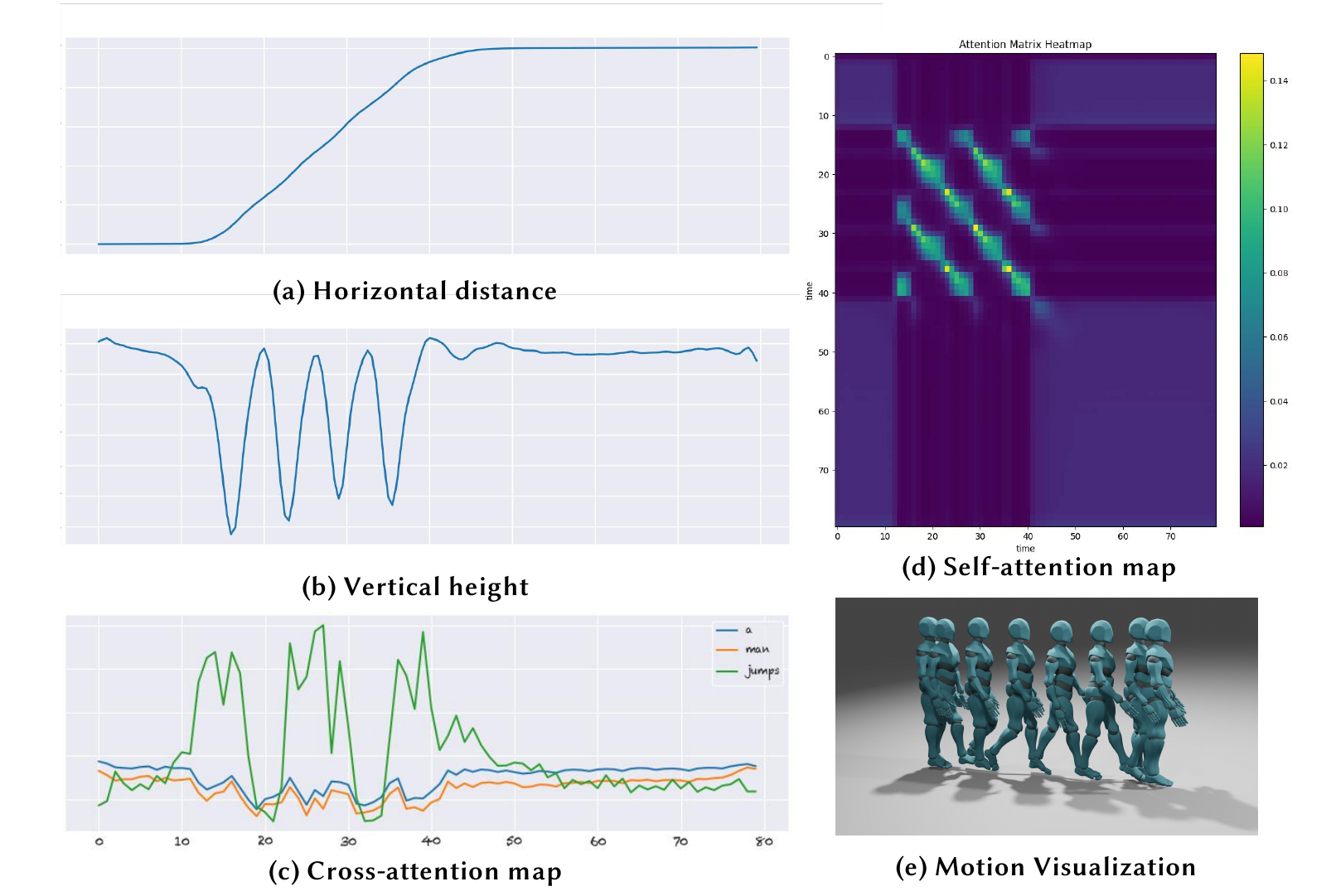}
\captionsetup{font=small}
\vspace{-1.9em}
\caption{\textbf{Empirical study of attention patterns in a consistent walking sequence.} We use the example: ``\texttt{a man walks.}''. (a) The horizontal distance traveled over time reflects a steady walking motion. (b) Vertical height changes indicate minimal variation, characteristic of walking actions. (c) The \textbf{cross-attention} map shows that the ``\texttt{walks}'' word maintains consistent activation. (d) The \textbf{self-attention} map highlights the repeated walking cycles, capturing the temporal stability. (e) Visualization of the motion sequence.}
\label{fig:study_attn_app2}
\end{figure}

Continuing with another case study, in~\cref{fig:study_attn_app2}, we examine how attention mechanisms respond to a sequence that primarily involves walking actions with varying intensity. 
In this instance, we observe that both the horizontal distance (\cref{fig:study_attn_app2}a) and vertical height (\cref{fig:study_attn_app2}b) reflect the man walks straight. The cross-attention map (\cref{fig:study_attn_app2}c) reveals how the word ``walks'' related to walking maintains consistent activation, indicating that MotionCLR has a word-level understanding throughout the sequence. The self-attention map (\cref{fig:study_attn_app2}d) further emphasizes repeated walking patterns, demonstrating that the mechanism effectively identifies the temporal consistency and structure of the walking phases. The motion visualization (\cref{fig:study_attn_app2}e) reinforces this finding, showing a clear, uninterrupted walking motion.

More importantly, we can observe that the walking action consists of a total of five steps: three steps with the right foot and two with the left foot. The self-attention map (\cref{fig:study_attn_app2}d) clearly reveals that steps taken by the same foot exhibit similar patterns, while movements between different feet show distinct differences. This observation indicates that the self-attention mechanism effectively captures the subtle variations between repetitive motions, further demonstrating its sensitivity to nuanced motion capture capability within the sequence.

Besides, different from the jumping, the highlights in the self-attention map of the walking are rectangular. The reason is that the local movements of walking are similar. In contrast, the jumping includes several sub-actions, resulting in the highlighted areas in the self-attention maps being elongated. 

\clearpage
\section{Implementation and Evaluation Details}

\subsection{Compared Baselines}

Here, we introduce details of baselines in~\cref{tab:humanml3d_main} for our comparison. 



\textbf{MDM}~\citep{mdm} uses a diffusion-based approach with a transformer-based design for generating human motions. It excels at handling various generation tasks, achieving satisfying results in text-to-motion tasks.

\textbf{MLD}~\citep{mld} uses a diffusion process on motion latent space for conditional human motion generation. By employing a Variational AutoEncoder (VAE), it efficiently generates vivid motion sequences while reducing computational overhead.

\textbf{MotionDiffuse}~\citep{motiondiffuse} is a diffusion model-based text-driven framework for motion generation. It provides diverse and fine-grained human motions, supporting probabilistic mapping and multi-level manipulation based on text prompts.


\textbf{ReMoDiffuse}~\citep{remodiffuse} integrates retrieval mechanisms into a diffusion model for motion generation, enhancing diversity and consistency. It uses a Semantic-Modulated Transformer to incorporate retrieval knowledge, improving text-motion alignment.

\textbf{MoMask}~\citep{momask} introduces a masked modeling framework for 3D human motion generation using hierarchical quantization. It outperforms other methods in generating motions and is applicable to related tasks without further fine-tuning.

\subsection{Evaluation Details}

\label{sec:app_evaluation}

\textbf{Motion (de-)emphasis.} To evaluate the effectiveness of motion (de-)emphasis application, we construct 200 prompts (\aka HVerb-wild)to verify the algorithm. All of these prompts are constructed by researchers manually. We take some samples from our evaluation set as follows. 
\begin{lstlisting}[basicstyle=\ttfamily\small, frame=single, keywordstyle={}, numbers=none, frame=none]
... ...
3 the figure leaps high
4 a man is waving hands
... ...
\end{lstlisting}
Each line in the examples represents the index of the edited word in the sentence, followed by the corresponding prompt. These indices indicate the key verbs or actions that are subject to the (de-)emphasis during the evaluation process. The prompts were carefully selected to cover a diverse range of actions, ensuring that our method is tested on different types of motion descriptions. For instance, in the prompt ``\texttt{3 the figure leaps high}'', the number 3 indicates that the word ``\texttt{leaps}'' is the third word in the sentence and is the target action for (de-)emphasis. This format ensures a systematic evaluation of how the model responds to adjusting attention weights on specific actions across different prompts.

\textbf{Example-based motion generation.} To further evaluate our example-based motion generation algorithm, we randomly constructed 7 test prompts. We used t-SNE~\citep{scikitlearn} visualization to analyze how closely the generated motions resemble the provided examples in terms of motion textures. For each case, the generated motion was assessed based on two criteria: (1) similarity to the example, and (2) diversity across different generated results from the same example.

\textbf{Action counting.} To thoroughly evaluate the effectiveness of our action counting method, we constructed a test set based on HVerb-wild. These prompts were manually designed by researchers to ensure diversity. Each prompt corresponds to a motion sequence generated by our model, and the ground truth action counts were labeled by researchers based on the observable actions within the generated motions.

\subsection{User Study Details}

\begin{figure}[h]
\centering
\includegraphics[width=\linewidth]{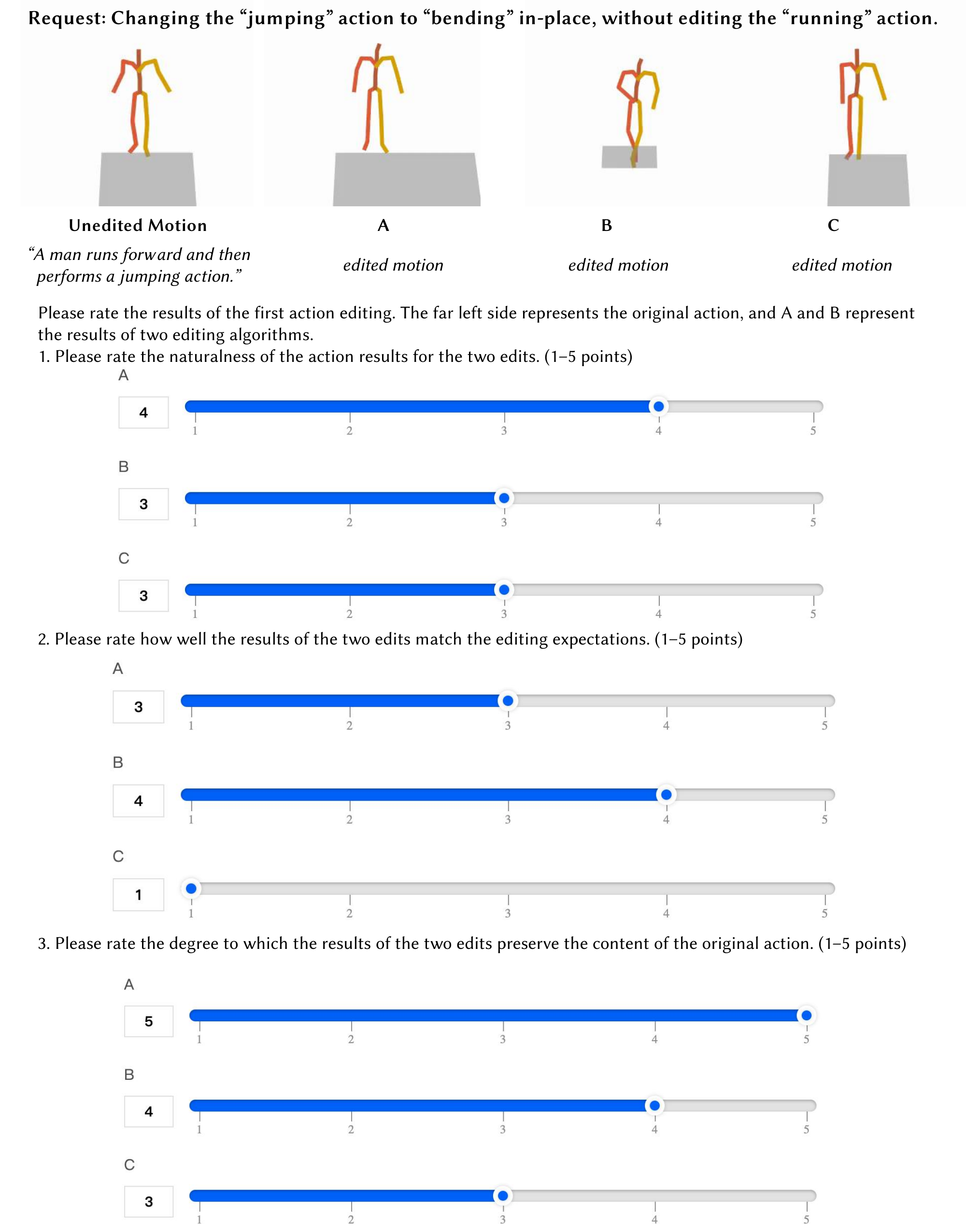}
\caption{An example interface from the user study.}
\label{fig:user-study-example}
\end{figure}

To evaluate the effectiveness of our editing methods, we conducted a user study with participants rating the quality of edited motion results. \cref{fig:user-study-example} illustrates an example of the interface and questions used during the study (replacement case for example here). Participants were asked to compare the original motion (leftmost column) with the results of 3 editing methods and rate them across three metrics (mentioned in the main text).
This study provides insight into the subjective perception of motion quality and highlights the differences between various editing approaches.

\newpage
\section{Details of Motion Editing}

\label{sec:app_detial_edit}

\vspace{-0.8em}
In this section, we will provide more technical details about the motion editing algorithms. 
\vspace{-0.8em}

\subsection{Pseudo Codes of Motion Editing}

\vspace{-0.3em}

\textbf{Motion (de-)emphasizing.} Motion (de-)emphasizing mainly manipulate the cross-attention weights of the attention map. Key codes are shown in the \texttt{L16-18} of Code~\ref{lst:motion_emphasizing}.

\setlength{\TPHorizModule}{1em} 
\setlength{\TPVertModule}{1em}

\begin{textblock}{10}(10.7,30.7) 
\begin{tikzpicture}
    \node[fill=gray!20, text opacity=1, fill opacity=0.5, text width=13.8cm, minimum height=5.5em, align=left] at (0,0) { };
\end{tikzpicture}
\end{textblock}

\lstset{style=pythonstyle}
\begin{lstlisting}[caption={Pseudo codes for motion (de-)emphasizing.}, label={lst:motion_emphasizing}]
def forward(self, x, cond, reweighting_attn, idxs):
    B, T, D = x.shape
    N = cond.shape[1]
    H = self.num_head
    
    # B, T, 1, D
    query = self.query(self.norm(x)).unsqueeze(2).view(B, T, H, -1)
    # B, 1, N, D
    key = self.key(self.text_norm(cond)).unsqueeze(1).view(B, N, H, -1)
    
    # B, T, N, H
    attention = torch.einsum('bnhd,bmhd->bnmh', query, key) / math.sqrt(D // H)
    weight = self.dropout(F.softmax(attention, dim=2))

    # reweighting attention for motion (de-)emphasizing
    if reweighting_attn > 1e-5 or reweighting_attn < -1e-5:
        for i in range(len(idxs)):
            weight[i, :, 1 + idxs[i]] = weight[i, :, 1 + idxs[i]] + reweighting_attn
        
    value = self.value(self.text_norm(cond)).view(B, N, H, -1)
    y = torch.einsum('bnmh,bmhd->bnhd', weight, value).reshape(B, T, D)
    return y
\end{lstlisting}

\vspace{-1.5em}

\textbf{In-place motion replacement.} The generation of two motions (\texttt{B=2}) are reference and edited motions. As the cross-attention map determines when to execute the action. Therefore, replacing the cross-attention map directly is a straightforward way, which is shown in \texttt{L16-17} of Code~\ref{lst:replace}.

\begin{textblock}{10}(10.7,62) 
\begin{tikzpicture}
    \node[fill=gray!20, text opacity=1, fill opacity=0.5, text width=13.8cm, minimum height=3.5em, align=left] at (0,0) { };
\end{tikzpicture}
\end{textblock}

\lstset{style=pythonstyle}
\begin{lstlisting}[caption={Pseudo codes for in-place motion replacement.}, label={lst:replace}]
def forward(self, x, cond, manipulation_steps_end):
    B, T, D = x.shape
    N = cond.shape[1]
    H = self.num_head
    
    # B, T, 1, D
    query = self.query(self.norm(x)).unsqueeze(2).view(B, T, H, -1)
    # B, 1, N, D
    key = self.key(self.text_norm(cond)).unsqueeze(1).view(B, N, H, -1)
    
    # B, T, N, H
    attention = torch.einsum('bnhd,bmhd->bnmh', query, key) / math.sqrt(D // H)
    weight = self.dropout(F.softmax(attention, dim=2))

    # replacing the attention map directly 
    if self.step <= manipulation_steps_end:
        weight[1, :, :, :] = weight[0, :, :, :]

    value = self.value(self.text_norm(cond)).view(B, N, H, -1)
    y = torch.einsum('bnmh,bmhd->bnhd', weight, value).reshape(B, T, D)
    return y
\end{lstlisting}

\vspace{-1.5em}

\newpage

\textbf{Motion sequence shifting.} Motion sequence shifting aims to correct the atomic motion in the temporal order you want. We only need to shift the temporal order of $\mathbf{Q}$s, $\mathbf{K}$s, and $\mathbf{V}$s in the self-attention to obtain the shifted result. Key codes are shown in the \texttt{L13-24} and \texttt{L32-36} of Code~\ref{lst:motion_shifting}.

\setlength{\TPHorizModule}{1em} 
\setlength{\TPVertModule}{1em}

\begin{textblock}{10}(10.7,22.3) 
\begin{tikzpicture}
    \node[fill=gray!20, text opacity=1, fill opacity=0.5, text width=13.8cm, minimum height=13.3em, align=left] at (0,0) { };
\end{tikzpicture}
\end{textblock}

\begin{textblock}{10}(10.7,42.2) 
\begin{tikzpicture}
    \node[fill=gray!20, text opacity=1, fill opacity=0.5, text width=13.8cm, minimum height=6.3em, align=left] at (0,0) { };
\end{tikzpicture}
\end{textblock}

\lstset{style=pythonstyle}
\begin{lstlisting}[caption={Pseudo codes for motion sequence shifting.}, label={lst:motion_shifting}]
def forward(self, x, cond, time_shift_steps_end, time_shift_ratio):
    B, T, D = x.shape
    H = self.num_head
    
    # B, T, 1, D
    query = self.query(self.norm(x)).unsqueeze(2)
    # B, 1, T, D
    key = self.key(self.norm(x)).unsqueeze(1)
    query = query.view(B, T, H, -1)
    key = key.view(B, N, H, -1)

    # shifting queries and keys
    if self.step <= time_shift_steps_end:
        part1 = int(key.shape[1] * time_shift_ratio)
        part2 = int(key.shape[1] * (1 - time_shift_ratio))
        q_front_part = query[0, :part1, :, :]
        q_back_part = query[0, -part2:, :, :]
        new_q = torch.cat((q_back_part, q_front_part), dim=0)
        query[1] = new_q
        
        k_front_part = key[0, :part1, :, :]
        k_back_part = key[0, -part2:, :, :]
        new_k = torch.cat((k_back_part, k_front_part), dim=0)
        key[1] = new_k
                
    # B, T, N, H
    attention = torch.einsum('bnhd,bmhd->bnmh', query, key) / math.sqrt(D // H)
    weight = self.dropout(F.softmax(attention, dim=2))
    value = self.value(self.text_norm(cond)).view(B, T, H, -1)

    # shifting values
    if self.step <= time_shift_steps_end:
        v_front_part = value[0, :part1, :, :]
        v_back_part = value[0, -part2:, :, :]
        new_v = torch.cat((v_back_part, v_front_part), dim=0)
        value[1] = new_v
    y = torch.einsum('bnmh,bmhd->bnhd', weight, value).reshape(B, T, D)
    return y
\end{lstlisting}

\clearpage

\textbf{Example-based motion generation.} To generate diverse motions driven by the same example, we only need to shuffle the order of queries in self-attention, which is shown in \texttt{L13-23} of Code~\ref{lst:example_based}.

\begin{textblock}{10}(10.7,21.1) 
\begin{tikzpicture}
    \node[fill=gray!20, text opacity=1, fill opacity=0.5, text width=13.8cm, minimum height=12.4em, align=left] at (0,0) { };
\end{tikzpicture}
\end{textblock}

\lstset{style=pythonstyle}
\begin{lstlisting}[caption={Pseudo codes for example-based motion generation.}, label={lst:example_based}]
def forward(self, x, cond, steps_end, _seed, chunk_size, seed_bar):
    B, T, D = x.shape
    H = self.num_head
    
    # B, T, 1, D
    query = self.query(self.norm(x)).unsqueeze(2)
    # B, 1, T, D
    key = self.key(self.norm(x)).unsqueeze(1)
    query = query.view(B, T, H, -1)
    key = key.view(B, N, H, -1)
    
    # shuffling queries
    if self.step == steps_end:
        for id_ in range(query.shape[0]-1):
            with torch.random.fork_rng():
                torch.manual_seed(_seed)
                tensor = query[0]
                chunks = torch.split(tensor, chunk_size, dim=0)
                shuffled_index = torch.randperm(len(chunks))
                shuffled_chunks = [chunks[i] for i in shuffled_index]
                shuffled_tensor = torch.cat(shuffled_chunks, dim=0)
                query[1+id_] = shuffled_tensor
                _seed += seed_bar
                
    # B, T, T, H
    attention = torch.einsum('bnhd,bmhd->bnmh', query, key) / math.sqrt(D // H)
    weight = self.dropout(F.softmax(attention, dim=2))
    value = self.value(self.text_norm(cond)).view(B, N, H, -1)
    y = torch.einsum('bnmh,bmhd->bnhd', weight, value).reshape(B, T, D)
    return y
\end{lstlisting}

\textbf{Motion style transfer.} In the generation of two motions (\texttt{B=2}), we only need to replace the query of the second motion with the first one, which is shown in \texttt{L13-14} of Code~\ref{lst:style_transfer}.

\begin{textblock}{10}(10.7,58.7) 
\begin{tikzpicture}
    \node[fill=gray!20, text opacity=1, fill opacity=0.5, text width=13.8cm, minimum height=3.7em, align=left] at (0,0) { };
\end{tikzpicture}
\end{textblock}

\newcommand{\highlight}[1]{%
    \colorbox{yellow!50}{\parbox{\dimexpr\linewidth-2\fboxsep}{#1}}}

\lstset{style=pythonstyle}
\begin{lstlisting}[caption={Pseudo codes for motion style transfer.}, label={lst:style_transfer}]
def forward(self, x, cond, steps_end):
    B, T, D = x.shape
    H = self.num_head
    
    # B, T, 1, D
    query = self.query(self.norm(x)).unsqueeze(2)
    # B, 1, T, D
    key = self.key(self.norm(x)).unsqueeze(1)
    query = query.view(B, T, H, -1)
    key = key.view(B, N, H, -1)

    # style transfer
    if self.step <= self.steps_end:
        query[1] = query[0]
                
    # B, T, T, H
    attention = torch.einsum('bnhd,bmhd->bnmh', query, key) / math.sqrt(D // H)
    weight = self.dropout(F.softmax(attention, dim=2))
    value = self.value(self.text_norm(cond)).view(B, N, H, -1)
    y = torch.einsum('bnmh,bmhd->bnhd', weight, value).reshape(B, T, D)
    return y
\end{lstlisting}

\begin{figure}[!ht]
    \centering
    \vspace{-1.5em}
    \includegraphics[width=0.9\linewidth]{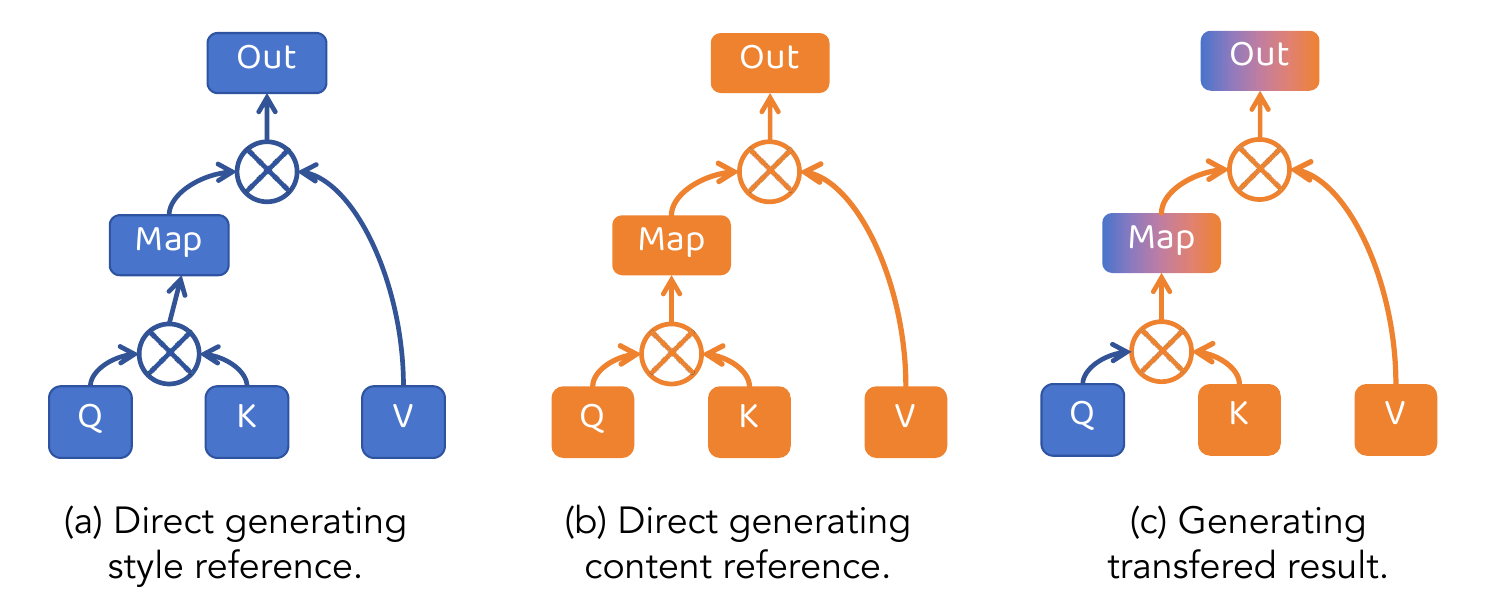}
    \caption{The illustration of motion style transfer process. (a) Direct generating style reference: The style information is generated directly using the query ($\mathbf{Q}$), key ($\mathbf{K}$), and value ($\mathbf{V}$) from the style reference motion sequence (blue). (b) Direct generating content reference: The content information is generated directly from the content reference motion sequence (orange). (c) Generating transferred result: The final transferred motion sequence combines the style from the style reference sequence with the content from the content reference sequence, using $\mathbf{Q}$ from the style reference (blue) and $\mathbf{K}$, $\mathbf{V}$ from the content reference (orange).}
    \label{fig:style_transfer_detail}
\end{figure}

\subsection{Supplementary for Motion Style Transfer}
As discussed in the main text, motion style transfer is accomplished by replacing the query ($\mathbf{Q}$) from the content sequence ($\mathbf{M_2}$) with that from the style sequence ($\mathbf{M_1}$). This replacement ensures that while the content features from $\mathbf{M_2}$ are preserved, the style features from $\mathbf{M_1}$ are adopted, resulting in a synthesized motion sequence that captures the style of $\mathbf{M_1}$ with the content of $\mathbf{M_2}$.
\cref{fig:style_transfer_detail} provides a visual explanation of this process. The self-attention mechanism plays a crucial role, where the attention map determines the correspondence between the style and content features. The pseudo code snippet provided in Code~\ref{lst:style_transfer} exemplifies this process. By setting ``\texttt{query[1] = query[0]}'' in the code, the query for the second motion ($\mathbf{M_2}$) is replaced by that of the first motion ($\mathbf{M_1}$), which effectively transfers the motion style from $\mathbf{M_2}$ to $\mathbf{M_1}$. In summary, this motion style transfer method allows one motion sequence to adopt the style characteristics of another while maintaining its content.


\newpage

\section{Details of Action Counting in a Motion}

\label{sec:app_detial_count}

The detailed process of action counting is described in Code~\ref{lst:count}. The attention map is first smoothed using a Gaussian filter to eliminate noise, ensuring that minor fluctuations do not affect peak detection. We then downsample the smoothed matrix to reduce computational complexity and normalize it within a 0-1 range for consistent peak detection across different motions.

The pseudo code provided demonstrates the complete process, including peak detection using height and distance thresholds. The experimental results indicate that this approach is more reliable and less sensitive to noise compared to using the root trajectory, thus confirming the effectiveness of our method in accurately counting actions within a generated motion sequence.

\lstdefinestyle{pythonstyle}{language=Python, basicstyle=\ttfamily, keywordstyle=\bfseries, commentstyle=\itshape, stringstyle=\color{red}, showstringspaces=false}
\begin{lstlisting}[caption={Pseudo codes for action counting.}, label={lst:count}]
"""
Input: matrix (the attention map array with shape (T, T))
Output: float (counting number)
"""

# Apply Gaussian smoothing via gaussian_filter in scipy.ndimage
smoothed_matrix = gaussian_filter(matrix, sigma=0.8)

# Attention map down-sampling
downsample_factor = 4
smoothed_matrix = downsample_matrix(smoothed_matrix, downsample_factor)

# Normalize the matrix to 0-1 range
normalized_matrix = normalize_matrix(smoothed_matrix)

# Detect peaks with specified height and distance thresholds
height_threshold = normalized_matrix.mean() * 3 # you can adjust this
distance_threshold = 1  # you can adjust this
peaks_positions_per_row = detect_peaks_in_matrix(normalized_matrix, height=height_threshold, distance=distance_threshold)

# Display the peaks positions per row
total_peak = sum([len(i) if len(i) > 0 else 0 for i in peaks_positions_per_row])
sum_ = sum([1 if len(i) > 0 else 0 for i in peaks_positions_per_row])

return total_peak / sum_
\end{lstlisting}

. 
\end{document}